\documentclass{article}

  \usepackage[main, final]{neurips_2026}

\usepackage[utf8]{inputenc} 
\usepackage[T1]{fontenc}    
\usepackage{hyperref}       
\usepackage{url}            
\usepackage{booktabs}       
\usepackage{amsfonts}       
\usepackage{nicefrac}       
\usepackage{microtype}      
\usepackage{xcolor}         

\usepackage{amsmath}
\usepackage{amssymb}
\usepackage{amsthm}
\usepackage{graphicx}
\usepackage{comment}
\usepackage[inline]{enumitem}
\usepackage[olditem,oldenum]{paralist} 
\usepackage{multirow}
\usepackage{subcaption}
\usepackage[capitalize,noabbrev]{cleveref}
\usepackage{wrapfig}
\usepackage{pdfpages}
\theoremstyle{plain}
\newtheorem{theorem}{Theorem}[section]
\newtheorem{property}{Property}[section]

\theoremstyle{definition}

\theoremstyle{remark}
\newtheorem{remark}[theorem]{Remark}

\title{Learning Optimal Transport Plans Via Autoregressive Token Regression}

\author{%
  Ivan Jaen Marquez \qquad Sotirios Panagiotis Chytas \qquad Vikas Singh \\
  University of Wisconsin--Madison \\
  \texttt{ivanjaenm@cs.wisc.edu} \quad
  \texttt{chytas@wisc.edu} \quad
  \texttt{vsingh@biostat.wisc.edu} \\
  \centering
\href{https://github.com/vsingh-group/TOR-OT}{\texttt{https://github.com/vsingh-group/TOR-OT}}
}

\begin{document}

\maketitle

\begin{abstract}
Optimal Transport (OT) provides a principled framework for finding the minimum cost coupling between probability
distributions.
Classical solvers remain the gold standard for this task. At the same time, OT is an ideal testbed for
studying whether learning-based amortized models can internalize the structure of
OT type problems. Here, ground-truth solutions are available and 
marginal constraints are explicitly defined. Also, exact discrete solutions have well-characterized
sparsity. We describe ToR-OT, a transformer-based model that learns to predict
discrete OT plans through autoregressive token regression. We reframe OT into a
structured sequence generation task. ToR-OT decodes sparse transport plans as
sequences instead of solving an optimization task at test time. We focus on
three properties: (a) Zero-shot generalization: pretraining on a large synthetic
corpus of OT instances allows a single model to generalize to unseen distribution
pairs and heterogeneous cost metrics without retraining, (b) Sparse primal
decoding: by exploiting the inherent sparsity in exact discrete OT solutions, the
decoder predicts only nonzero entries (avoids dense coupling outputs), 
(c) Amortized inference: on low-dimensional discrete problems,
ToR-OT gives accurate approximate plans and compares favorably with
per-instance neural OT baselines. Overall, ToR-OT is not yet a
replacement for classical solvers, but a step toward understanding
how autoregressive transformers can learn structure of constrained
optimization problems like OT.
\end{abstract}

\section{Introduction}\label{sec:intro}
Optimal Transport (OT) gives a principled framework for comparing probability distributions. It computes the minimum cost of transforming one measure into another, and finds  use 
in fluid dynamics \cite{ot_fluids}, image processing \cite{ot_image_proc}, brain connectivity analysis \cite{ot_brain_connectivity}, and machine learning, especially in domain adaptation and generative modeling \cite{wgan}. Typically, OT solvers \cite{ot_book,NIPS2013_af21d0c9} treat each instance independently, solving each  pair of distributions from scratch. 
This instance-specific setup leaves open the question of whether some regularities shared across related problems can be learned and reused.

The above observation of learning/amortization has indeed been studied in the last few years, and a number of results have reported success with ``neural'' or learning-based approaches to OT \cite{korotin2023neural, ot_net}. Neural Optimal Transport (NeuralOT) methods parameterize the dual potentials or transport maps using neural networks. In a number of situations, this allows solution of large-scale problems that can approach classical solvers. 
However, these approaches frequently remain instance-specific in a different sense. Each model is optimized to solve OT between {\em one specific pair of input measures} (e.g., a Gaussian source distribution and a mixture-of-Gaussians target distribution). It requires retraining or at minimum, fine-tuning when shown new source-target pairs. Meta OT \cite{MetaOT} mitigates
these issues where a single model learns to predict dual potentials across multiple OT instances. This is accomplished by training on a distribution of problem pairs (e.g., transporting between different pairs of MNIST digits). But Meta OT is still trained for specific problem domains: a model trained on image-to-image transport does not directly port over to probability distributions over labels. Other recent ideas \cite{jiang2025embedding} have explored embedding distributions into latent spaces to compute transport maps, offering further improvements. Despite these developments, zero/few-shot generalization across  heterogeneous OT problem instances, e.g., different distribution or cost types, in the way that large language models generalize across diverse downstream tasks is not possible at this time to our knowledge.

On the other hand, there is broad agreement that transformer-based models show strong capabilities beyond the language domain. Recent works show that autoregressive transformers can solve arithmetic problems \cite{lee2023teaching}, discover symbolic structures in partial differential equations \cite{bhatnagar2025equationsinsightsunravelingsymbolic} and generate PDE solver code \cite{li2025codepdeinferenceframeworkllmdriven}. Surprisingly, a result last year showed that these models can even perform numerical regression by representing real values as token sequences \cite{song2025decodingbased}. This capability, called decoding-based regression, enables transformers trained with standard cross-entropy loss to predict ``numeric'' outcomes with good accuracy, comparable to traditional regression methods. This suggests that autoregressive pretraining could transfer to structured prediction problems whose outputs admit a meaningful sequential representation.

The above discussion raises a natural question: can autoregressive token regression, when shown a sufficiently large sample size of examples during pretraining, learn to solve optimization problems beyond simple regression? We study this question in the context of OT, a structured prediction task with well-defined ground truth solutions, a clear optimization objective, and a nice geometric structure. Unlike standard regression where outputs are single scalars, OT requires predicting sparse transport plans (matrices with constrained marginals). So, it is a compelling test case for whether transformer models can provide sensible results.

\paragraph{Contributions.} We give evidence that autoregressive token prediction, the mechanism behind LLMs, can be successfully adapted to solve OT-type structured optimization problems when shown sufficient training data. 
Our ToR-OT is not positioned as a replacement for classical OT solvers. 
Instead, we use OT as a controlled setting to investigate the degree to which transformers can internalize structural regularities in such constrained optimization problems. OT is useful for this purpose because its solutions are globally coupled, ground-truth optima can be obtained by classical solvers, and its polyhedral structure gives well-characterized sparsity. To this end, our experiments test distinct scientific hypotheses about learned OT structure, including causal interventions/linear probes that test whether the model represents and uses marginal-capacity information during generation.
We present a model that treats transport plan prediction as sequence generation rather than numerical optimization. 
We achieve (a) zero-shot generalization across unseen distribution pairs and (b) amortized efficiency through row-wise autoregressive decoding that predicts only nonzero entries.

\section{Background}\label{sec:background}
We first review two concepts used throughout the paper: (i) the discrete OT solution of interest (that we aim to predict), and (ii) a token regression module for predicting numerical outputs.

\subsection{Optimal Transport}
\label{sec:background_ot}

Let $\mu = \sum_{i=1}^n a_i \delta_{x_i}$ and $\nu = \sum_{j=1}^m b_j \delta_{y_j}$ be two empirical measures, where $a \in \Delta_n := \{a \in \mathbb{R}^n_+ : \sum_i a_i = 1\}$ and $b \in \Delta_m := \{b \in \mathbb{R}^m_+ : \sum_j b_j = 1\}$ are probability mass vectors supported on points $\{x_i\}_{i=1}^n \subset \mathcal{X}$ and $\{y_j\}_{j=1}^m \subset \mathcal{Y}$ respectively. Let $C \in \mathbb{R}_{+}^{n \times m}$ denote a ground-cost matrix with entries $C_{ij} = c(x_i, y_j)$ (e.g., Euclidean distance $C_{ij}=\|x_i-y_j\|_2$). In the discrete OT setting, the output is a \emph{transport plan} $\pi \in \mathbb{R}_{+}^{n \times m}$ where $\pi_{ij}$ gives the amount of mass moved from $x_i$ to $y_j$.

Based on this definition, we can distinguish two variations: the (i) \textbf{Monge} and (ii) \textbf{Kantorovich} formulations.
The Monge formulation (\Cref{eq:monge_discrete}) seeks a deterministic assignment $\sigma:\{1,\dots,n\}\to\{1,\dots,m\}$ and enforces \emph{no mass splitting}. The Kantorovich relaxation (\Cref{eq:kantorovich_discrete}) replaces the map $\sigma$ with a matrix $\pi$ that can split mass across targets and satisfies linear marginal constraints. The feasible set $\Pi(a,b) = \{\pi \ge 0 : \pi\mathbf{1}_m = a, \pi^\top\mathbf{1}_n = b\}$ is a convex polyhedron in $\mathbb{R}^{n \times m}$.

\noindent
\begin{minipage}[c]{0.51\textwidth}
\begin{equation}
\label{eq:monge_discrete}
\min_{\sigma}\;\Big\{\sum_{i=1}^{n} a_i C_{i,\sigma(i)}
\;\Big|\;
b_j=\!\!\sum_{i:\,\sigma(i)=j} a_i\ \ \forall j
\Big\}.
\end{equation}
\end{minipage}
\hfill
\begin{minipage}[c]{0.46\textwidth}
\begin{equation}
\label{eq:kantorovich_discrete}
\min_{\pi\ge 0}\;
\Big\{\langle \pi, C\rangle
\;\Big|\;
\pi\mathbf{1}_m=a,\ \pi^\top\mathbf{1}_n=b
\Big\}.
\end{equation}
\end{minipage}

\begin{wrapfigure}{!b}{0.41\textwidth}
  \vspace{0pt} 
  \centering
  \includegraphics[width=0.37\textwidth]{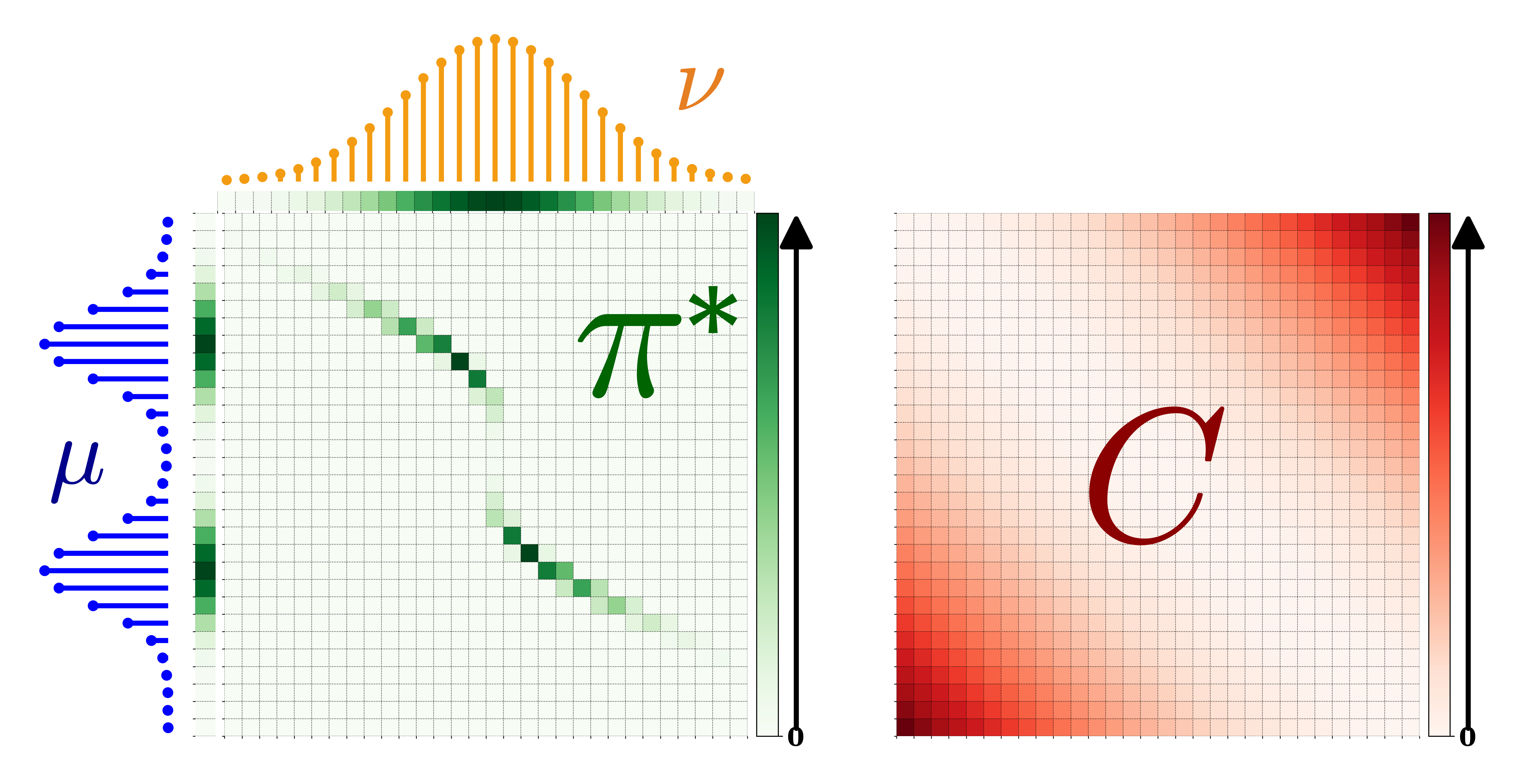}
  \caption{\footnotesize{Toy discrete OT instance showing the source measure $\mu$, target measure $\nu$, cost $C$, and an optimal plan $\pi^\star\in\mathbb{R}_+^{n\times m}$}}
  \label{fig:background_ot_toy}
    \vspace{-25pt} 
\end{wrapfigure}

Between the two formulations above, we write our prediction task around the Kantorovich \emph{coupling matrix} $\pi$ instead of the Monge \emph{map} $\sigma$. While a valid Monge map~(\ref{eq:monge_discrete}) may not exist when $n \neq m$ or when mass weights are unequal, the Kantorovich relaxation~(\ref{eq:kantorovich_discrete}) guarantees a feasible solution for any $(a,b)$ pair. Also, its optimal coupling matrix satisfies a fundamental sparsity property that makes sequence prediction a good fit:

\begin{property}[Sparsity]
For the discrete Kantorovich problem~\eqref{eq:kantorovich_discrete}, an optimal plan $\pi^*$ exists with at most $n+m-1$ nonzeros \cite[Prop.~3.4]{peyre2019computational}.
\label{ot_plan_sparsity_property}
\end{property}

We now have the main object of interest: the prediction target ($\pi \in \mathbb{R}_+^{n \times m}$), shown in \Cref{fig:background_ot_toy}.

\subsection{Token Regression (Decoding-Based Regression)}
\label{sec:token_regression}

As briefly mentioned above, decoding-based regression~\cite{song2025decodingbased} represents a real-valued target $y\in\mathbb{R}$ using a finite sequence of discrete tokens $(t_1,\dots,t_K)$ that encodes $y$ under a chosen numeric format, for example, a string such as \texttt{10.452} or \texttt{1e-10}. A causal decoder \cite{jurafsky2023speech} then models the conditional distribution over the token sequence given an input representation $\phi(\mathbf{x})\in\mathbb{R}^d$ via the standard autoregressive factorization:
\begin{equation}
\label{eq:token_reg_ar}
p_\theta(t_1,\ldots,t_K \mid \phi(\mathbf{x}))
= \prod_{k=1}^{K} p_\theta\!\left(t_k \mid \phi(\mathbf{x}), t_{<k}\right),
\end{equation}
where $t_{<k}=(t_1,\ldots,t_{k-1})$ denotes the prefix tokens.
Training follows the standard language model machinery of optimizing with next-token cross-entropy.
Notice that by representing numbers as short token sequences and training with next-token prediction, decoding gives numeric strings parsed to $\mathbb{R}$.

\begin{remark}[Variable-length outputs]
Although originally set up for scalar regression, the same mechanism can easily output \emph{variable-length} structured objects, e.g., lists, tuples, or sparse representations. We will use exactly this idea to serialize sparse OT plans.
\end{remark}

\section{Why Is OT a Structured Prediction Problem?}\label{sec:tor_ot_seq2seq}

We can start by asking: why would predicting an OT plan be difficult for standard neural heads?
One possible reason is that the transport plan $\pi$ which solves~\eqref{eq:kantorovich_discrete} is a globally \emph{constrained} object. Every entry must satisfy nonnegativity ($\pi_{ij} \ge 0$), row marginals  ($\sum_j \pi_{ij} = a_i$) for all $i$, as well as
column marginals ($\sum_i \pi_{ij} = b_j$) for all $j$.
These constraints couple all the entries because changing $\pi_{ij}$ requires compensating adjustments elsewhere. Otherwise, we will lose feasibility. Further, the sparsity pattern varies across instances because some solutions concentrate mass on few edges and others ``spread'' it.

\noindent\textbf{Possible mismatch?} We know that a dense regression head which outputs all $n \times m$ entries assumes (i) fixed output dimension (but $n,m$ vary across instances) and (ii) independent predictions (but entries are constrained). Further, sparsity is ignored although typical plans have $\mathcal{O}(n+m)$ nonzeros, not $\mathcal{O}(nm)$.
We reframe OT plan prediction as a \emph{structured output} with instance dependent complexity. While this is ill-suited for fixed size dense regressors, it is natural for variable-length sequences.

Although \Cref{ot_plan_sparsity_property} above is usually 
viewed as a geometric property of the transportation polytope, 
for sequence modeling one benefit is that any extreme-point OT plan can be described 
in an encoding of length $\mathcal{O}(n+m)$ rather than $\mathcal{O}(nm)$. 
Basically, instead of a dense $n \times m$ matrix, 
we can represent the plan as a sparse list of $({\rm source}, {\rm target}, {\rm mass})$ triples:
\begin{equation}
\pi \;\equiv\; \{(i_k, j_k, w_k)\}_{k=1}^{K}, \qquad K \le n+m-1,
\end{equation}
where each triple means moving mass $w_k$ from source $i_k$ to target $j_k$.

{\bf Serializing description for autoregressive decoding.}
We will simply predict the plan \emph{row by row}: for each source $i$, emit (target index, mass) pairs until the row budget $a_i$ is allocated, then proceed to row $i+1$. Autoregression 
will condition on previous rows, implicitly tracking remaining column capacities (the $b_j$ not yet consumed).
Formally, for row $i$, we can emit:
\begin{equation}
{\rm Row}_i = \{(j_\ell^{(i)}, w_\ell^{(i)})\}_{\ell=1}^{K_i}, \quad \text{where}\quad \sum_{\ell=1}^{K_i} w_\ell^{(i)} = a_i.
\label{ot_plan_row_rep}
\end{equation}
Now, conditioning on $\mathrm{Row}_{<i}$ shows to the decoder the remaining column-capacity state, allowing it to learn column-marginal behavior without an explicit constraint enforcement. The main benefit is that sparsity allows a compact serialization: essentially, OT plans become short sequences (length $\mathcal{O}(n+m)$) aligned with autoregressive generation.

\subsection{Reformulating OT as a Seq2Seq Task}

We cast OT as a sequence prediction task, an approach we call \textbf{ToR-OT} (\textbf{To}ken \textbf{R}egression for \textbf{OT}). Given a problem triplet $(\mu,\nu,c)$ which includes source distribution, target distribution, and cost type, the model produces a token sequence encoding the sparse plan $\pi$ via the row-wise representation~\eqref{ot_plan_row_rep}.

Autoregressive decoding fits this structure for two reasons. First, the output length is inherently variable: different instances induce different numbers of active entries, so a decoder emitting tokens until an \texttt{EOS} marker will avoid the inefficiency of fixed-size outputs. Second, the row-major serialization allows a clean probabilistic factorization. Generating row $i$ while conditioning on rows $\pi_{<i,:}$ allows implicit tracking of remaining target capacities. Formally, the decoder models
\begin{equation}
\label{eq:seq_factorization}
p(y \mid \mu,\nu,c) \;=\; \prod_{t=1}^{T} p(y_t \mid y_{<t}, \mu,\nu,c),
\end{equation}
where $y = (y_1,\ldots,y_T)$ is the complete token sequence encoding all rows.

\begin{remark}
\label{rem:paradigm}
We do {\em not} solve an optimization problem at test time. We will \emph{generate} a structured sequence whose semantics encode an OT plan. 
\end{remark}

\subsection{Will ToR-OT Generalize? Test-Time Compute Cost?}

The sequence formulation yields two main benefits at inference time, which we discuss below.

\begin{asparaenum}
\item \textbf{Test-time compute efficiency.}
Rather than running a new solve for each query, ToR-OT amortizes inference into generating a short token sequence, with no iterative optimization. Because only nonzero entries are emitted, we can avoid $\mathcal{O}(nm)$ dense outputs and the sequence length scales as $\mathcal{O}(n+m)$.

\item \textbf{Zero-shot generalization.}
Second, because the source and target measures $(\mu,\nu)$ are passed directly as input context, a single pretrained model can generalize to new distribution families without test-time retraining. 

\begin{remark}[Amortization and zero-shot generalization]
\label{rem:amortization}
ToR-OT is trained on {\em solved} OT instances: tuples $(\mu, \nu, c, \pi^*)$ where $\mu, \nu$ are sampled uniformly at random and $\pi^*$ is the optimal plan from a classical solver.
At test time, the model is evaluated on \emph{new} distribution family pairs {\em unseen} during training (e.g., Laplace source, Gamma target), and must generalize the transport structure learned across training pairs to predict plans for these unseen combinations. This is different from solving each instance from scratch or training separate networks per fixed $(\mu,\nu)$ pair.
ToR-OT amortizes learning \emph{across} distribution families, enabling zero-shot inference on unseen pairs.
\end{remark}

\end{asparaenum}


\section{ToR-OT: Tokenization, Set-to-Sequence Architecture, and Training}\label{sec:tor_ot_architecture}
\begin{wrapfigure}{r}{0.42\textwidth}
    \centering
    \vspace{-2pt}
    \includegraphics[width=\linewidth]{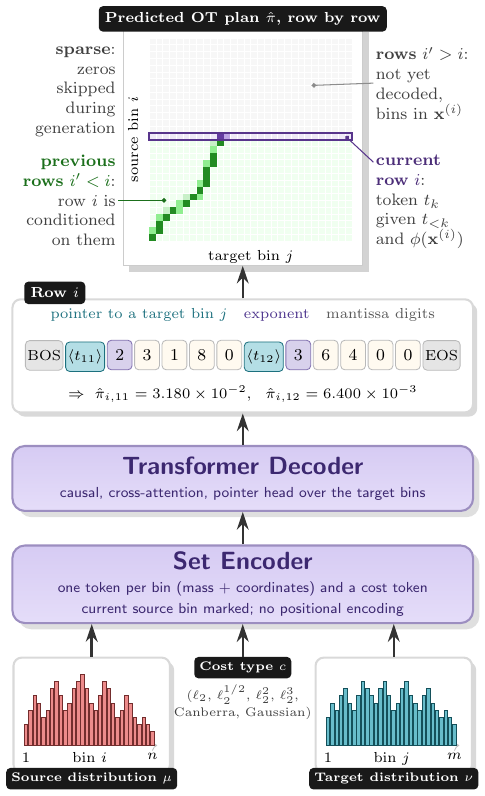}
    \vspace{-14pt}
    \caption{\footnotesize{ToR-OT architecture. Motivated by \cite{song2025decodingbased}, we view Optimal Transport as a sparse token-based multi-regression problem. Given a pair of source and target distributions, we train a Transformer to predict the nonzero values of the OT plan $\pi^*$ as a sequence of digit tokens.}}
    \label{fig:ToR-OT_architecture}
    \vspace{-14pt}
\end{wrapfigure}

We now give details for ToR-OT: it is a transformer-based model that predicts the OT plan $\pi(\mu, \nu)$ given two input measures $\mu, \nu$ and a cost metric $c$ (e.g., $\ell_2$, $\ell_2^2$). The plan is generated one row at a time
(\Cref{fig:ToR-OT_architecture}): for row $i$, a set encoder will embed the remaining problem (the unallocated source bins and residual target capacities), and a causal Transformer
decoder will emit the nonzero entries of $\hat\pi_{i,:}$ as a short token sequence. The masses of the row are then
subtracted from the target masses, and the next row is generated.

\paragraph{Serializing rows as token sequences.} A row is simply the token sequence:
\[
\texttt{BOS}\ \big(\langle t_j \rangle\ \langle E_k \rangle\ d_1\ d_2\ d_3\ d_4\big)^{*}\ \texttt{EOS},
\]
with one group of six tokens per nonzero entry, in increasing target index $j$. The pointer token $\langle t_j\rangle$
identifies the target bin, and the mass is written in scientific notation, $\hat\pi_{ij} = d_1.d_2 d_3 d_4 \times 10^{-k}$: that is, 
an exponent token $\langle E_k \rangle$, $k \in \{1, \dots, 9\}$, followed by four mantissa digits (masses below
$10^{-9}$ are dropped). For example, a source bin $i$ sending mass $0.0318$ to target 11 and $0.0064$ to target 12 is
\[
\texttt{BOS}\ \langle t_{11} \rangle\ \langle E_2 \rangle\ 3\ 1\ 8\ 0\ \langle t_{12} \rangle\ \langle E_3 \rangle\
6\ 4\ 0\ 0\ \texttt{EOS}.
\]
All other entries in row $i$ are implicitly zero.

\begin{remark}[Design rationale]
\label{rem:vocab_design}
A small, fixed alphabet (e.g., $|\mathcal{V}| = 15$) ensures:
(i) stable training across varying support sizes $n,m$,
(ii) arbitrary numeric precision via digit strings, and
(iii) compatibility with standard causal language modeling.
\end{remark}

\begin{remark}[Sequence length]
\label{rem:sequence_length}
A row with $K_i$ nonzero entries will have $6K_i + 2$ tokens. For problems where optimal plans are sparse (Property~\ref{ot_plan_sparsity_property}), the outputted rows remain relatively short, typically $K_i = \mathcal{O}(1)$ entries per row, giving total sequence length $\mathcal{O}(n+m)$ tokens.
\end{remark}

\subsection{Encoding Residual Transport State $\phi(\mathbf{x}^{(i)})$}

The key question is what information the decoder must condition on to generate feasible or near-feasible rows. For
row $i$, the encoder receives the \emph{remaining problem}
\begin{equation}
\label{eq:capacity}
\mathbf{x}^{(i)} = \Big(c,\ \{(x_{i'}, a_{i'})\}_{i'=i}^{n},\ \{(y_j, r^{(i)}_j)\}_{j=1}^{m}\Big), \qquad
r^{(i)}_j = b_j - \sum_{i' < i} \hat\pi_{i'j},
\end{equation}
that is, the cost type $c$, the source bins not yet decoded with their masses, and all target bins with their
remaining capacities. It maps $\mathbf{x}^{(i)}$ to one latent state per element: $\phi(\mathbf{x}^{(i)}) = \big(h_c,\ h^{\text{src}}_{i}, \ldots, h^{\text{src}}_{n},\ h^{\text{tgt}}_{1}, \ldots,
h^{\text{tgt}}_{m}\big)$.
The decoder attends to all of these states, with the target states $h^{\text{tgt}}_j$ 
forming the candidate set for pointer selection (\Cref{sec:decoder}).

\paragraph{Tracking residual state across rows.} Row $i$ is conditioned on the past rows through the remaining
capacities $r^{(i)}_j$~\eqref{eq:capacity}. Once the past rows are fixed, the completion of
the plan depends on them through $r^{(i)}$. Since the encoder ignores input order, bin $i$ carries a learned
marker. We also retain the unallocated source bins $i' > i$ in the input so the model can account for downstream demand. Basically, if a later source bin can reach a given target at lower cost, the current row should preserve capacity for it.

\paragraph{Coordinate-based bin embeddings.} Each bin becomes one token, embedded from its coordinate and its mass
(Appendix~\ref{app:training}), and the encoder is a Transformer without positional encoding. A bin is thus identified by its coordinate (not by its place in the input), and so, relabeling the bins only permutes the states. The same model therefore works for any support size within its training range, and the same architecture also works for two-dimensional supports.

\paragraph{Conditioning on heterogeneous cost metrics.} The cost type $c$ enters as one additional learned token. Every bin attends to
it, and a single model covers multiple costs. To
summarize, $\phi(\mathbf{x}^{(i)})$ provides the geometry, the remaining capacities and the cost one needs to generate
row $i$.

\subsection{Autoregressive Decoding Under Marginal Constraints}
\label{sec:decoder}

The decoder needs to be set up to answer the question whether row-wise generation will/can ignore global constraints (column marginals). Our decoder conditions on {\em both} the encoder embedding $\phi(\mathbf{x})$ and previous row tokens. The prefix encodes used column capacities: if $\sum_{i'<i} \hat\pi_{i'j}$ mass has been allocated to target $j$, the model learns to respect the remaining capacity $r^{(i)}_j = b_j - \sum_{i'<i} \hat\pi_{i'j}$. We do {\em not} hard-enforce the constraints. In terms of architecture, the decoder is just a standard causal Transformer with $L$ layers, $H$ attention heads, and hidden dimension $d$. It receives $\phi(\mathbf{x})$ as a prefix and generates the token sequence $\mathbf{t} = (t_1, \ldots, t_K)$ via:
\begin{equation}
\label{eq:decoder}
p_\theta(t_1, \ldots, t_K \mid \phi(\mathbf{x})) = \prod_{k=1}^{K} p_\theta(t_{k} \mid \phi(\mathbf{x}), t_{<k}).
\end{equation}
For each row $i$, the decoder outputs (index, mass) pairs until an \texttt{EOS} token signals completion.

\begin{remark}[Feasibility]
\label{rem:feasibility}
Autoregressive generation of row $i$ conditions on $\hat\pi_{<i,:}$ and the problem instance $(\mu,\nu,c)$. The prefix $\pi_{<i,:}$ encodes used column capacities, guiding the model toward marginal-respecting plans. However, ToR-OT does not enforce constraints via projection. Feasibility is learned from training data, and so is not guaranteed. We will discuss this shortly. When strict feasibility is required, predicted plans can be post-processed.
\end{remark}

To close up this section, we provide the details of setting training data and points not covered above. 

\subsection{Synthetic Pretraining Corpus Generation and Training}
\label{sec:pretraining}

We now describe in more detail where the ground-truth token sequences come from.
We generate a large synthetic corpus of OT instances as follows. 
First, we sample source and target distributions uniformly at random.
Then, we sample the cost type $c$ from $\{\ell_2, \ell_2^{1/2}, \ell_2^2, \ell_2^3, \text{Canberra}, \text{Gaussian}\}$.
We then solve the Kantorovich LP with POT's network simplex solver and keep only the nonzero entries of the optimal plan $\pi^*$, i.e., \texttt{res = ot.solve(C, a, b)} and $\pi^* = \texttt{res.sparse\_plan}$.

\begin{remark}[Sanity checks]
\label{rem:sanity}
Before tokenization, we verify that training targets satisfy:
nonnegativity: $\pi^*_{ij} \ge 0$ for all $i,j$ as well as row 
and column  constraints. Plans failing these checks ($<0.1\%$ of generated instances, due to numerical errors) are discarded.
\end{remark}

\begin{remark}[Coverage across random distributions]
\label{rem:coverage}
The training corpus spans diverse problem characteristics such as bin sizes, 1D/2D distributions, multiple cost metrics covering six different geometries.
This diversity provides \emph{transport structure}, i.e., how mass flows respond to cost and marginal changes rather than memorizing specific instance types.
\end{remark}

\paragraph{Training.} Our training objective is the standard cross-entropy loss over the generated token sequence. Given the digit-level tokenization, this transforms the task into a series of discrete classification tasks,
$\mathcal{L}(\theta) = -\sum_{k=1}^{K} \log p_\theta(t_k \mid \phi(\mathbf{x}), t_{<k})$.
To handle long sequences from row-wise serialization, during training our implementation uses FlashAttention-2 \cite{dao2024flashattention}. 

\subsection{Decoding Complexity and Runtime Scaling}
As described earlier, for Kantorovich formulation \eqref{eq:kantorovich_discrete}, there is always an optimal plan solution $\pi$ with at most $n+m-1$ nonzero entries \cite{peyré2025optimaltransportmachinelearners}, and the number of predicted nonzero bins is $\mathcal{O}(n)$. So, we can distinguish a \textit{worst-case} (highly uneven sparsity) and \textit{average-case} (approximately uniform sparsity) 
runtime of a transformer-based model.
Even when the \emph{total} number of nonzeros is $\mathcal{O}(n+m)$, they can still be concentrated in a few rows. For the balanced case $n=m$, we can have $K_1 = \mathcal{O}(n)$ and $K_i=\mathcal{O}(1)$ for $i>1$ while still satisfying $\sum_i K_i = \mathcal{O}(n)$. Then, $\sum_{i=1}^n K_i^2 = \mathcal{O}(n^2)$
so row-wise decoding with full self-attention has worst-case cost $\mathcal{O}(n^2)$.
If each row has a bounded number of targets, $K_i=\mathcal{O}(1)$, then $\sum_i K_i = \mathcal{O}(n)$ 
and $\sum_{i=1}^n K_i^2 = \mathcal{O}(n)$.
This means linear scaling in the self-attention under row-wise decoding. Here, the inference cost is dominated by the model's fixed per-token cost (and not the quadratic output size).

\subsection{Context Relative to Classical and Dual Solvers}
ToR-OT is a prototype for studying whether autoregressive models can internalize globally coupled constraints (and is not intended to replace exact solvers yet). We distinguish between two settings:

\begin{asparaenum}[\bfseries (a)]
\item \textbf{When ToR-OT helps.} ToR-OT is useful when solving many similar OT problems repeatedly, as in matching problems that recur within the same application domain, e.g., as in ~\cite{pachauri2012incorporating}. In this regime, the formulation amortizes learning across instances and can produce accurate approximate plans without per-instance neural optimization. Because we decode row-wise, generating a complete plan requires sequential row predictions rather than a single dense matrix output.

\item \textbf{When ToR-OT is a sub-optimal fit.} Classical OT solvers are always preferable when we seek certified optimality, or when constraints/costs fall far outside the training distribution. ToR-OT is also not a good fit if the plan is effectively dense (where sequence lengths grow and the advantage of predicting only nonzeros shrinks), such as in entropic OT~\cite{NIPS2013_af21d0c9}, where adding an entropy regularizer spreads mass across many edges.
\end{asparaenum}

\noindent\textbf{Primal versus dual OT prediction.}
In the learned OT setting, one key modeling choice is whether to learn OT structure in the primal or dual space. Several learning-based methods such as Meta OT and UNOT predict vector-valued {\em dual} potentials to avoid predicting the full matrix-valued coupling. These dual formulations typically use an \(\varepsilon\)-regularized OT objective: for a fixed \(\varepsilon>0\), one solves the regularized problem accurately, but it is an approximation of the original {\em non-entropic} Kantorovich problem. The choice of \(\varepsilon\) introduces a bias--stability tradeoff since larger \(\varepsilon\) values improve smoothness and numerical stability, while smaller values better approximate exact OT (albeit harder to optimize). ToR-OT uses the exact primal formulation of the Kantorovich LP (\Cref{eq:kantorovich_discrete}) and predicts its sparse primal coupling. Thus, while dual prediction avoids the dense $n \times m$ coupling by learning vector-valued potentials, ToR-OT instead uses the inherent sparsity of the exact primal solution to obtain a compact sequence representation of the coupling itself.

\section{Experiments and Results}\label{sec:results}
\providecommand{\tbd}[1]{\textcolor{red}{#1}}

We evaluate ToR-OT across a range of settings to test what a transformer model learns when trained on OT data. The
first four tasks test predictive and structural properties of the learned model and the last asks whether the model uses OT's
constraint structure during generation. We ask whether ToR-OT: (1) generalizes to unseen
distributions; (2) gives fast per-query inference through amortized pretraining, relative to per-instance neural baselines;
(3) handles several cost types with a single model; (4) yields approximately feasible plans with row-wise autoregressive decoding; and (5) uses the remaining marginal capacities during generation rather than pattern-matching inputs to
outputs.

\subsection{Experimental Setting}
\textbf{Model.} All one-dimensional results come from a single encoder--decoder Transformer (157.6M parameters; 12 heads, 6 encoder and 12 decoder layers) that generates the plan row by row. It is trained once, on 5M instances
covering six cost types ($\ell_2$, $\ell_2^{1/2}$, $\ell_2^2$, $\ell_2^3$, Canberra and Gaussian), support sizes from 5 to 100 bins
and marginals drawn uniformly from the probability simplex, with the exact plans of POT as targets. There is no fine-tuning per
cost, evaluation set or support size. Training took 24.4 wall-clock hours with data parallelism on 8 NVIDIA A100 GPUs.
A second model with the same architecture and recipe is trained on two-dimensional histograms (\Cref{sec:2d}).
Additional training and evaluation details are in Appendix~\ref{app:training}, and ablations in Appendix~\ref{app:scaling}.

\begin{figure*}[!b]
    \centering
    \includegraphics[width=\textwidth]{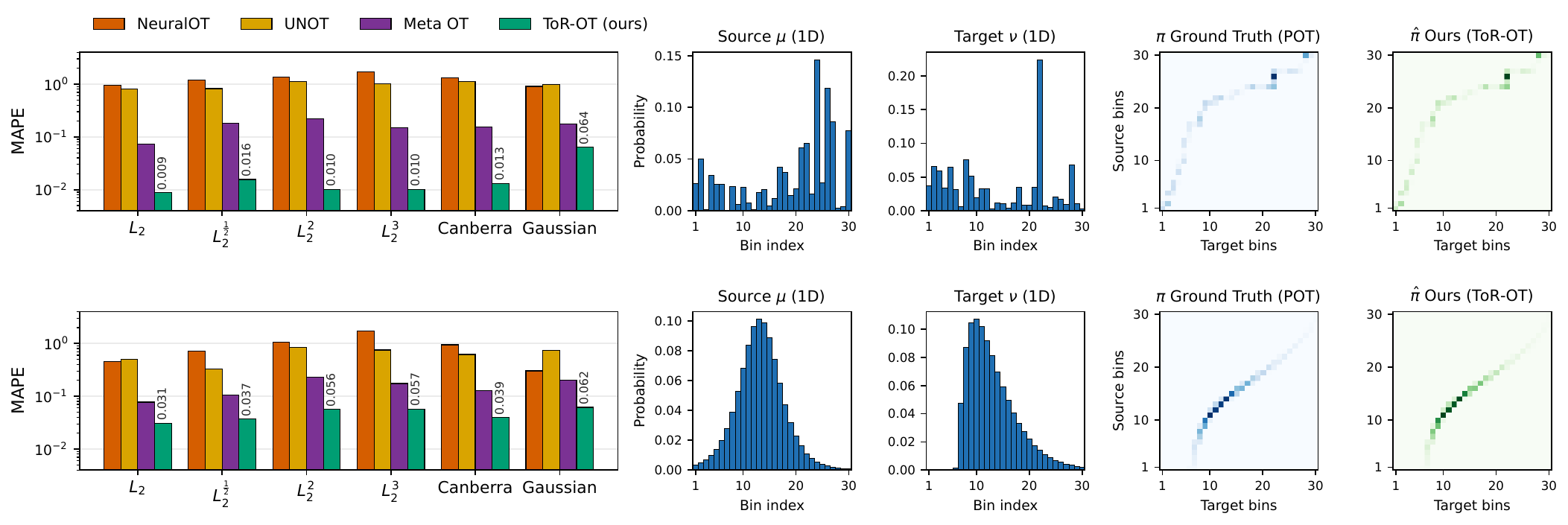}
    \vspace{-20pt}
    \caption{\footnotesize{Test set (top row) results and OOD (bottom row) results. (left) ToR-OT vs NeuralOT, Meta OT and UNOT. On all six costs, our model performs favorably. (right) a sample of the evaluation set, along with its ground truth and predicted OT plan. In the case of the test set, both source and target distributions are randomly sampled, while in the OOD case, they correspond to well-defined distributions, like Beta and Exponential.}}
    \label{fig:tor_ot_evaluation}
\end{figure*}

\textbf{Baselines.} We treat the exact solver of \textbf{POT} \cite{JMLR:v22:20-451} as the ground truth and compare
against three neural OT methods: \textbf{NeuralOT} \cite{korotin2023neural}, which trains a network for every new pair,
and two amortized solvers of entropic OT, \textbf{Meta OT} \cite{MetaOT} and \textbf{UNOT} \cite{geuter2025universal}.
As both assume a fixed cost, we train Meta OT and UNOT separately for each cost with the official code, choosing the regularization $\varepsilon \in \{0.01, 0.001\}$ on a held-out validation set, and reporting the mean over three training
seeds.

\textbf{Evaluation.} Problems have $n = m = 30$ bins ($n = 100$ and $200$ in Appendix~\ref{app:scaling-size}). The \textbf{test set} has 1{,}000 random pairs per cost from the training distribution (unseen instances); the \textbf{OOD set} has the 144 ordered pairs of 12 named distributions, each pair binned on a common grid of 30 bins that covers both distributions, so the narrower one can have empty bins
(Appendix~\ref{app:full-1d}). We
report the Mean Absolute Percentage Error (MAPE), $\sum_{ij} |\hat\pi_{ij} - \pi_{ij}| / \sum_{ij} \pi_{ij}$, of each
method against the solution of the problem it solves. Meta OT and UNOT solve entropic OT and are compared with the
converged Sinkhorn plan at their $\varepsilon$. ToR-OT and NeuralOT target the exact OT and are compared with the optimal
plan closest to their prediction, since the optimal plan is not always unique and POT returns
one of them. Other metrics and every cost are in Appendix~\ref{app:full-1d}.

\subsection{One-Dimensional Distributions}

We first evaluate two experimental families of one-dimensional distributions: \begin{enumerate*}
    \item \textbf{Test set}, and
    \item \textbf{OOD}
\end{enumerate*}. The test set corresponds to 1{,}000 held-out test pairs per cost of randomly generated source and target
distributions, similar in nature to the training set used for the pretraining. OOD stands for 144 pairs of
well-defined distributions (e.g., Beta, Cauchy, Exponential, Laplace) not seen during training.

\begin{wraptable}{r}{0.48\textwidth}
\vspace{-1.2em}
\centering
\footnotesize
\setlength{\tabcolsep}{4pt}
\caption{\footnotesize{Reduction of ToR-OT's mean MAPE relative to Meta OT, the strongest baseline, for every cost ($n = 30$).}}
\vspace{-5pt}
\label{tab:improvement}
\resizebox{\linewidth}{!}{%
\begin{tabular}{l|rrrrrr}
    \toprule
    Cost & $\ell_2$ & $\ell_2^{1/2}$ & $\ell_2^2$ & $\ell_2^3$ & Canberra & Gaussian \\
    \midrule
    Test set & $88\%$ & $91\%$ & $95\%$ & $93\%$ & $91\%$ & $64\%$ \\
    OOD set & $60\%$ & $65\%$ & $75\%$ & $67\%$ & $69\%$ & $69\%$ \\
    \bottomrule
\end{tabular}%
}
\vspace{-1em}
\end{wraptable}

\subsubsection{Test-Set Samples}
\Cref{fig:tor_ot_evaluation} (top row) reports test-set performance. On the left, we see that ToR-OT consistently produces more accurate solutions than the baselines across all six cost functions, each method scored against the solution of the problem it solves.

Meta OT is the strongest baseline on every cost, and ToR-OT reduces its error by roughly an order of magnitude on five
of the six costs (\Cref{tab:improvement}). The Gaussian cost is the hardest for ToR-OT, yet it still clearly improves
on Meta OT there. UNOT and NeuralOT are far less accurate in this setting, even though NeuralOT is trained separately on
every pair. All six costs are handled by a single ToR-OT model, without per-cost training.

\subsubsection{Out-of-Distribution Samples}
\begin{figure*}[!t]
    \centering
    \includegraphics[width=\textwidth]{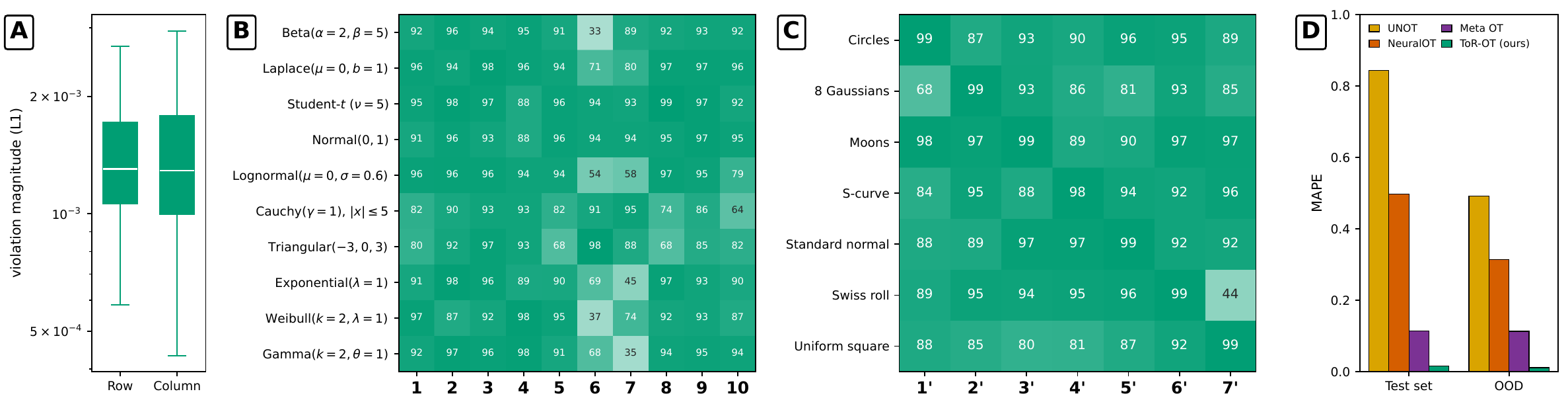}
    \vspace{-15pt}
    \caption{\footnotesize{\textbf{(A)} Violation of marginal constraints in ToR-OT across rows and columns. \textbf{(B)} Relative improvement of ToR-OT's MAPE over Meta OT, the strongest baseline, under $\ell_2^2$ for 100 of the 144 pairs of OOD one-dimensional distributions (all 144 are in Appendix~\ref{app:full-1d}). The difference is always positive and reaches up to $99\%$. X-axis labels: 1: Beta, 2: Laplace, 3: Student's t, 4: Normal, 5: Lognormal, 6: Cauchy, 7: Triangular, 8: Exponential, 9: Weibull, 10: Gamma. \textbf{(C)} The same for all 49 pairs of OOD two-dimensional distributions. The difference is always positive and reaches up to $99\%$. X-axis labels: 1': Circles, 2': 8 Gaussians, 3': Moons, 4': S curve, 5': Standard Normal, 6': Swiss roll, 7': Uniform square. \textbf{(D)} Mean MAPE of every method on the 2D test and OOD sets.}}
    \label{fig:tor_ot_results_plots}
\end{figure*}

Out-of-Distribution (OOD) samples are pairs of well-defined distributions, such as Normal, Exponential, and Beta, that differ strongly from the randomly generated train-set distributions. Evaluating on OOD samples tests both applicability to common distributions and generalization to unseen instances, ruling out memorization. Indeed, OOD accuracy keeps improving with training and, in a smaller-scale study, is unchanged when the model is trained on only a tenth of the corpus (Appendix~\ref{app:scaling}). In the OOD setting (\Cref{fig:tor_ot_evaluation}, bottom row), ToR-OT's error grows relative to the test set on five of the six costs, but it remains the lowest of all methods on every cost, 60--75\% below that of Meta OT (\Cref{tab:improvement}). \Cref{fig:tor_ot_results_plots} (B) shows the relative MAPE improvement over Meta OT under $\ell_2^2$ for 100 of the 144 distribution pairs; over all 144, ToR-OT is more accurate on 134, with a median improvement of 91\%. Per-pair results for the other five costs, and every method (for all metrics), are in Appendix~\ref{app:full-1d}.
\subsubsection{Runtime}
While classical solvers like POT are optimized for small-scale instances, ToR-OT operates in the amortized
regime. In the runtime comparison we can ask a more specific question: after pretraining, how does a single amortized
sequence model compare to per-instance neural OT methods? NeuralOT requires a dedicated optimization loop for each new
pair, whereas ToR-OT generates a plan in 1.09\,s on one A100 GPU. Including its 24.4\,h of training, ToR-OT is cheaper than NeuralOT after fewer than 100 instances. Meta OT and UNOT are faster still, but they solve entropic OT and their MAPE is at least 2.5 times ToR-OT's on every cost and set. The runtime result should therefore be viewed as some evidence for
amortization relative to per-instance neural baselines (not a replacement for classical OT solvers). More runtime
details can be found in Appendix~\ref{app:runtime}.

\subsubsection{Feasibility of predicted plans}
Despite no hard constraint enforcement, ToR-OT yields near-feasible plans. Figure~\ref{fig:tor_ot_results_plots}
\textbf{(A)} reports the violation magnitudes on rows $\|\hat{\pi}\mathbf{1}-a\|_1$ and columns
$\|\hat{\pi}^\top\mathbf{1}-b\|_1$ on the test set. The median for both is about 0.1\% of the total mass, and no
instance exceeds 1\%. Column residuals, however, vary more across instances than row residuals, on every test and OOD
set, and pooled over all sets, row residuals are also slightly smaller. Both effects are consistent with the autoregressive
structure: each row is generated with its own mass as input, so its sum is controlled directly, while every column sum
accumulates the errors of all rows. When improved feasibility is needed, two options are possible: a lightweight
Sinkhorn post-processing step and a symmetric two-pass approach. Full details and ablations for both strategies are in
Appendix~\ref{app:feasibility}.

\subsection{Mechanistic Interpretability}
Our goal here is to determine  whether the transformer represents and uses structural regularities of OT. Three experiments (Appendix~\ref{app:mechanistic}) give good evidence. 
We use a \textit{causal intervention}, modifying the generation prefix to exhaust a column's budget. This forces $r^{(i)}_j = b_j - \sum_{i'<i}\hat\pi_{i'j} = 0$, while leaving model weights and the problem instance unchanged. We see that the model adapts routing in 99\% of cases, tracking remaining capacity as a continuous quantity. \textit{Linear probing} recovers remaining column capacities from final-layer hidden states with $R^2=0.7135$. Accuracy rises as generation progresses, ruling out static input encoding. \textit{Attention pattern analysis} over 144 maps shows three consistent head types: global constraint heads, running row-budget accumulators, and spatial allocation heads.
These results are consistent with the model representing and using OT constraint state during generation.

\subsection{Extension to Multivariate Distributions: The Two-Dimensional Case}
\label{sec:2d}
\begin{figure*}[!t]
    \centering
    \includegraphics[width=\textwidth]{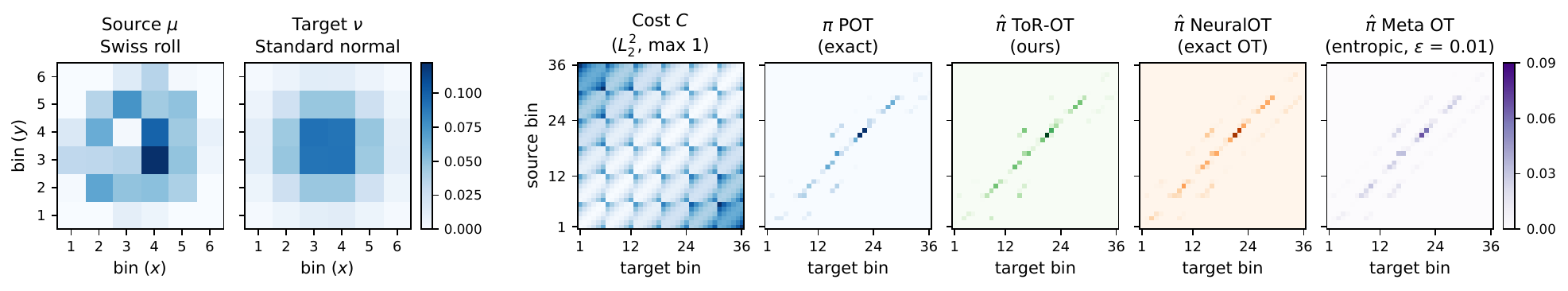}
    \vspace{-15pt}
    \caption{\footnotesize{ToR-OT, NeuralOT and Meta OT on a 2D OOD pair (Swiss roll $\to$ Standard Normal under $\ell_2^2$): the $6 \times 6$ histograms, the cost matrix and the plans over the 36 flattened bins, on one color scale. The exact same formulation as 1D is used. The only difference is that the 2D model is trained on 2D histograms, with each bin embedded from its 2D coordinate; the six costs are the same as in 1D.}}
    \label{fig:cost_variable_lp1_combined}
\end{figure*}

Finally, we extend the evaluation to the case of 2D histograms and focus on in-distribution test performance and OOD generalization. The OOD evaluation set comprises 49 pairs of well-defined 2D distributions (e.g. Standard Normal, S curve). This setting highlights that ToR-OT translates to 2D without any fundamental changes to the architecture (by flattening 2D bins into 1D). The only additional input is a 2D-specific cost for the model to capture the underlying geometry (\Cref{fig:cost_variable_lp1_combined}).

The 2D model shares the architecture and training recipe of the 1D model and is trained on 4M pairs of 2D histograms on $k \times k$ grids ($k$ from 3 to 10). We evaluate on a $6 \times 6$ grid ($n = m = 36$), with a test set of 1{,}000 random pairs per cost. Per pair, ToR-OT is more accurate than Meta OT on all 49 OOD pairs, with a median improvement of 93\% and at least 44\% (\Cref{fig:tor_ot_results_plots} (C)). On the input pair of \Cref{fig:cost_variable_lp1_combined}, ToR-OT's plan follows the sparse structure of the exact plan, while NeuralOT's has a larger error. These gains are consistent across all six costs, where ToR-OT outperforms Meta OT on every OOD pair of distinct distributions (Appendix~\ref{app:full-2d}).

\section{Related work}\label{sec:results}
The ``neural'' OT methods primarily focus on parametrizing the transport map with neural networks \cite{seguy2018large, korotin2023neural}. They usually exploit a connection between dual OT formulation and NNs, derived from Brenier's theorem \cite{Brenier_theorem}. Proposals in \cite{10.5555/3600270.3600767} and \cite{pmlr-v202-uscidda23a} leverage  this idea to propose novel extensions such as inferring OT maps for multiple pairs of measures, some of them with a pre-specified context.
One common feature this first set of neural OT methods share is that they seek to solve individual problem instances, similar to traditional optimization solvers. There is no learning from previously solved instances. This limitation has motivated a shift toward learned, amortized solvers that can transfer structure across instances.
Approaches such as Meta OT \cite{MetaOT, asadulaev2024neural} are good representatives: by training on a distribution of OT problems, they make inference on new pairs faster. They learn shared structure across tasks, and related approaches embed distributions into a latent space to produce problem-adapted OT maps or operators. While this improves reuse, it is often domain-based, since one usually restricts it to a fixed cost or fixed support size.
A more recent result, Universal Neural OT \cite{geuter2025universal} targets generalization across OT problems under a fixed cost. It uses Fourier Neural Operators (FNO) as a discretization-invariant architecture that can operate on inputs of varying resolution. To our knowledge, LLMs/Transformer models to solve OT problems in the style of ToR-OT have not been reported.



\section{Conclusion and Limitations}\label{sec:conclusions}
We described ToR-OT, a novel transformer-based token-regression formulation for predicting OT plans between probability distributions. We have shown that ToR-OT can effectively internalize the structure of OT and provide highly accurate predictions on both the test set and out-of-distribution samples, while offering amortized runtime efficiency. One takeaway is that sparse \emph{primal} OT plans can be effectively learned by sequence models. Operating in the exact discrete setting gives provably sparse plans (at most $n{+}m{-}1$ nonzeros, with no regularization bias), instance-dependent variable support that a standard autoregressive model handles naturally. 
And since ground truth is available, the mechanistic analysis reported here becomes possible. This is quite different from the neural OT literature's focus on continuous transport maps or entropy-regularized couplings.

{\em Limitations.} While we view ToR-OT as a promising step towards the study of transformers for OT, there are some limitations. First, scaling to larger supports (more bins) may benefit from richer source--target interactions in the encoder, e.g., attention biased by pairwise costs, which the model currently infers from bin coordinates and the cost type. Second, extending to higher-dimensional measures remains challenging: our study focuses on 1D/2D settings; a unified model that transfers across dimensions or operates on arbitrary (non-grid) supports would be a meaningful step toward more general learned OT solvers. Third, handling more general cost structures is required to better capture instance-specific geometry, especially under variable or data-dependent costs.

\begin{ack}
Marquez, Chytas and Singh were supported in part by 
NIH R01 AG092220. 
\end{ack}

\bibliographystyle{unsrtnat}
\bibliography{references}

\newpage
\section*{NeurIPS Paper Checklist}

\begin{enumerate}

\item {\bf Claims}
    \item[] Question: Do the main claims made in the abstract and introduction accurately reflect the paper's contributions and scope?
    \item[] Answer: \answerYes{} 
    \item[] Justification: We support our claims on OT generalization, amortized efficiency through the six core hypothesis tested in Section 5 Experiments and Results.

\item {\bf Limitations}
    \item[] Question: Does the paper discuss the limitations of the work performed by the authors?
    \item[] Answer: \answerYes{} 
    \item[] Justification: We discuss limitations in Section 7.

\item {\bf Theory assumptions and proofs}
    \item[] Question: For each theoretical result, does the paper provide the full set of assumptions and a complete (and correct) proof?
    \item[] Answer: \answerNA{} 
    \item[] Justification: This is an empirical paper and we do not include theoretical results.

    \item {\bf Experimental result reproducibility}
    \item[] Question: Does the paper fully disclose all the information needed to reproduce the main experimental results of the paper to the extent that it affects the main claims and/or conclusions of the paper (regardless of whether the code and data are provided or not)?
    \item[] Answer: \answerYes{} 
    \item[] Justification: \cref{sec:tor_ot_architecture} detail the model, dataset and training routine. Appendix \ref{app:training} describe the specific settings used for all experiments. The code for the experiments and plots will be released upon acceptance.

\item {\bf Open access to data and code}
    \item[] Question: Does the paper provide open access to the data and code, with sufficient instructions to faithfully reproduce the main experimental results, as described in supplemental material?
    \item[] Answer: \answerYes{} 
    \item[] Justification: We have disclosed all the information needed to reproduce the main experimental results of the paper. The code will be open-sourced upon acceptance, together with the instructions on how to reproduce the experiments.

\item {\bf Experimental setting/details}
    \item[] Question: Does the paper specify all the training and test details (e.g., data splits, hyperparameters, how they were chosen, type of optimizer) necessary to understand the results?
    \item[] Answer: \answerYes{} 
    \item[] Justification: The technical details pertaining to the exact setup are given in \cref{sec:tor_ot_architecture} and appendix \ref{app:training}. The code with instructions on how to run the full experiments will be open-sourced upon acceptance.

\item {\bf Experiment statistical significance}
    \item[] Question: Does the paper report error bars suitably and correctly defined or other appropriate information about the statistical significance of the experiments?
    \item[] Answer: \answerYes{} 
    \item[] Justification: Box plots and error bars are given and described where relevant.

\item {\bf Experiments compute resources}
    \item[] Question: For each experiment, does the paper provide sufficient information on the computer resources (type of compute workers, memory, time of execution) needed to reproduce the experiments?
    \item[] Answer: \answerYes{} 
    \item[] Justification: We provided sufficient information on the total computer resources needed for training our model in experimental part of \Cref{sec:results}. Similarly we report inference runtimes.
    
\item {\bf Code of ethics}
    \item[] Question: Does the research conducted in the paper conform, in every respect, with the NeurIPS Code of Ethics \url{https://neurips.cc/public/EthicsGuidelines}?
    \item[] Answer: \answerYes{} 
    \item[] Justification: We have conducted research under the NeurIPS Code Of Ethics.

\item {\bf Broader impacts}
    \item[] Question: Does the paper discuss both potential positive societal impacts and negative societal impacts of the work performed?
    \item[] Answer: \answerNA{} 
    \item[] Justification: Our research aims to understand weather transformers can internalize the OT structure. It is a first step toward amortized, generalizable learned Transformer-based solvers and current research stage has minimal societal impact.
    
\item {\bf Safeguards}
    \item[] Question: Does the paper describe safeguards that have been put in place for responsible release of data or models that have a high risk for misuse (e.g., pre-trained language models, image generators, or scraped datasets)?
    \item[] Answer: \answerNA{} 
    \item[] Justification: Our paper poses no such risks.

\item {\bf Licenses for existing assets}
    \item[] Question: Are the creators or original owners of assets (e.g., code, data, models), used in the paper, properly credited and are the license and terms of use explicitly mentioned and properly respected?
    \item[] Answer: \answerYes{} 
    \item[] Justification: All the assets used in the paper are properly licensed. Our paper cites the original papers that produced the two libraries we used:
    \begin{itemize}
        \item POT 0.9.6.post1: \url{https://github.com/PythonOT/POT/releases/tag/0.9.6.post1}), distributed under MIT License.
        \item Flash Attention 2.8.3: \url{https://github.com/Dao-AILab/flash-attention}, distributed under BSD 3-Clause License.
    \end{itemize}

\item {\bf New assets}
    \item[] Question: Are new assets introduced in the paper well documented and is the documentation provided alongside the assets?
    \item[] Answer: \answerYes{} 
    \item[] Justification: Details for the synthetic dataset generation, model architecture and specific code details (e.g. training with data pallalelization, using FlashAttention) are included in the main body of the paper. We will make our code available and provide additional documentation to reproduce our experiments.

\item {\bf Crowdsourcing and research with human subjects}
    \item[] Question: For crowdsourcing experiments and research with human subjects, does the paper include the full text of instructions given to participants and screenshots, if applicable, as well as details about compensation (if any)? 
    \item[] Answer: \answerNA{} 
    \item[] Justification: 

\item {\bf Institutional review board (IRB) approvals or equivalent for research with human subjects}
    \item[] Question: Does the paper describe potential risks incurred by study participants, whether such risks were disclosed to the subjects, and whether Institutional Review Board (IRB) approvals (or an equivalent approval/review based on the requirements of your country or institution) were obtained?
    \item[] Answer: \answerNA{} 
    \item[] Justification: 

\item {\bf Declaration of LLM usage}
    \item[] Question: Does the paper describe the usage of LLMs if it is an important, original, or non-standard component of the core methods in this research? Note that if the LLM is used only for writing, editing, or formatting purposes and does \emph{not} impact the core methodology, scientific rigor, or originality of the research, declaration is not required.
    \item[] Answer: \answerNA{} 
    \item[] Justification: 

\end{enumerate}
\newpage
\appendix
\newcommand{\appitem}[1]{\item Appendix~\ref{#1}: \nameref{#1}\dotfill\pageref{#1}}

\section*{Appendix}
\subsection*{Table of Contents}
\begin{itemize}
    \appitem{app:training}       
    \appitem{app:runtime}        
    \appitem{app:feasibility}    
    \appitem{app:mechanistic}    
    \appitem{app:scaling}        
    \appitem{app:full-1d}        
    \appitem{app:full-2d}        
\end{itemize}

\newpage
\appendix
\newenvironment{apitems}{\begin{itemize}\setlength{\itemsep}{0pt}\setlength{\parskip}{0pt}}{\end{itemize}}

\section{Experimental Details: ToR-OT Configuration}
\label{app:training}

\subsection{Training Data (1D)}
\begin{apitems}
    \item \textbf{Corpus:} \texttt{ivanjaenm/torot\_data\_5M}; 5M training and 2.5K validation instances
    \item \textbf{Support:} bin centres of a regular grid on $[0,1]$; $n = m$ uniform in $\{5, \dots, 100\}$
    \item \textbf{Marginals:} $a, b \sim \mathrm{Dirichlet}(1, \dots, 1)$, independent
    \item \textbf{Costs:} $L_2$, $L_2^{1/2}$, $L_2^2$, $L_2^3$, Canberra, Gaussian ($\sigma = 0.2$), one per instance;
          $L_2^p = \lVert x - y \rVert_2^p$
    \item \textbf{Targets:} exact plans from POT's network simplex (\texttt{ot.emd})
    \item \textbf{Augmentation:} transpose with prob.\ 0.5;
\end{apitems}

\subsection{Model}
\begin{apitems}
    \item \textbf{Type:} encoder--decoder Transformer
    \item \textbf{Parameters:} 157.6M
    \item \textbf{Width / heads:} 768 / 12 (head dimension 64)
    \item \textbf{Layers:} 6 encoder, 12 decoder
    \item \textbf{Feed-forward:} SwiGLU, hidden width 2{,}048
    \item \textbf{Normalization:} pre-norm RMSNorm; QK-norm in all attention layers
    \item \textbf{Encoder input:} cost-type token + one token per bin. No positional encoding
    \item \textbf{Decoder:} causal, RoPE, cross-attention to the encoder
    \item \textbf{Column index:} pointer head over the target-bin encoder states
\end{apitems}

\subsection{Tokenization and Loss}
\begin{apitems}
    \item \textbf{Row sequence:} $\texttt{BOS}\,\big(\langle j \rangle\, \langle E_k \rangle\, d_1 d_2 d_3 d_4\big)^{*}\,\texttt{EOS}$,
          entries in increasing $j$
    \item \textbf{Mass format:} $0.d_1 d_2 d_3 d_4 \times 10^{-(k-1)}$, $k \in \{1, \dots, 9\}$ (4 significant digits)
    \item \textbf{Vocabulary:} 23 fixed tokens + one pointer per target bin
    \item \textbf{Loss:} cross-entropy; weight 1 for structure, pointer and exponent tokens, $0.5^{p-1}$ for mantissa
          digit $p$; ordinal label smoothing 0.05 per neighbouring digit; z-loss $10^{-4}$
\end{apitems}

\subsection{Optimization}
\begin{apitems}
    \item \textbf{Optimizer:} AdamW, $\beta = (0.9, 0.95)$, weight decay 0.1 (weight matrices only)
    \item \textbf{Learning rate:} $10^{-4}$, 2{,}000 warmup steps, cosine decay to $10^{-5}$
    \item \textbf{Steps:} 203{,}000 ($\approx$1.5 passes over the row examples)
    \item \textbf{Batch per GPU:} $\le$256 rows and $\le$32{,}768 encoder tokens; micro-batches of $\le$24{,}576 decoder
          tokens; global $\le$2{,}048 rows per step
    \item \textbf{Gradient clipping:} 1.0
    \item \textbf{Precision:} bf16
    \item \textbf{EMA:} 0.999 (EMA weights are evaluated and released)
    \item \textbf{Checkpoint:} lowest validation loss, checked every 5{,}000 steps (step 200{,}000)
    \item \textbf{Hardware:} 8$\times$ NVIDIA A100 40\,GB, 0.43\,s per step, $\approx$24.5\,h
\end{apitems}

\subsection{Evaluation}
\begin{apitems}
    \item \textbf{Test set:} 1{,}000 random pairs per cost (training distribution, disjoint seeds)
    \item \textbf{OOD set:} 144 pairs per cost (12 named distributions of the submitted version)
    \item \textbf{Support sizes:} $n = 30$ (main results), 100, 200 (2$\times$ the largest training size); 6{,}864 instances each
    \item \textbf{2D:} 1{,}000 test and 49 OOD pairs (7 named distributions) per cost, $6 \times 6$ grid
    \item \textbf{Reference plan:} the optimal plan closest to the prediction (LP over the optimal face); equals POT's plan when the optimum is unique
\end{apitems}

\subsection{2D Model}
\begin{apitems}
    \item \textbf{Settings:} same architecture, tokenizer, loss and optimizer as above
    \item \textbf{Corpus:} \texttt{ivanjaenm/torot\_data\_2M\_2d}; 2M instances on $k \times k$ grids in $[0,1]^2$,
          $k$ uniform in $\{3, \dots, 10\}$ ($n = k^2 \in [9, 100]$); 93.0M row examples
    \item \textbf{Training:} 2$\times$ NVIDIA A100 40\,GB, 248{,}000 steps, $\approx$26\,h
\end{apitems}

\newpage
\section{Runtime comparisons}
\label{app:runtime}

\begin{figure}[h]
    \centering

        \vspace{0pt}
        \centering
        \footnotesize
        \setlength{\tabcolsep}{6pt}

        \begin{tabular}{l r| r r r}
            \toprule
            \multirow{2}{*}{Method} & \multirow{2}{*}{Single sample} 
            & \multicolumn{3}{c}{Amortized analysis} \\
            & & $100$ & $500$ & $1000$ \\
            \midrule
            NeuralOT & $2627.08$ & $2627.08$ & $2627.08$ & $2627.08$ \\
            \textbf{ToR-OT}   & $\mathbf{1.09}$    & $\mathbf{2593.18}$ & $\mathbf{519.60}$  & $\mathbf{259.57}$ \\
            \bottomrule
        \end{tabular}
        \captionof{table}{\footnotesize{Runtime comparison (in seconds) of NeuralOT and ToR-OT. Provided an extensive pretraining on multiple OT instances, ToR-OT is much faster in estimating the OT map of a new sample. Additionally, even when we take into account the pretraining stage, the amortized runtime is below NeuralOT as few as 500 samples.}}
        \label{tab:runtime_comparison_table}
        \setlength{\tabcolsep}{6pt}

        \vspace{0pt}
        \centering

        \includegraphics[width=0.7\linewidth]{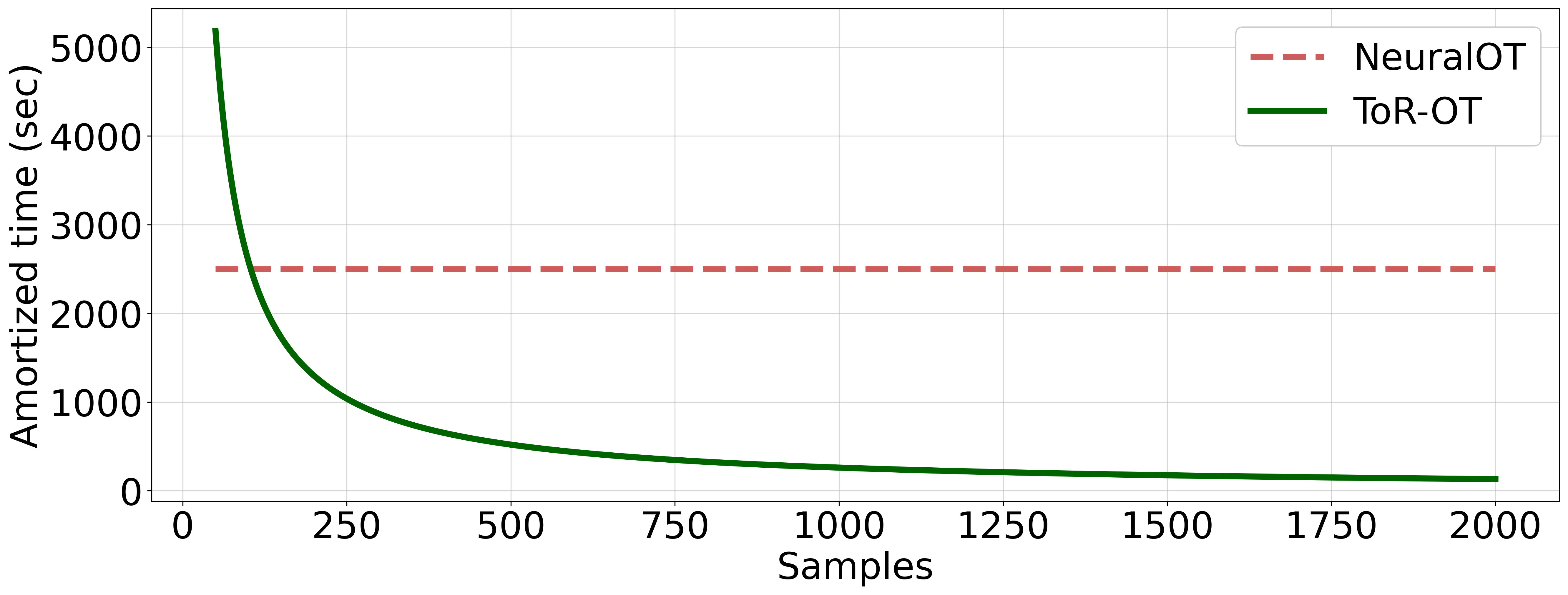}

        \captionof{figure}{\footnotesize{Amortized runtime of ToR-OT taking into account the pretraining time. For as few as 100 samples, ToR-OT drops below NeuralOT, which requires a new training loop for each new sample, with an average runtime of $2600$ seconds.}}
        \label{fig:amortized_time_comparison}

\end{figure}

The single-sample runtime comparison favors ToR-OT because its training cost is not included at inference time. To make the comparison more complete, we also report the amortized runtime obtained by spreading the pretraining cost over an increasing number of test samples. The right part of Table~\ref{tab:runtime_comparison_table} shows that the amortized cost decreases rapidly with the number of samples, falling below NeuralOT after as few as 100 samples. The full amortization curve is shown in \Cref{fig:amortized_time_comparison}.
\newpage
\section{Post-processing for improved OT feasibility}
\label{app:feasibility}

\subsection{Sinkhorn projections}
Although ToR-OT produces near-feasible plans (marginal violations of order $10^{-1}$--$10^{-2}$ in $L_1$), applications requiring improved marginal feasibility can use a lightweight post-processing step. We apply a few Sinkhorn iterations to the predicted plan $\hat{\pi}$ to project it onto the feasible set $\Pi(a,b)$ as shown in table \Cref{tab:sinkhorn_feasibility}.

\begin{table}[h]
\centering
\footnotesize
\begin{tabular}{l r r}
\toprule
Metric & Canonical ToR-OT & \textbf{ToR-OT + Sinkhorn} \\
\midrule
Max Marginal Violation & $1.15 \times 10^{-2}$ & $\mathbf{1.7 \times 10^{-4}}$ \\
Prediction Error (MAPE) & $1.52 \times 10^{-1}$ & $\mathbf{7.5 \times 10^{-2}}$ \\
\bottomrule
\end{tabular}
\vspace{0.5cm}
\caption{\footnotesize{Improved feasibility and accuracy with a few Sinkhorn iterations on the test set.}}
\label{tab:sinkhorn_feasibility}
\end{table}

Sinkhorn post-processing reduces the maximum marginal violation by nearly two orders of magnitude while simultaneously improving prediction accuracy by approximately $50\%$. Notably, ToR-OT already outperforms NeuralOT before any post-processing; after post-processing, the solution is both near-feasible to a smaller numerical tolerance and more accurate. This hybrid approach — generate a high-quality sparse plan with ToR-OT, then project to improved feasibility — provides a practical path to certified feasibility without sacrificing the efficiency gains of amortized inference.

\subsection{Symmetric decoding via transposed averaging}
An additional technique to improve feasibility is symmetric decoding: we run two forward passes with the source and target distributions swapped, then average the resulting plans. This reduces asymmetry between row and column marginal residuals as \cref{tab:symmetric_two_pass} shows.

\begin{table}[h]
\centering
\footnotesize

\begin{tabular}{l r r}
\toprule
Metric & Canonical (Single Pass) & \textbf{Symmetric (Two Pass Avg.)} \\
\midrule
Max Marginal Violation & $0.0272$ & $\mathbf{0.0218}$ \\
Prediction Error (MAPE) & $0.3025$ & $\mathbf{0.2538}$ \\
Total Inference Time & $1.133$s & $\mathbf{2.299}$s \\
\bottomrule
\end{tabular}
\vspace{0.5cm}
\caption{\footnotesize{Effect of symmetric two-pass averaging on feasibility, accuracy, and inference time.}}
\label{tab:symmetric_two_pass}
\end{table}

Symmetric decoding reduces asymmetry by $\sim 20\%$ and improves accuracy by $\sim 16\%$, at the cost of doubling inference time.

\newpage
\section{Mechanistic Interpretability studies}
\label{app:mechanistic}
We report three mechanistic studies designed to distinguish whether ToR-OT has internalized the mathematical structure of OT. These studies were motivated by methods from the mechanistic interpretability literature~\cite{bietti2023birth,olsson2022context}.

\subsection{Causal Intervention (inspired by induction head analysis)}
For a target row $i$, we modify the serialized prefix of previous rows to simulate exhausting column $j$'s global capacity: we set the cumulative allocation $\sum_{i' < i} \pi_{i'j}$ equal to $b_j$, so no mass remains available for column $j$. This is done without changing the global problem instance $(\mu, \nu, C)$ or any model weights, isolating the causal effect of the prefix state on generation. Across 100 such interventions, the model adapted its routing in \textbf{99\%} of cases: mean mass reduction to column $j$ was \textbf{87.3\%}, with \textbf{42\%} of interventions showing complete avoidance (mass $\to 0$) and the remaining \textbf{57\%} showing proportional reduction. The two modes of response indicate the model tracks remaining capacity as a \emph{continuous} quantity, not a binary flag. Since the only change was in the prefix state while the global instance remained fixed, this is direct causal evidence that ToR-OT conditions generation on the evolving global constraint state — not on shallow distributional patterns from the input.

\subsection{Linear Probing of Residual Streams}
We extracted final-layer hidden states immediately before each mass prediction and trained linear probes (ridge regression) to recover the remaining column capacity vector $r_j = b_j - \sum_{i' < i} \pi_{i'j} \in \mathbb{R}^{30}$ on held-out sequences. Ridge regression achieves $R^2 = \mathbf{0.7135}$, showing that the remaining constraint budget is \emph{linearly accessible} from the residual stream. The probing accuracy increases monotonically as generation progresses within a row, consistent with two-timescale learning dynamics~\cite{bietti2023birth}: the model dynamically updates its representation of the constraint state rather than encoding it statically from the input. This complements the causal intervention: the probing shows \emph{that} the constraint state is encoded; the intervention shows \emph{that it is used}.

\subsection{Attention Pattern Analysis}
We visualized last-layer self-attention during generation over 144 attention maps, grouping character-level tokens into three interpretable blocks: \emph{Prefix} (encoding $a$, $b$, $C$, $i/n$), \emph{Index tokens} ($j_k$), and \emph{Weight tokens} ($w_k$). Three consistent patterns appear across maps: (i) \textbf{Global heads} — several heads maintain persistent attention to the Prefix block throughout decoding, providing continuous reference to global marginal constraints, analogous to the global heads identified in~\cite{bietti2023birth}. (ii) \textbf{Running accumulator heads} — in longer rows, specific heads show weight-to-weight off-diagonal bands, attending to previously emitted mass tokens when computing subsequent weights, consistent with a running accumulator tracking the remaining row budget. (iii) \textbf{Spatial allocation heads} — tight index-to-index staircase patterns suggest local spatial boundary enforcement under the cost geometry. These three head types coordinate simultaneously across global constraint tracking, accumulated mass tracking, and spatial allocation, inconsistent with independent per-token prediction and supporting the interpretation that the model has learned a structured internal algorithm for OT plan generation.

\subsection{Why Autoregressive Decoding for OT?}

The convergent evidence above raises a structural question: why is autoregressive decoding specifically well-suited to OT, rather than, say, a fixed-size dense output head? Two properties of OT's mathematical structure suggest a natural alignment.

First, OT feasibility constraints are \emph{globally coupled}: each entry $\pi_{ij}$ constrains all others sharing row $i$ or column $j$. The serialized prefix $\pi_{<i,:}$ encodes accumulated column allocations, making the remaining capacity $b_j - \sum_{i' < i} \pi_{i'j}$ implicitly computable from context — without any explicit constraint enforcement mechanism. This is precisely what the causal intervention and linear probing results confirm: the model tracks this evolving capacity state in its hidden representations and uses it causally during generation.

Second, the polyhedral structure of the Kantorovich LP guarantees that optimal plans have \emph{at most} $n + m - 1$ nonzero entries, but the exact support is instance-dependent and varies in size. Autoregressive generation with an end-of-sequence token handles this variable-length structured output naturally, whereas a fixed-size dense prediction head must either over-allocate capacity or impose an arbitrary sparsification threshold post hoc. The three head types observed in the attention analysis: global constraint heads, running accumulators, and spatial allocation heads, correspond directly to the three computational demands this structure imposes: track global budgets, accumulate row mass, and resolve local cost geometry.

Together, these observations suggest that the strong empirical performance of ToR-OT is not incidental: structural alignments between the autoregressive mechanism and the mathematical properties of OT make this an effective inductive bias for learning to predict sparse transport plans.
\newpage
\clearpage
\makeatletter
\section{Scaling}
\label{app:scaling}
\label{app:scaling}

This appendix details the experiments on problem and training scale: support
sizes beyond the main setting (\ref{app:scaling-size}), training length (\ref{app:scaling-training}) and, on a smaller
testbed, corpus size, model size and design choices (\ref{app:scaling-testbed}).

\subsection{Support size}
\label{app:scaling-size}
The 1D model is trained on supports of 5 to 100 bins. We evaluate it on test and OOD sets built as in the
main experiments at $n = m = 100$, inside the training range, and at $n = m = 200$, twice the largest
training size; Meta OT and UNOT are retrained at every $n$ (three seeds each). Table~\ref{tab:app-support-size} compares
ToR-OT with Meta OT, the strongest baseline, each scored against the solution of the problem it solves.

On the test set, ToR-OT's error grows moderately from $n = 30$ to $n = 100$ (0.009--0.064
to 0.027--0.062), while Meta OT's grows several-fold
(0.072--0.221 to 0.28--0.73),
so the gap widens with $n$. On the OOD set at $n = 100$, ToR-OT remains more accurate on most costs, but falls
behind Meta OT on $L_2^2$ and $L_2^3$. At $n = 200$ the model does not extrapolate for $L_2^2$, $L_2^3$ and Gaussian: its plans
are far from optimal (MAPE above 1). For $L_2$, $L_2^{1/2}$ and Canberra, the error grows but stays below Meta OT's.
Supports larger than those seen in training therefore require training on them.

The plan length grows linearly with $n$ (386 tokens at $n = 30$,
1,296 at $n = 100$ and 2,596 at $n = 200$; log--log slope
1.01), and so does the time to generate one plan on one GPU (2.5\,s,
8.6\,s and 18.3\,s; slope 1.03).

\begin{table}[!ht]
    \centering
    \small
    \caption{\footnotesize{Mean MAPE at three support sizes, each method against the solution of the problem it solves (lower is
    better; bold: the better of the two). ToR-OT is the single 1D model trained on $n \le 100$; Meta OT is retrained at
    every $n$ (mean over three training seeds). Test: 1{,}000 pairs per cost; OOD: the 144 pairs of the 12 named
    distributions, binned at each $n$.}}
    \label{tab:app-support-size}
    \begin{tabular}{llrrrrrr}
        \toprule
        & & \multicolumn{2}{c}{$n = 30$} & \multicolumn{2}{c}{$n = 100$} & \multicolumn{2}{c}{$n = 200$} \\
        \cmidrule(lr){3-4} \cmidrule(lr){5-6} \cmidrule(lr){7-8}
        Set & Cost & ToR-OT & Meta OT & ToR-OT & Meta OT & ToR-OT & Meta OT \\
        \midrule
        \multirow{6}{*}{Test} & $L_2$ & \textbf{0.009} & 0.072 & \textbf{0.027} & 0.436 & \textbf{0.302} & 0.701 \\
         & $L_2^{1/2}$ & \textbf{0.016} & 0.180 & \textbf{0.037} & 0.282 & \textbf{0.375} & 0.395 \\
         & $L_2^2$ & \textbf{0.010} & 0.221 & \textbf{0.031} & 0.612 & 1.770 & \textbf{0.804} \\
         & $L_2^3$ & \textbf{0.010} & 0.148 & \textbf{0.031} & 0.581 & 1.767 & \textbf{0.733} \\
         & Canberra & \textbf{0.013} & 0.154 & \textbf{0.032} & 0.733 & \textbf{0.371} & 0.945 \\
         & Gaussian & \textbf{0.064} & 0.176 & \textbf{0.062} & 0.733 & 1.727 & \textbf{0.326} \\
        \midrule
        \multirow{6}{*}{OOD} & $L_2$ & \textbf{0.031} & 0.078 & \textbf{0.090} & 0.438 & \textbf{0.203} & 0.448 \\
         & $L_2^{1/2}$ & \textbf{0.037} & 0.104 & \textbf{0.101} & 0.200 & \textbf{0.232} & 0.289 \\
         & $L_2^2$ & \textbf{0.056} & 0.229 & 0.319 & \textbf{0.301} & 1.326 & \textbf{0.486} \\
         & $L_2^3$ & \textbf{0.057} & 0.175 & 0.318 & \textbf{0.247} & 1.347 & \textbf{0.286} \\
         & Canberra & \textbf{0.039} & 0.127 & \textbf{0.099} & 0.651 & \textbf{0.227} & 0.755 \\
         & Gaussian & \textbf{0.062} & 0.200 & \textbf{0.313} & 0.714 & 1.348 & \textbf{0.338} \\
        \bottomrule
    \end{tabular}
\end{table}

\subsection{Training length}
\label{app:scaling-training}
\begin{figure}[!ht]
    \centering
    \includegraphics[width=\linewidth]{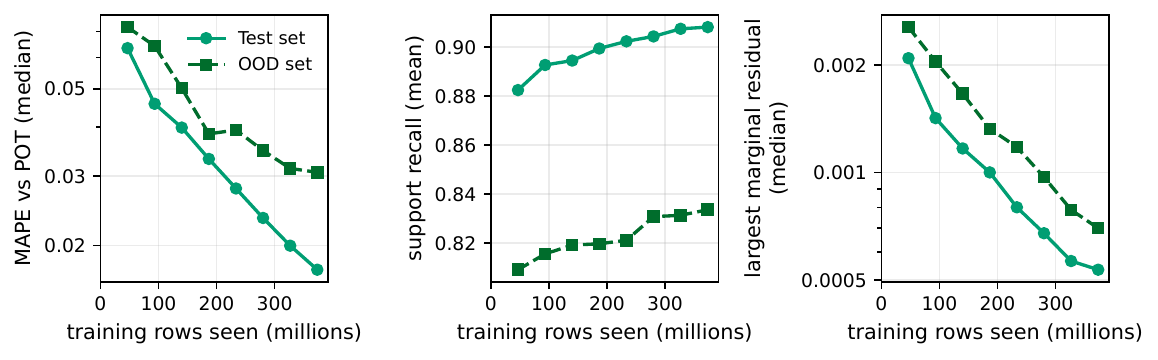}
    \caption{\footnotesize{Checkpoints of the 1D model (every 25{,}000 steps) vs training rows seen: median MAPE against POT's plan,
    mean support recall and median largest marginal residual, on 1{,}000 test pairs and the 864 named-distribution
    pairs of the corpus' OOD split.}}
    \label{fig:app-scaling-curve}
\end{figure}

To trace how accuracy depends on the amount of training data seen, at no additional training cost, we evaluate eight
checkpoints of the 1D model, every 25{,}000 steps from 47M to 374M
training rows ($\approx$1.5 passes over the corpus;
Fig.~\ref{fig:app-scaling-curve}). They are evaluated on fixed subsets of the corpus' own test splits: 1{,}000 random
pairs ($n$ from 5 to 100) and its 864 named-distribution pairs (144 per cost, $n = 30$), with MAPE against POT's plan,
computed during evaluation (for costs with tied optima, such as $L_2$, this overstates the error).

The error on the named-distribution pairs decreases steadily, from 0.072 to 0.031 (median), and
is still decreasing at the last checkpoint, as are the support recall and the marginal residuals, so longer training
would likely help further. The in-distribution error decreases faster (0.063 to 0.017), so the
ratio of OOD to test error grows from 1.1 to 1.8. All checkpoints come from one run, so the number of rows seen
and the learning-rate schedule change together; the effect of the corpus itself is isolated in
\ref{app:scaling-testbed}.

\subsection{Corpus size, model size and design choices}
\label{app:scaling-testbed}
\begin{figure}[!ht]
    \centering
    \includegraphics[width=0.9\linewidth]{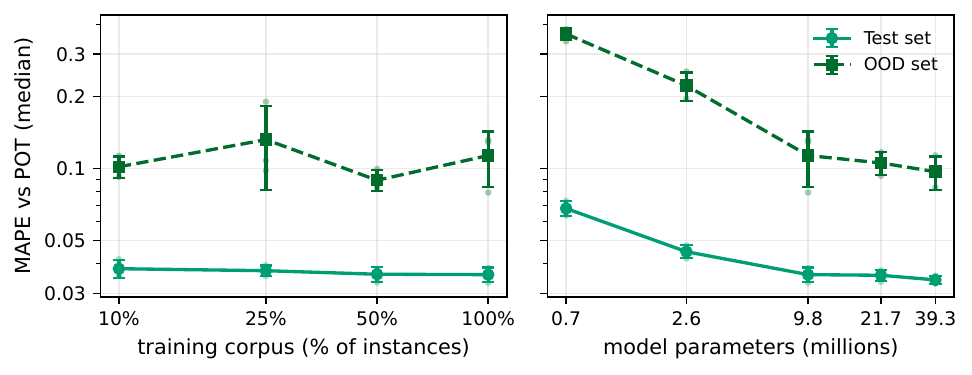}
    \caption{\footnotesize{Small testbed ($n \le 30$, the same training budget for every run): median MAPE against POT's plan vs
    the fraction of training instances (left) and the number of parameters (right). Line: mean over three training
    seeds; error bar: standard deviation; dots: the seeds.}}
    \label{fig:app-scaling-testbed}
\end{figure}

Retraining the full model for every configuration is too costly, so these ablations use a smaller testbed with the same
architecture, tokenizer and loss: a 9.8M-parameter model (width 256, 4 heads, 4 encoder and 6 decoder
layers) trained on the 270,508 instances with $n \le 30$ of a 1M-instance corpus (4.5M row examples).
Every run has the same training budget, 11{,}000 steps of 2{,}048 rows (22.5M rows, about one GPU-hour),
and every configuration is trained with three seeds. The models are evaluated on 1{,}311 test pairs and 864
named-distribution pairs ($n = 30$), with MAPE against POT's plan (Fig.~\ref{fig:app-scaling-testbed},
Table~\ref{tab:app-ablations}).

\textbf{Corpus size.} With the training budget fixed, the model trained on 10\% of the instances passes over them
50 times, against 5 times for the full corpus. Its test and OOD errors are
nevertheless the same within the variation across seeds (OOD: 0.089--0.132 for 10\% to 100\%;
the three seeds of the base recipe alone span 0.079--0.132). At this scale, the model does not
overfit the training instances even after many passes, and accuracy on unseen distributions does not depend on a large
pool of distinct instances.

\textbf{Model size.} Capacity matters more: the OOD error falls from 0.36 to 0.10 as
the model grows from 0.7M to 39.3M parameters, and the test error from
0.068 to 0.034, with smaller gains beyond 9.8M at this training
length.

\textbf{Design choices.} Perturbing the remaining capacities during training (App.~\ref{app:training}) matters most:
without it, the test error rises from 0.036 to 0.047 and the OOD error
from 0.113 to 0.165, as the model no longer sees the slightly inexact
capacities it produces at inference. The transpose augmentation and the digit-weighted loss change the errors by less
than the variation across seeds.

\begin{table}[!ht]
    \centering
    \small
    \caption{\footnotesize{Ablations on the small testbed: median MAPE against POT's plan, mean $\pm$ standard deviation over three
    training seeds. Each row changes one aspect of the base recipe.}}
    \label{tab:app-ablations}
    \begin{tabular}{llrcc}
        \toprule
        & Configuration & Parameters & Test set & OOD set \\
        \midrule
        Base recipe & width 256 (base) & 9.8M & 0.036 $\pm$ 0.003 & 0.113 $\pm$ 0.029 \\
        \midrule
        \multirow{4}{*}{Model size} & width 64 & 0.7M & 0.068 $\pm$ 0.005 & 0.365 $\pm$ 0.022 \\
         & width 128 & 2.6M & 0.045 $\pm$ 0.003 & 0.222 $\pm$ 0.030 \\
         & width 384 & 21.7M & 0.036 $\pm$ 0.002 & 0.105 $\pm$ 0.012 \\
         & width 512 & 39.3M & 0.034 $\pm$ 0.001 & 0.097 $\pm$ 0.015 \\
        \midrule
        \multirow{3}{*}{Training corpus} & 10\% of the instances & 9.8M & 0.038 $\pm$ 0.003 & 0.102 $\pm$ 0.011 \\
         & 25\% of the instances & 9.8M & 0.037 $\pm$ 0.002 & 0.132 $\pm$ 0.050 \\
         & 50\% of the instances & 9.8M & 0.036 $\pm$ 0.002 & 0.089 $\pm$ 0.009 \\
        \midrule
        \multirow{3}{*}{Design choices} & no capacity perturbation & 9.8M & 0.047 $\pm$ 0.007 & 0.165 $\pm$ 0.069 \\
         & no transpose augmentation & 9.8M & 0.036 $\pm$ 0.002 & 0.142 $\pm$ 0.083 \\
         & plain cross-entropy & 9.8M & 0.040 $\pm$ 0.003 & 0.107 $\pm$ 0.012 \\
        \bottomrule
    \end{tabular}
\end{table}

\newpage
\clearpage
\makeatletter
\providecommand{\appEdir}{figures_cameraready/appendix_E}

\section{Full results on 1D distributions}
\label{app:full-1d}

\providecommand{\appEtocline}[2]{\noindent\hspace*{1em}\makebox[2.2em][l]{\ref{#1}}#2\ \dotfill\ \pageref{#1}\par}
\noindent\textbf{Contents}\par\smallskip
\appEtocline{app:full-1d-euclidean}{Cost type: $L_2$}
\appEtocline{app:full-1d-sqrt-euclidean}{Cost type: $L_2^{1/2}$}
\appEtocline{app:full-1d-sqeuclidean}{Cost type: $L_2^2$}
\appEtocline{app:full-1d-cubic-euclidean}{Cost type: $L_2^3$}
\appEtocline{app:full-1d-canberra}{Cost type: Canberra}
\appEtocline{app:full-1d-gaussian}{Cost type: Gaussian}
\appEtocline{app:full-1d-all}{All six cost types}
\medskip

This appendix reports every method on every cost at $n = m = 30$, on the two sets of Fig.~3:
\begin{itemize}
    \item \textbf{Test set:} 1K random pairs per cost, marginals uniform on the simplex (training distribution).
    \item \textbf{OOD set:} the 144 ordered pairs of the 12 named distributions.
\end{itemize}
As in Fig.~3, each method is scored against the solution of the problem it solves. ToR-OT and NeuralOT (exact OT) are
compared with the optimal plan closest to their prediction, found by a linear program over the optimal face (POT's plan
whenever the optimum is unique). Meta OT and UNOT (entropic OT) are compared with the converged Sinkhorn plan at their
$\varepsilon$, given in each subsection. The metrics, for a prediction $\hat\pi$ and reference $\pi$:
\begin{itemize}
    \item \textbf{MAPE:} $\sum |\hat\pi_{ij} - \pi_{ij}| / \sum \pi_{ij}$ over the union of the two supports (the main metric).
    \item \textbf{MAE:} mean of $|\hat\pi_{ij} - \pi_{ij}|$ over the entries where the two plans differ. An entropic plan
          differs from its reference by small amounts on almost every entry, which lowers its MAE.
    \item \textbf{Hellinger:} $\tfrac{1}{\sqrt 2} \lVert \sqrt{\hat\pi} - \sqrt{\pi} \rVert_2$.
    \item \textbf{TV:} $\tfrac12 \sum |\hat\pi_{ij} - \pi_{ij}|$.
    \item \textbf{$|$OT cost diff.$|$:} $|\langle C, \hat\pi \rangle - \langle C, \pi \rangle|$.
\end{itemize}
The box plots use training seed 0 of the learned baselines; the numbers in the text are the means of Fig.~3 (three
seeds for the baselines). Each subsection has a summary figure and the plans of the 144 OOD pairs for POT and each
method. In these plan grids, the panel in row $i$ and column $j$ is the $30 \times 30$ plan from source distribution
$i$ to target distribution $j$, with its own color scale from 0 to its largest entry; Meta OT and UNOT solve entropic
OT, so their plans are blurred by design. Sec.~\ref{app:full-1d-all} compares all six costs.

\begin{figure}[!ht]
    \centering
    \includegraphics[width=\linewidth]{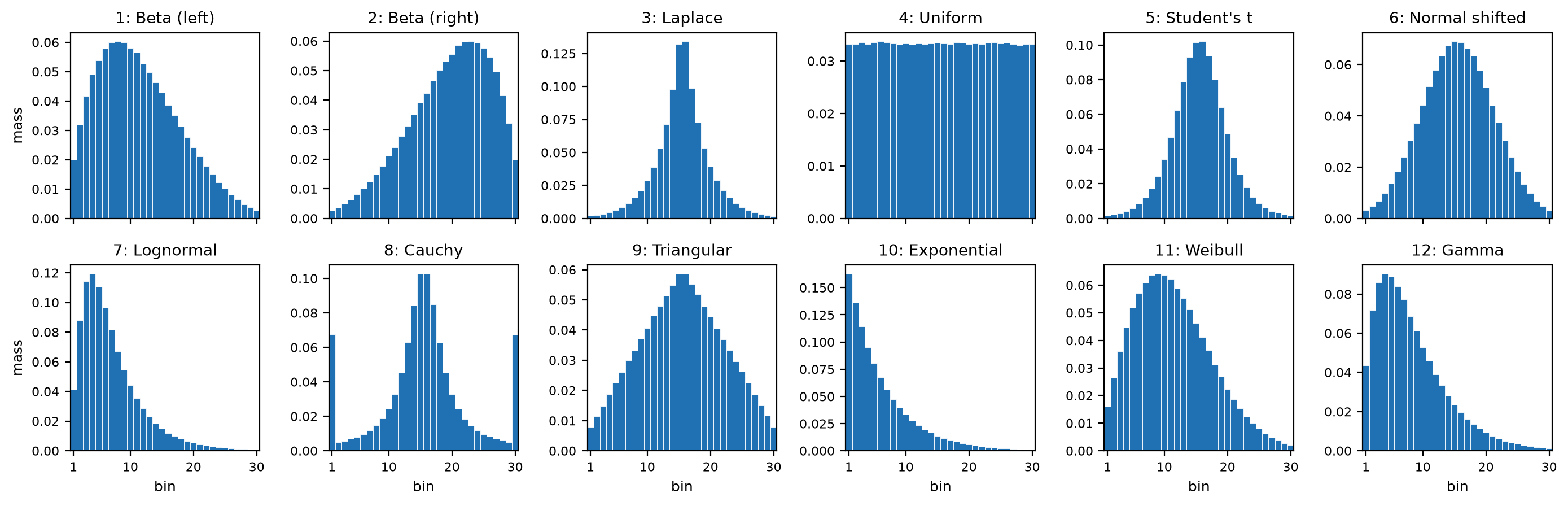}
    \caption{\footnotesize{The 12 distributions of the 1D OOD set, numbered as in the improvement heatmaps
    (Fig.~\ref{fig:appE-improvement}).}}
    \label{fig:appE-distributions}
\end{figure}

\subsection{Cost type: $L_2$}
\label{app:full-1d-euclidean}

\begin{figure}[!ht]
    \centering
    \includegraphics[width=\linewidth]{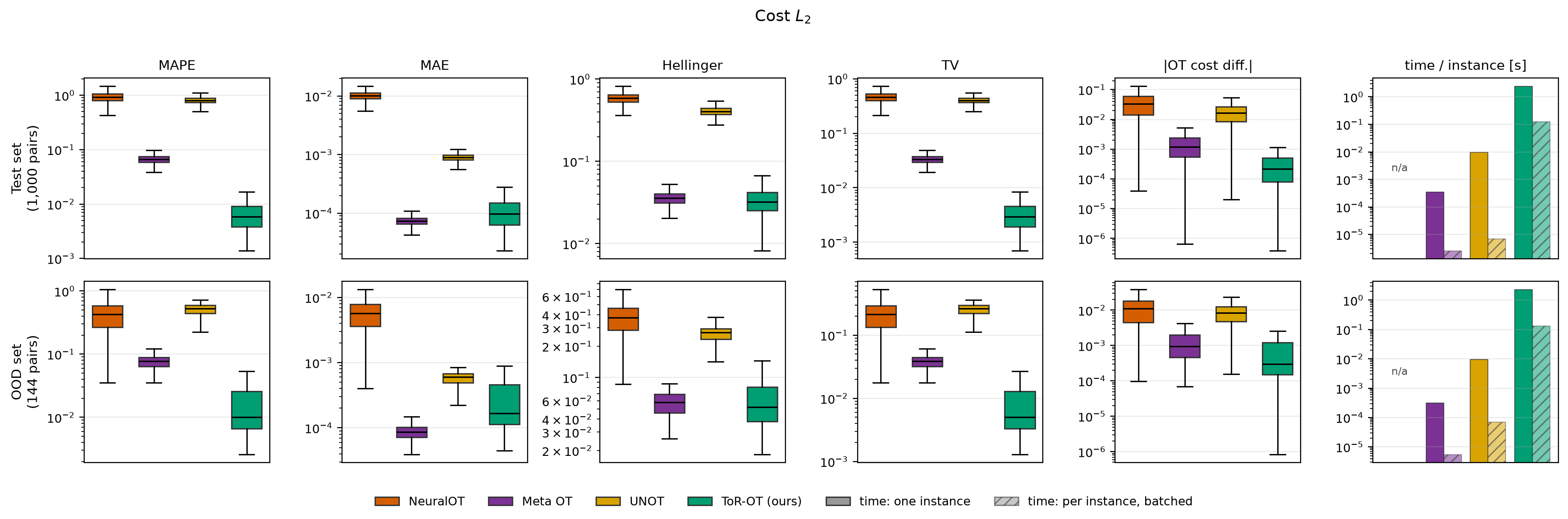}
    \caption{\footnotesize{Cost $L_2$: test set (top, 1{,}000 pairs) and OOD set (bottom, 144 pairs). Five plan metrics, each method against the solution of its own problem (box: quartiles; line: median; whiskers: 1.5 IQR; outliers not shown; log scale), and time per instance on one A100 GPU (solid: one instance; hatched: per instance, batched). NeuralOT optimizes each instance separately and its runs shared GPUs, so its times are not comparable (n/a).}}
    \label{fig:appE-summary-euclidean}
\end{figure}

\begin{figure}[!ht]
    \centering
    \includegraphics[width=0.69\linewidth]{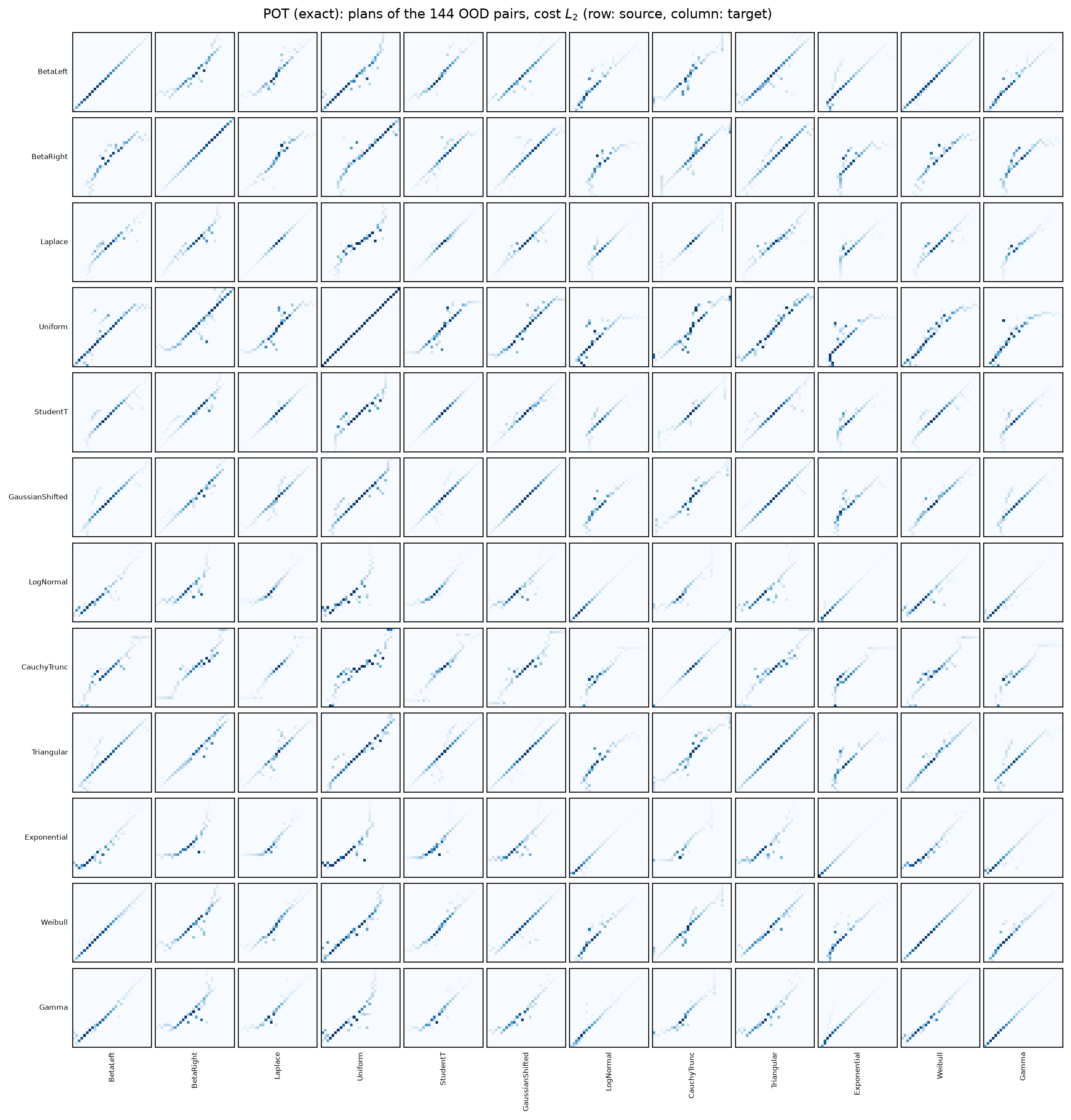}
    \caption{POT (exact): plans of the 144 OOD pairs, cost $L_2$.}
    \label{fig:appE-pairs-euclidean-exact}
\end{figure}

Cost $c(x, y) = |x - y|$; Meta OT uses $\varepsilon = 0.01$ and UNOT $\varepsilon = 0.01$. ToR-OT's mean MAPE is 0.009 on the test set and 0.031 on the OOD set, against 0.072 and 0.078 for Meta OT, the strongest baseline (Fig.~\ref{fig:appE-summary-euclidean}). In 1D the optimal plan for $L_2$ is often not unique: POT's plan (Fig.~\ref{fig:appE-pairs-euclidean-exact}) is one of several, and ToR-OT and NeuralOT are scored against the optimal plan closest to their prediction.

\begin{figure}[p]
    \centering
    \includegraphics[width=0.69\linewidth]{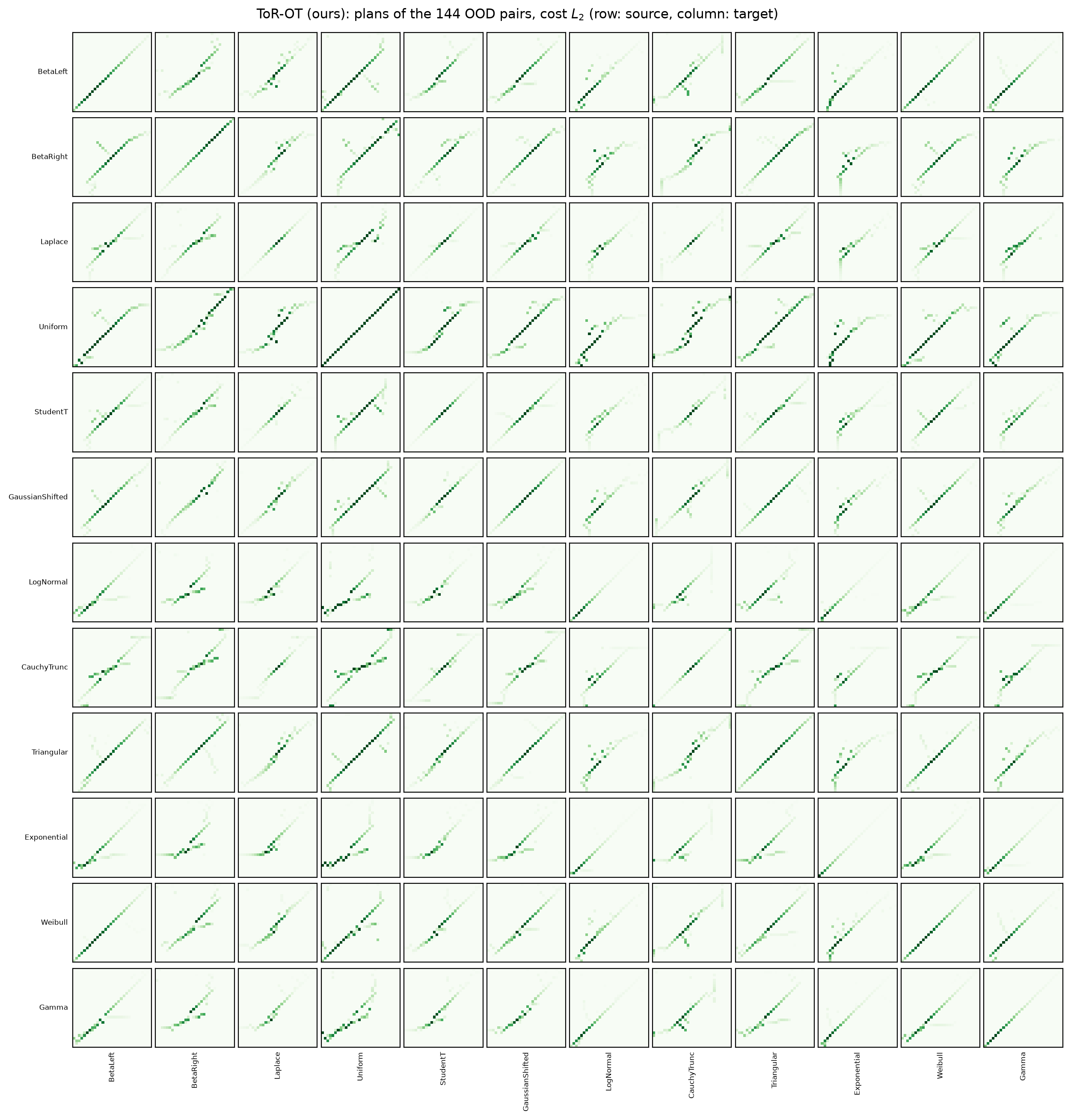}
    \caption{ToR-OT (ours): plans of the 144 OOD pairs, cost $L_2$.}
    \label{fig:appE-pairs-euclidean-v2row_lastrow}
    \vspace{0.5em}
    \includegraphics[width=0.69\linewidth]{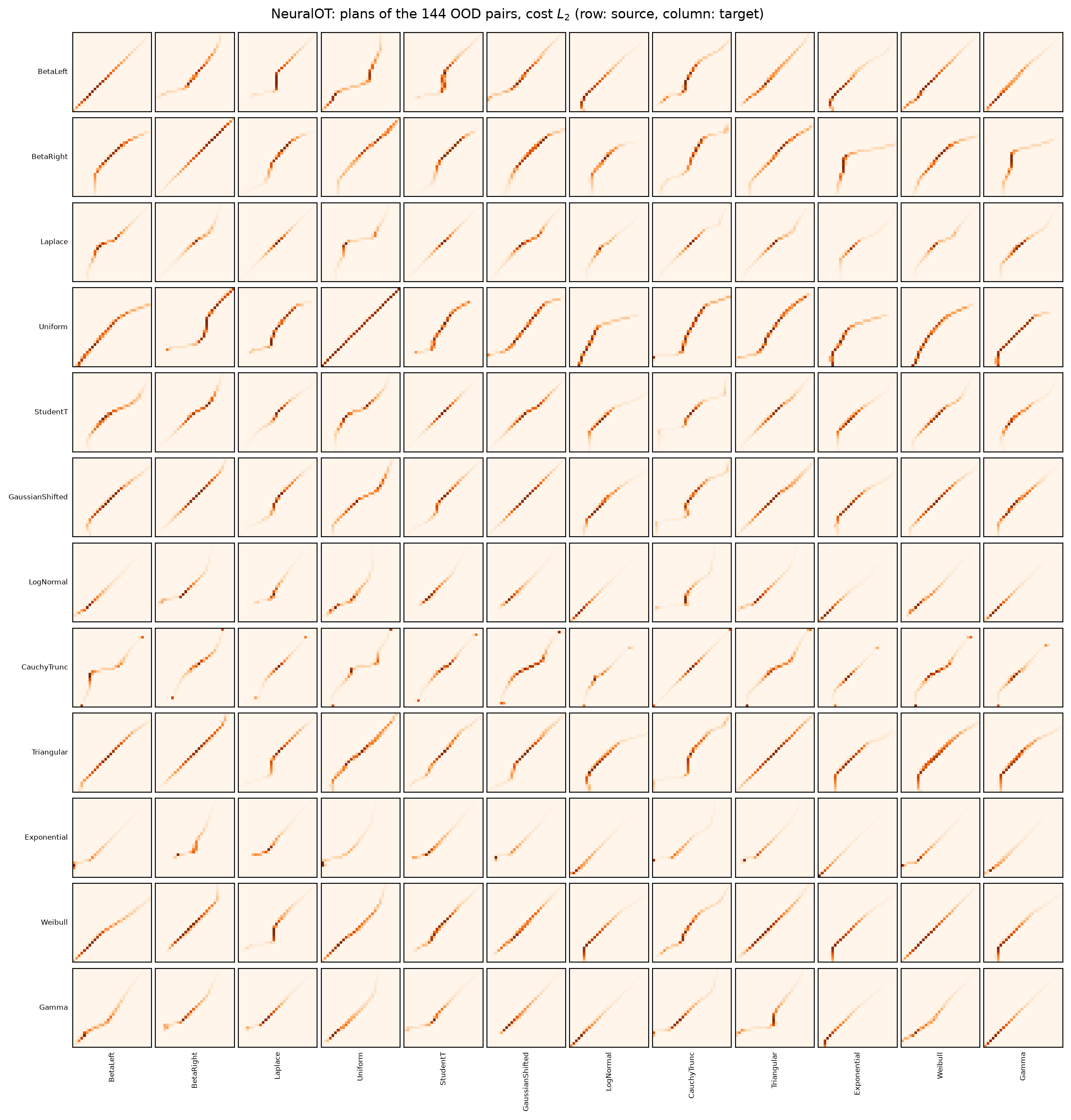}
    \caption{NeuralOT: plans of the 144 OOD pairs, cost $L_2$.}
    \label{fig:appE-pairs-euclidean-neuralot}
\end{figure}

\begin{figure}[p]
    \centering
    \includegraphics[width=0.69\linewidth]{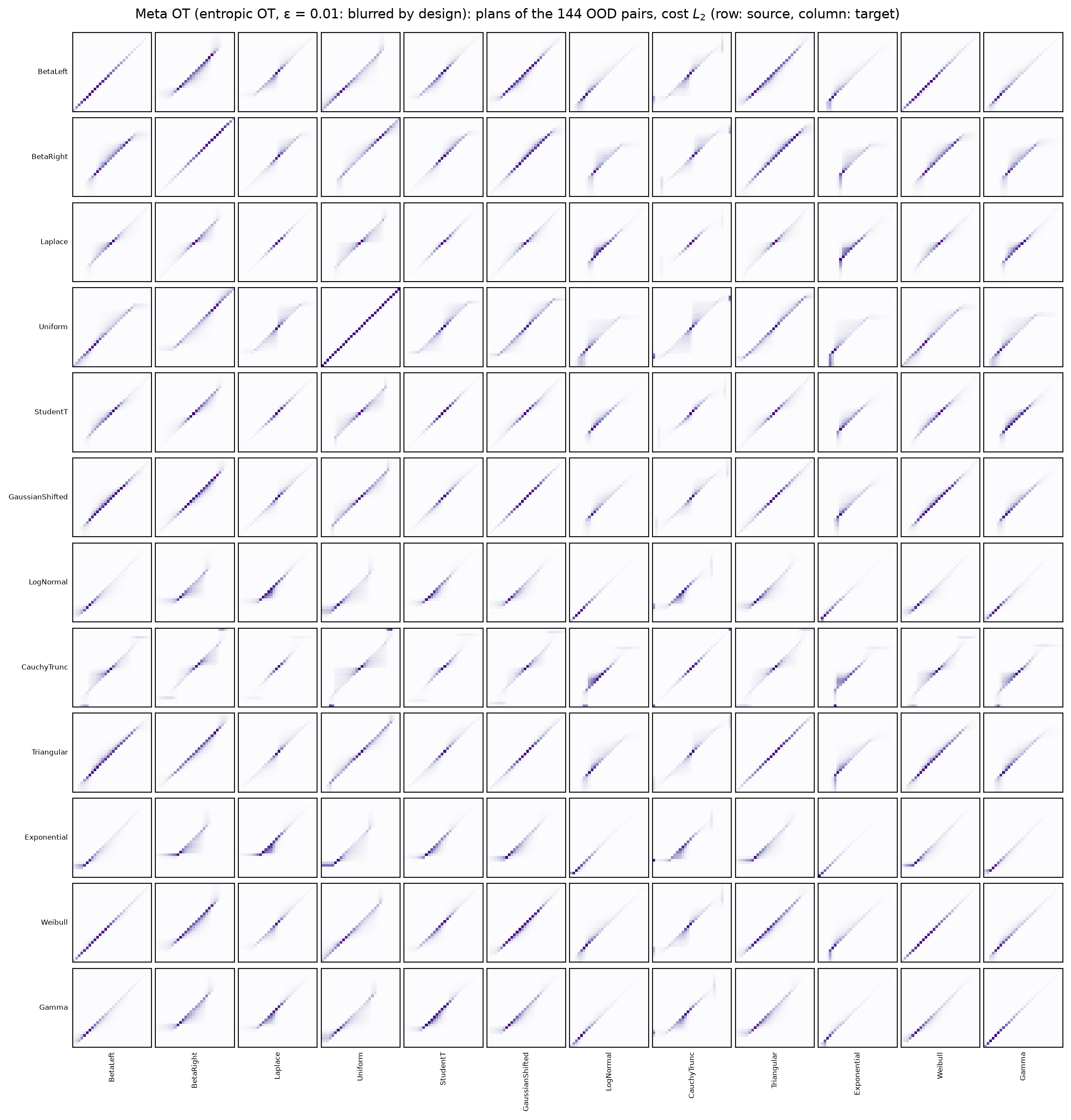}
    \caption{Meta OT (entropic, $\varepsilon = 0.01$): plans of the 144 OOD pairs, cost $L_2$.}
    \label{fig:appE-pairs-euclidean-metaot}
    \vspace{0.5em}
    \includegraphics[width=0.69\linewidth]{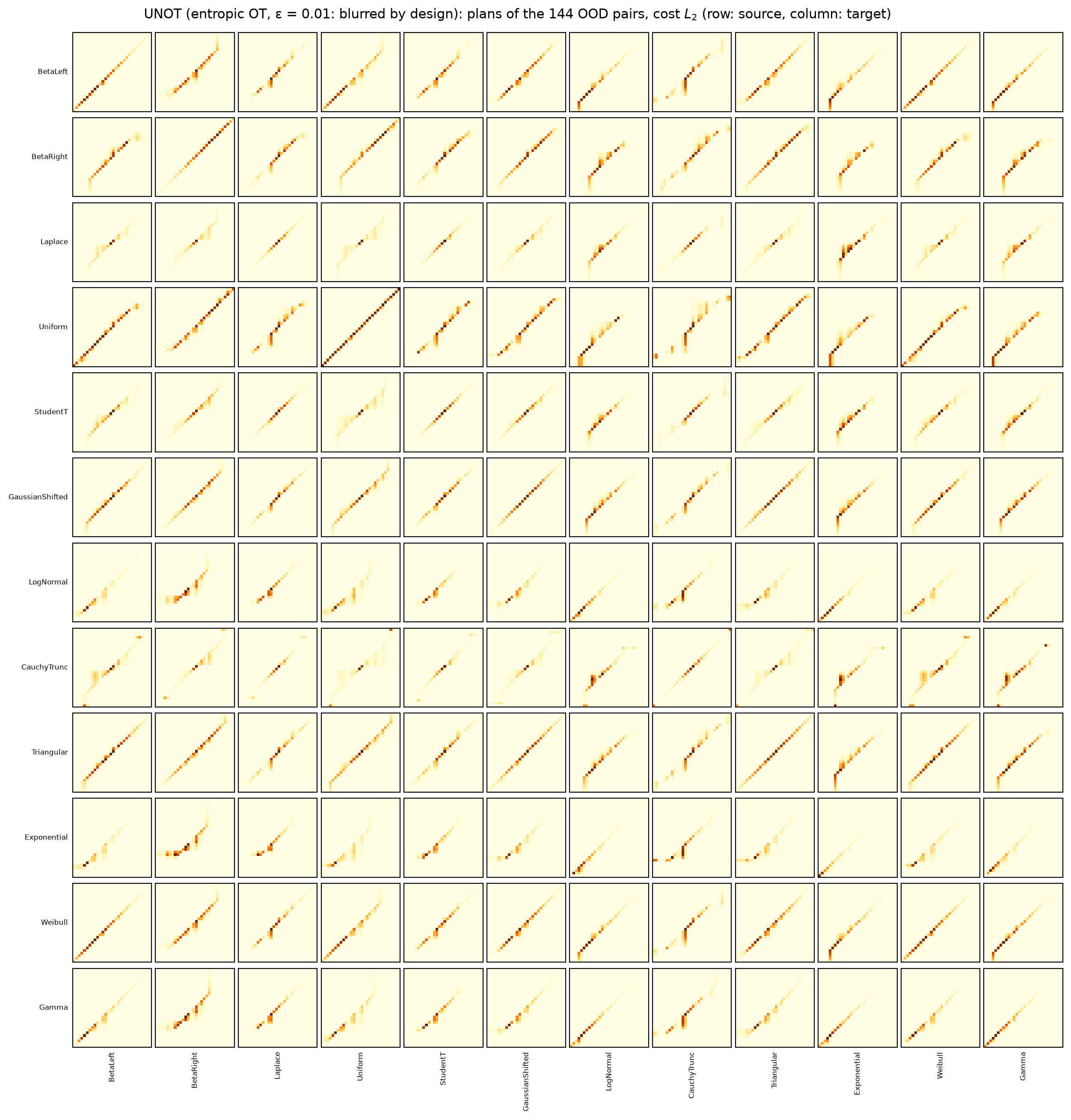}
    \caption{UNOT (entropic, $\varepsilon = 0.01$): plans of the 144 OOD pairs, cost $L_2$.}
    \label{fig:appE-pairs-euclidean-unot}
\end{figure}

\clearpage
\subsection{Cost type: $L_2^{1/2}$}
\label{app:full-1d-sqrt-euclidean}

\begin{figure}[!ht]
    \centering
    \includegraphics[width=\linewidth]{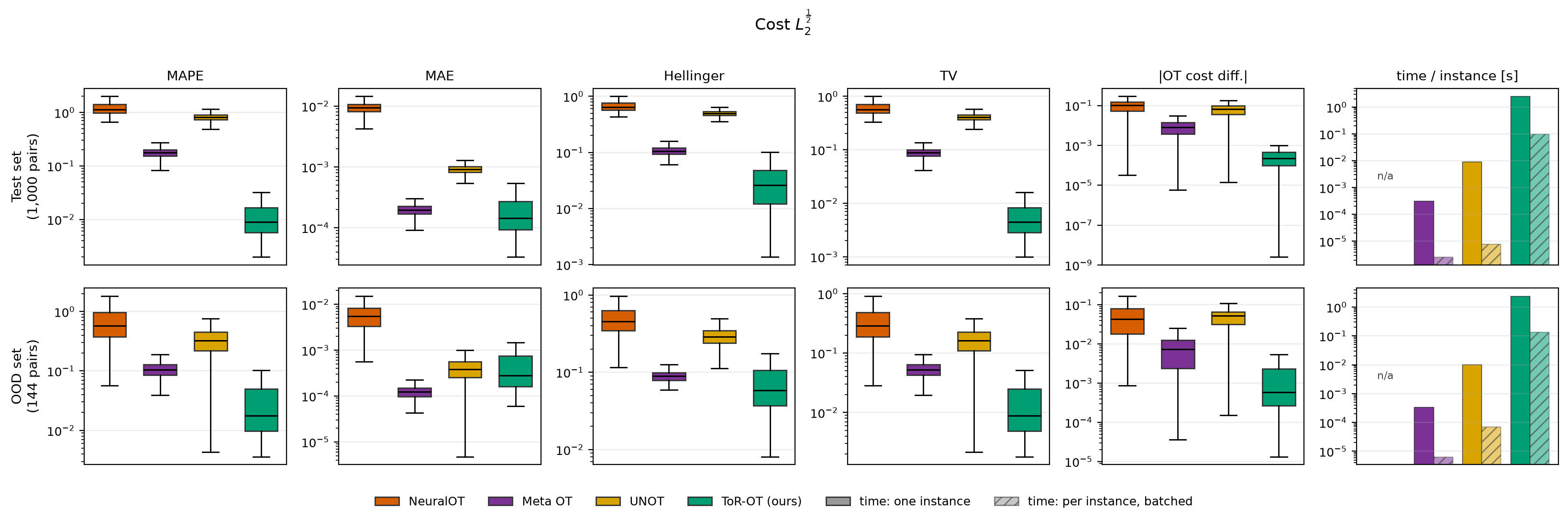}
    \caption{Cost $L_2^{1/2}$: test set (top, 1{,}000 pairs) and OOD set (bottom, 144 pairs). Five plan metrics, each method against the solution of its own problem (box: quartiles; line: median; whiskers: 1.5 IQR; outliers not shown; log scale), and time per instance on one A100 GPU (solid: one instance; hatched: per instance, batched). NeuralOT optimizes each instance separately and its runs shared GPUs, so its times are not comparable (n/a).}
    \label{fig:appE-summary-sqrt-euclidean}
\end{figure}

Cost $c(x, y) = |x - y|^{1/2}$ (concave); Meta OT uses $\varepsilon = 0.01$ and UNOT $\varepsilon = 0.01$. ToR-OT's mean MAPE is 0.016 on the test set and 0.037 on the OOD set, against 0.180 and 0.104 for Meta OT, the strongest baseline (Fig.~\ref{fig:appE-summary-sqrt-euclidean}).

\begin{figure}[!ht]
    \centering
    \includegraphics[width=0.69\linewidth]{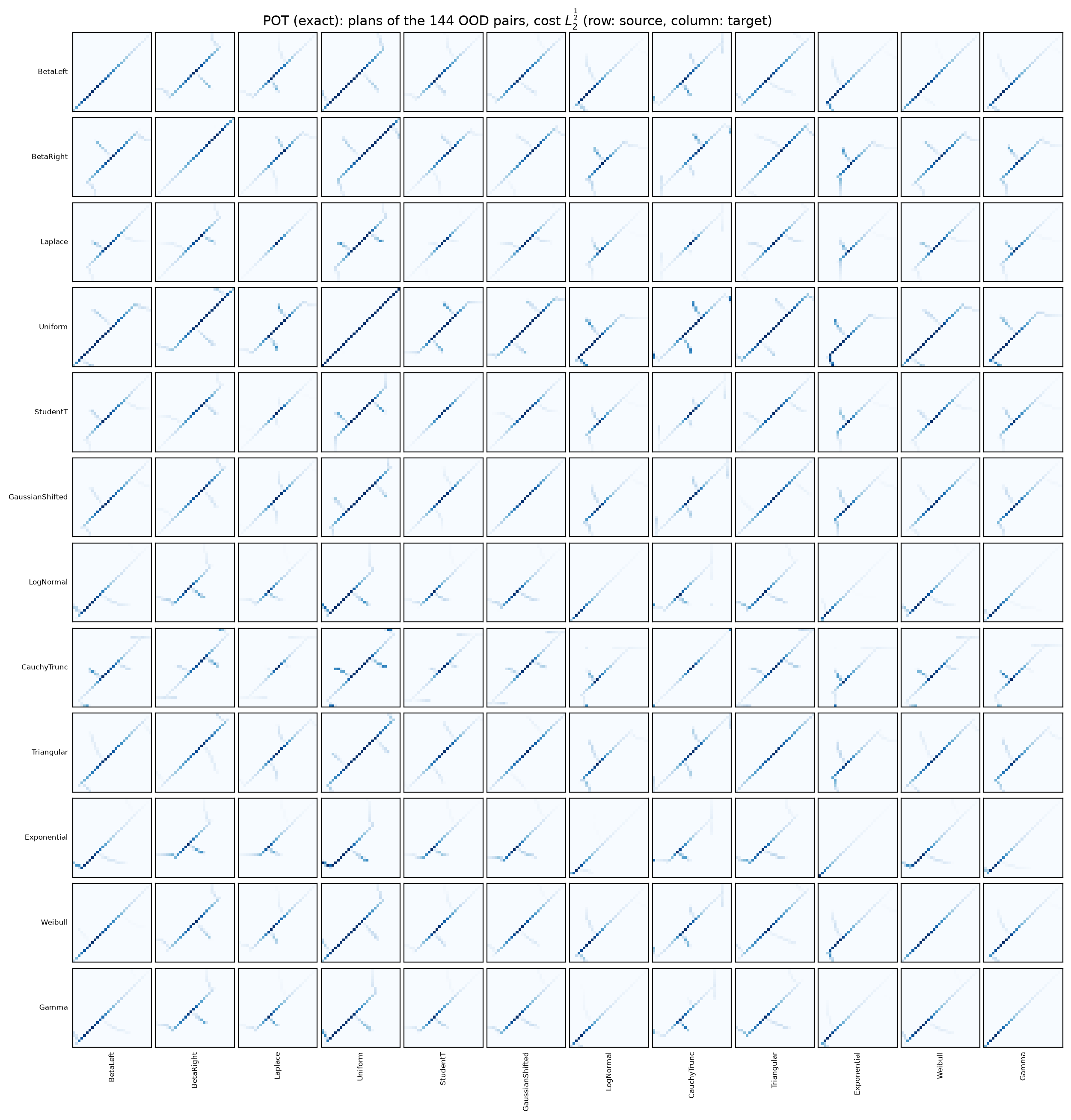}
    \caption{POT (exact): plans of the 144 OOD pairs, cost $L_2^{1/2}$.}
    \label{fig:appE-pairs-sqrt-euclidean-exact}
\end{figure}

\begin{figure}[p]
    \centering
    \includegraphics[width=0.69\linewidth]{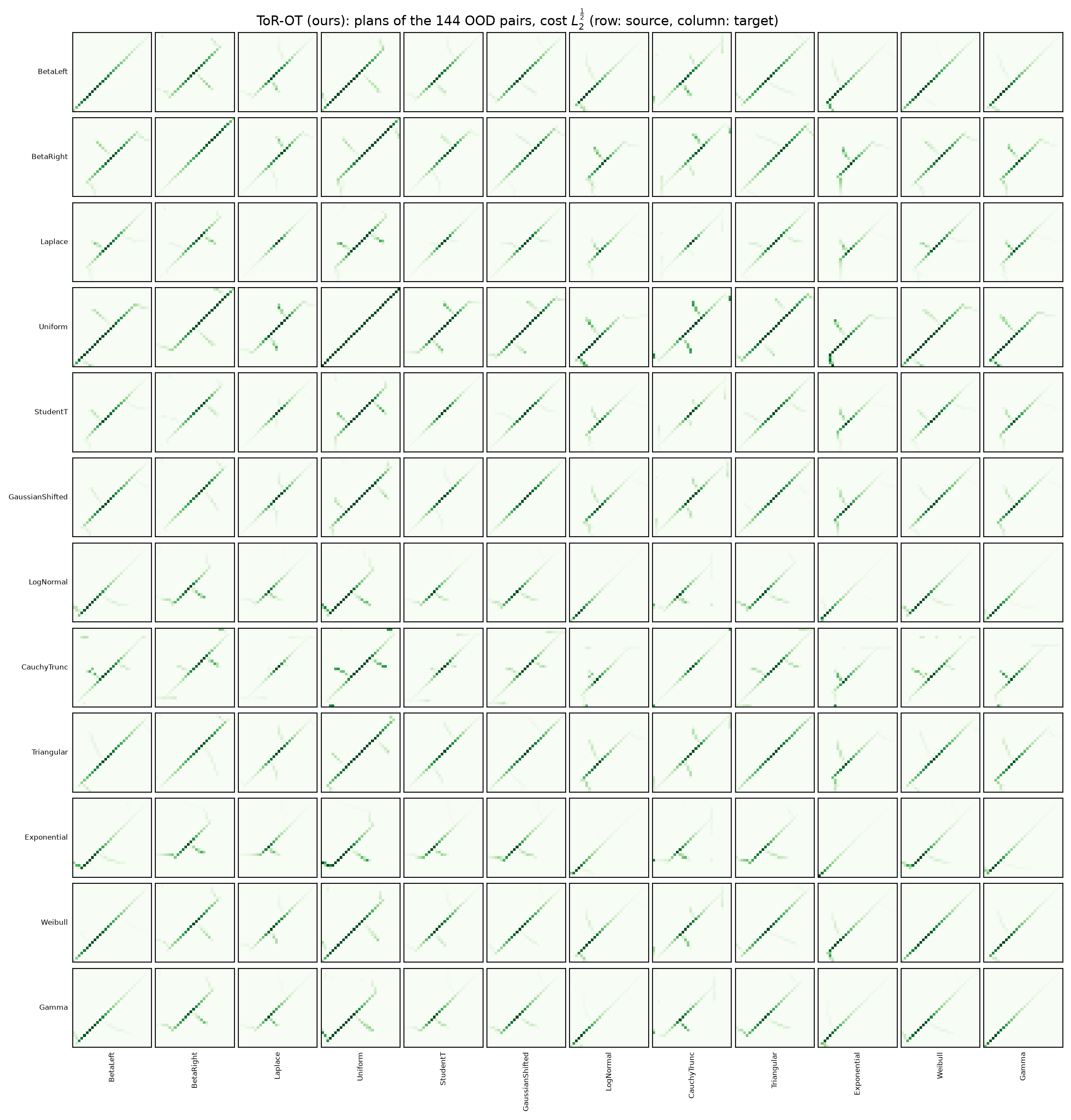}
    \caption{ToR-OT (ours): plans of the 144 OOD pairs, cost $L_2^{1/2}$.}
    \label{fig:appE-pairs-sqrt-euclidean-v2row_lastrow}
    \vspace{0.5em}
    \includegraphics[width=0.69\linewidth]{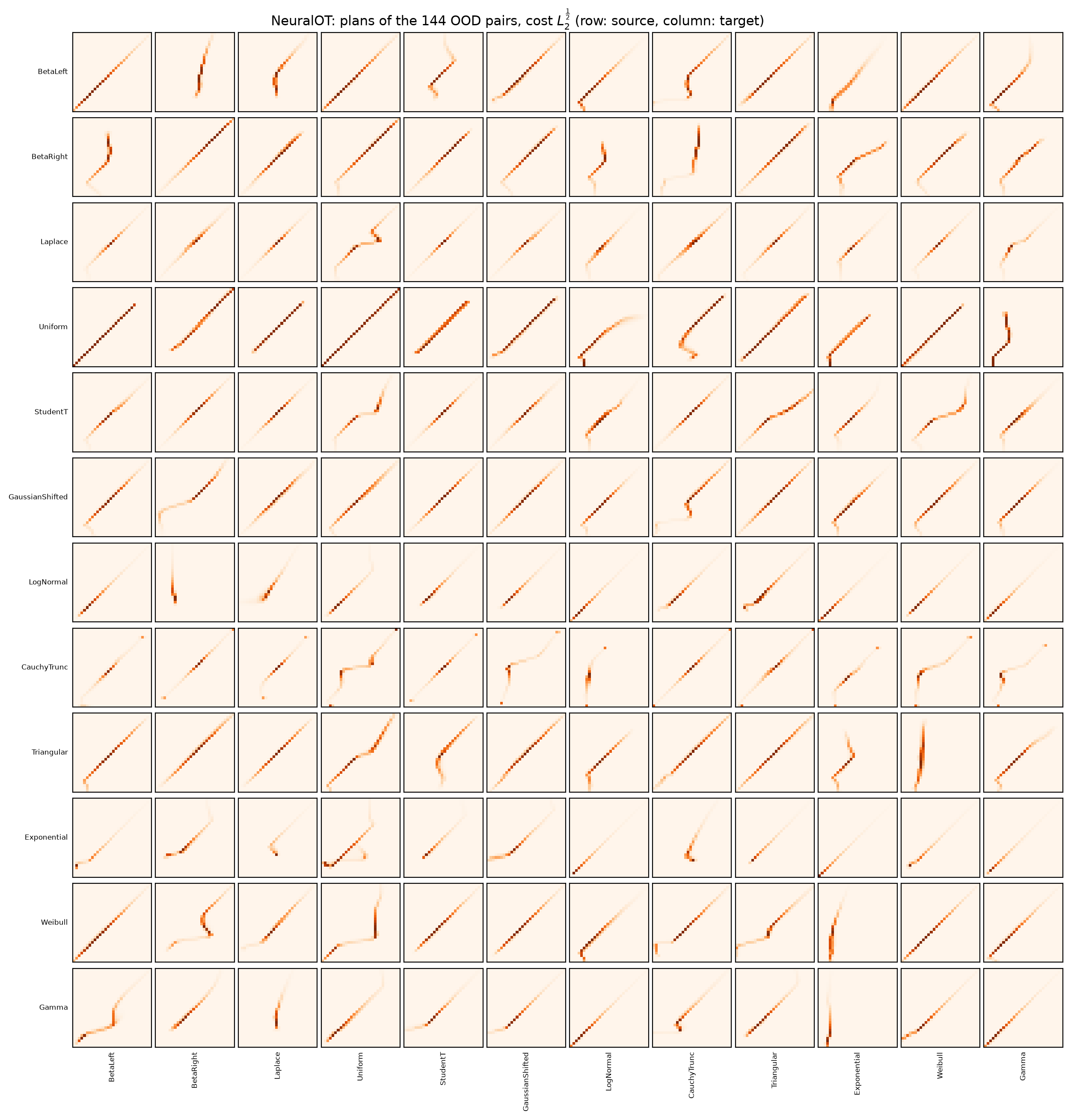}
    \caption{NeuralOT: plans of the 144 OOD pairs, cost $L_2^{1/2}$.}
    \label{fig:appE-pairs-sqrt-euclidean-neuralot}
\end{figure}

\begin{figure}[p]
    \centering
    \includegraphics[width=0.69\linewidth]{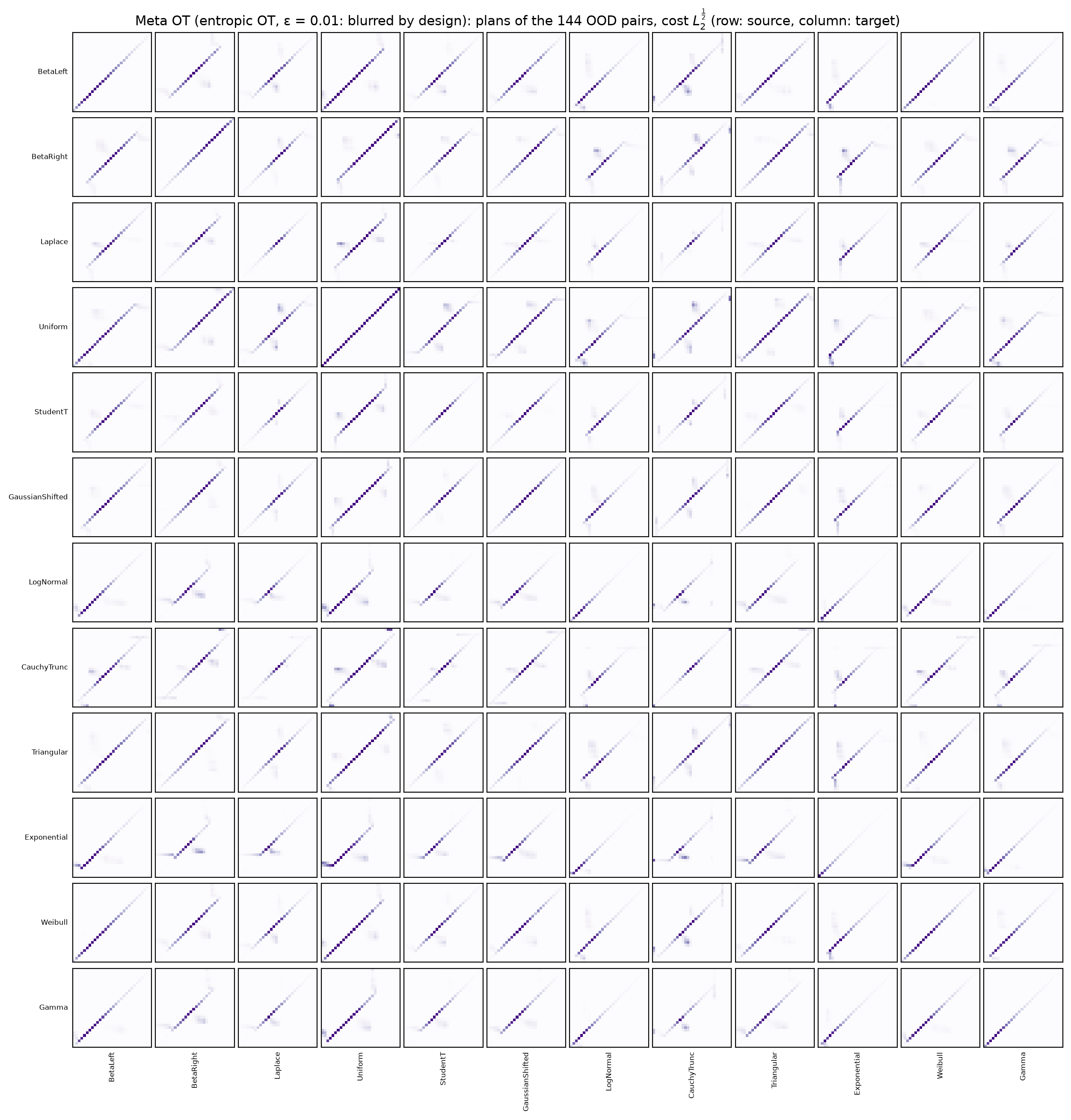}
    \caption{Meta OT (entropic, $\varepsilon = 0.01$): plans of the 144 OOD pairs, cost $L_2^{1/2}$.}
    \label{fig:appE-pairs-sqrt-euclidean-metaot}
    \vspace{0.5em}
    \includegraphics[width=0.69\linewidth]{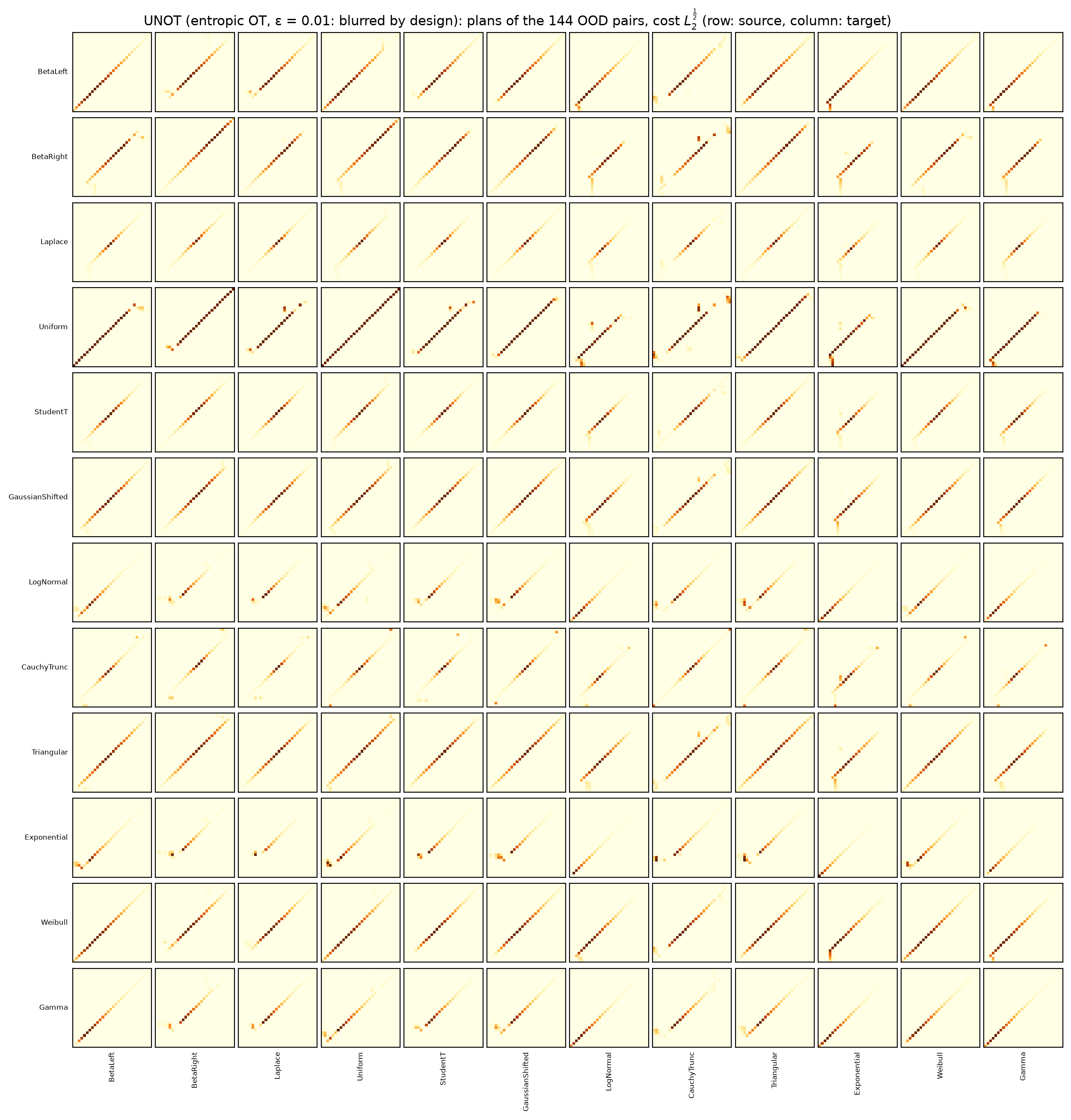}
    \caption{UNOT (entropic, $\varepsilon = 0.01$): plans of the 144 OOD pairs, cost $L_2^{1/2}$.}
    \label{fig:appE-pairs-sqrt-euclidean-unot}
\end{figure}

\clearpage
\subsection{Cost type: $L_2^2$}
\label{app:full-1d-sqeuclidean}

\begin{figure}[!ht]
    \centering
    \includegraphics[width=\linewidth]{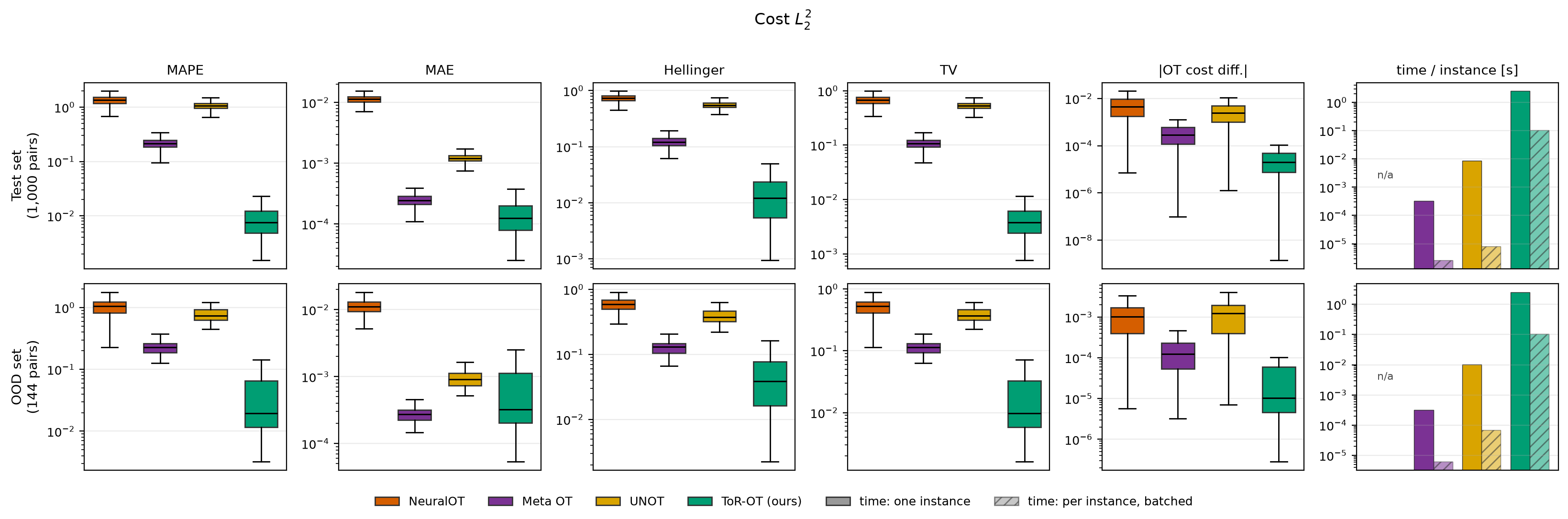}
    \caption{Cost $L_2^2$: test set (top, 1{,}000 pairs) and OOD set (bottom, 144 pairs). Five plan metrics, each method against the solution of its own problem (box: quartiles; line: median; whiskers: 1.5 IQR; outliers not shown; log scale), and time per instance on one A100 GPU (solid: one instance; hatched: per instance, batched). NeuralOT optimizes each instance separately and its runs shared GPUs, so its times are not comparable (n/a).}
    \label{fig:appE-summary-sqeuclidean}
\end{figure}

Cost $c(x, y) = |x - y|^2$; Meta OT uses $\varepsilon = 0.001$ and UNOT $\varepsilon = 0.001$. ToR-OT's mean MAPE is 0.010 on the test set and 0.056 on the OOD set, against 0.221 and 0.229 for Meta OT, the strongest baseline (Fig.~\ref{fig:appE-summary-sqeuclidean}).

\begin{figure}[!ht]
    \centering
    \includegraphics[width=0.69\linewidth]{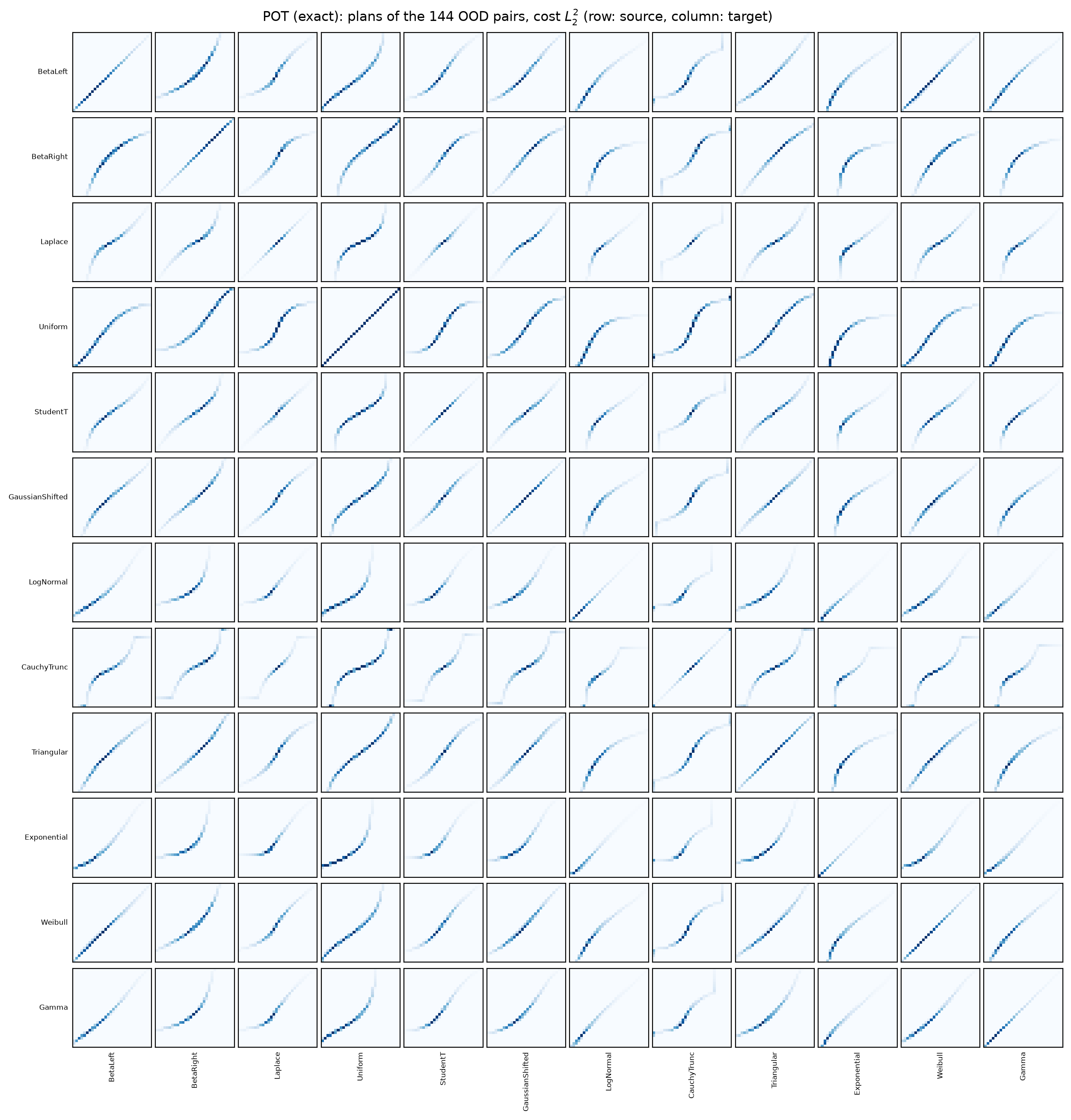}
    \caption{POT (exact): plans of the 144 OOD pairs, cost $L_2^2$.}
    \label{fig:appE-pairs-sqeuclidean-exact}
\end{figure}

\begin{figure}[p]
    \centering
    \includegraphics[width=0.69\linewidth]{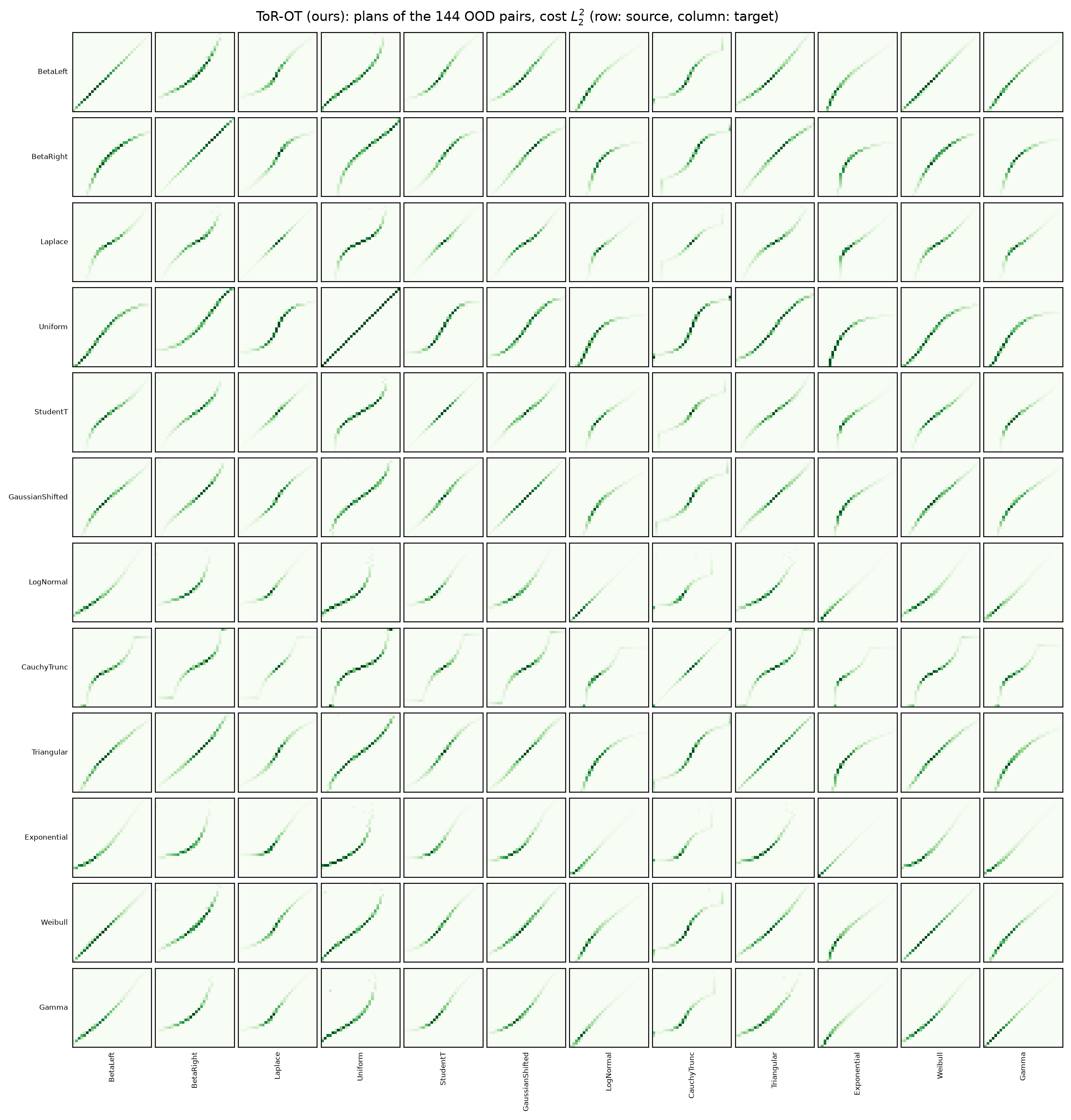}
    \caption{ToR-OT (ours): plans of the 144 OOD pairs, cost $L_2^2$.}
    \label{fig:appE-pairs-sqeuclidean-v2row_lastrow}
    \vspace{0.5em}
    \includegraphics[width=0.69\linewidth]{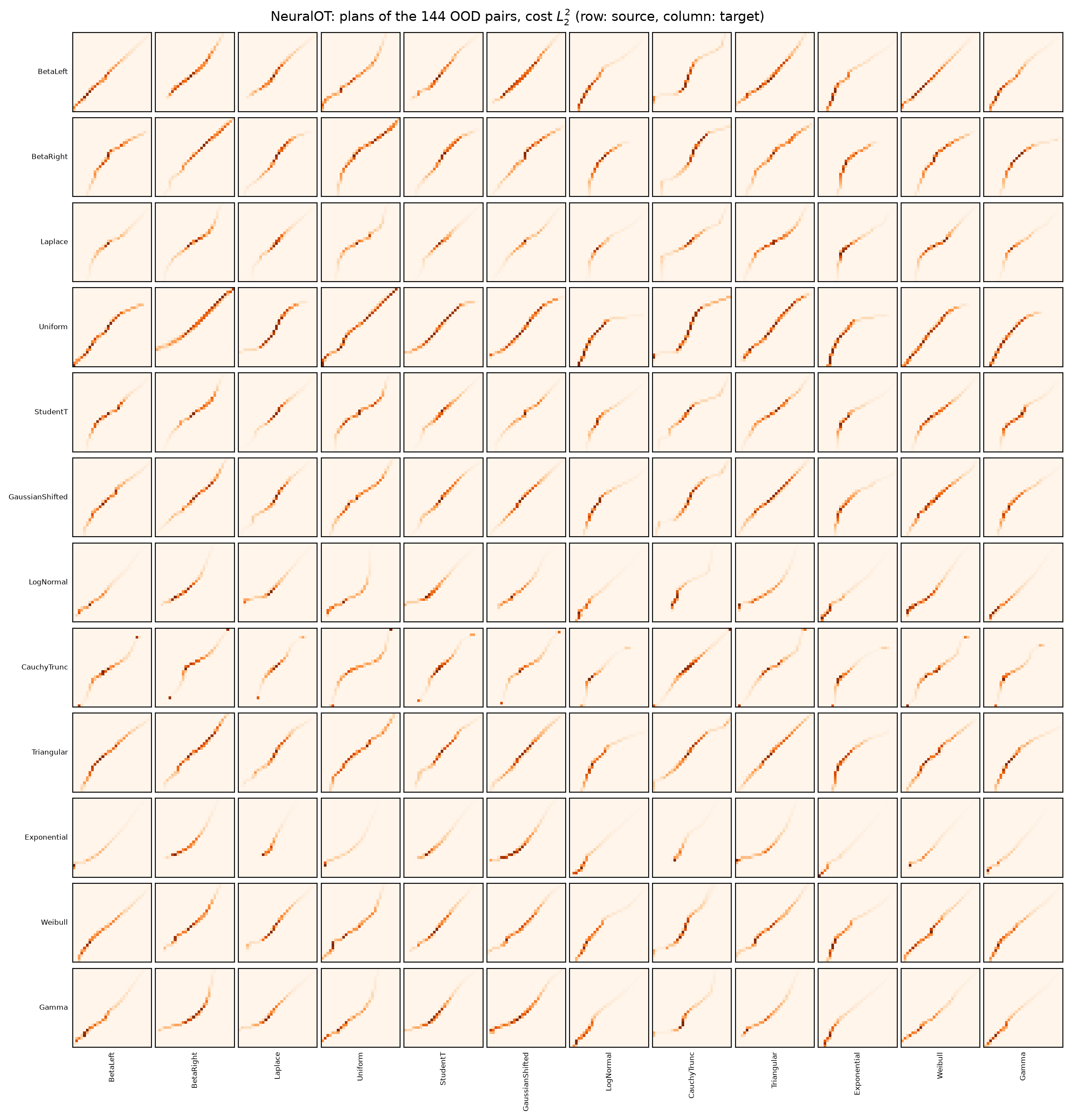}
    \caption{NeuralOT: plans of the 144 OOD pairs, cost $L_2^2$.}
    \label{fig:appE-pairs-sqeuclidean-neuralot}
\end{figure}

\begin{figure}[p]
    \centering
    \includegraphics[width=0.69\linewidth]{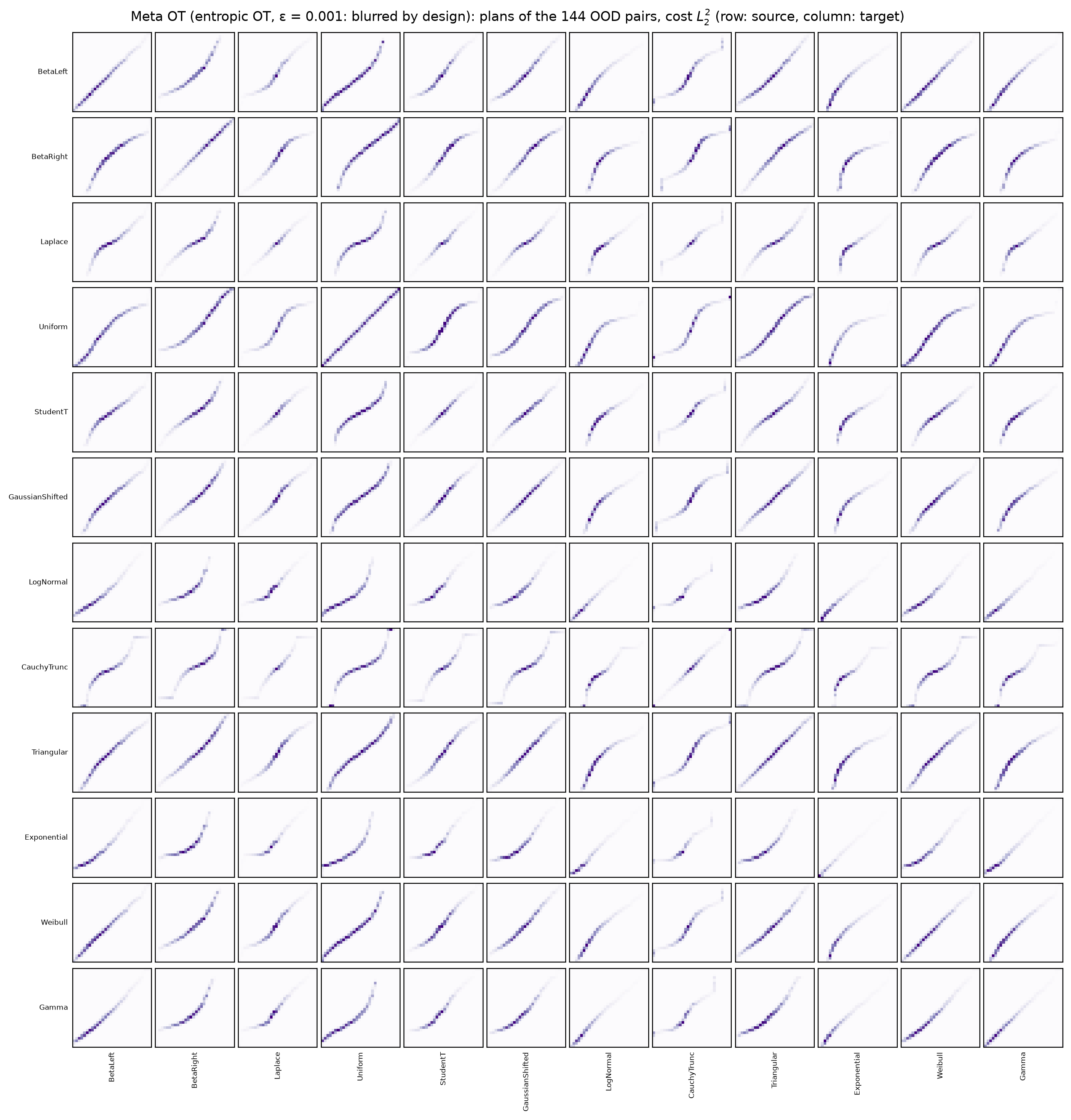}
    \caption{Meta OT (entropic, $\varepsilon = 0.001$): plans of the 144 OOD pairs, cost $L_2^2$.}
    \label{fig:appE-pairs-sqeuclidean-metaot}
    \vspace{0.5em}
    \includegraphics[width=0.69\linewidth]{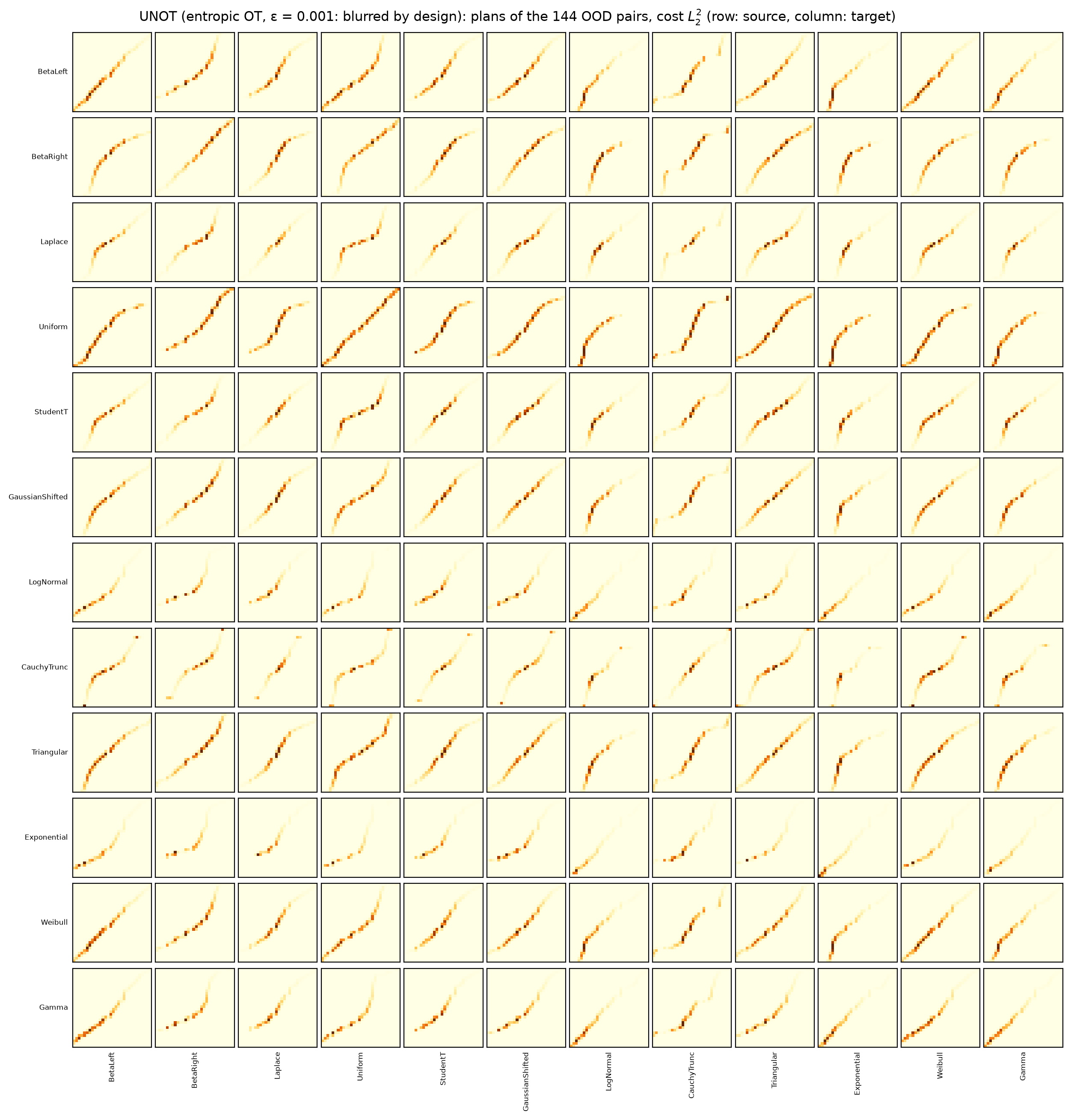}
    \caption{UNOT (entropic, $\varepsilon = 0.001$): plans of the 144 OOD pairs, cost $L_2^2$.}
    \label{fig:appE-pairs-sqeuclidean-unot}
\end{figure}

\clearpage
\subsection{Cost type: $L_2^3$}
\label{app:full-1d-cubic-euclidean}

\begin{figure}[!ht]
    \centering
    \includegraphics[width=\linewidth]{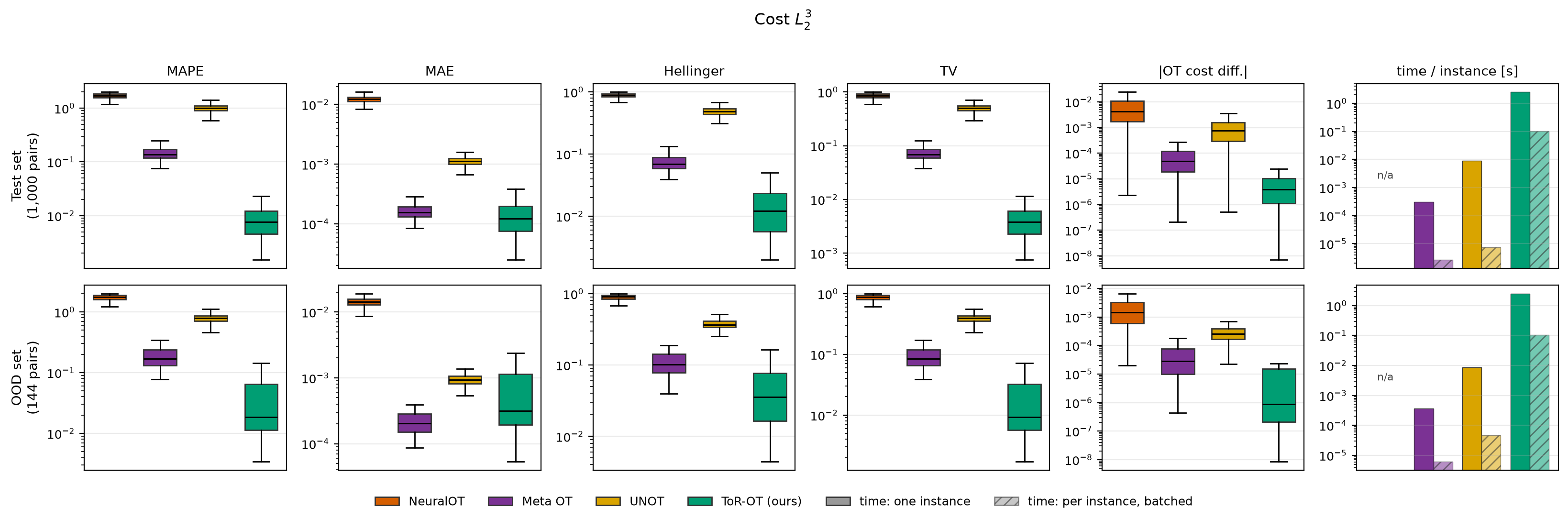}
    \caption{Cost $L_2^3$: test set (top, 1{,}000 pairs) and OOD set (bottom, 144 pairs). Five plan metrics, each method against the solution of its own problem (box: quartiles; line: median; whiskers: 1.5 IQR; outliers not shown; log scale), and time per instance on one A100 GPU (solid: one instance; hatched: per instance, batched). NeuralOT optimizes each instance separately and its runs shared GPUs, so its times are not comparable (n/a).}
    \label{fig:appE-summary-cubic-euclidean}
\end{figure}

Cost $c(x, y) = |x - y|^3$; Meta OT uses $\varepsilon = 0.001$ and UNOT $\varepsilon = 0.001$. ToR-OT's mean MAPE is 0.010 on the test set and 0.057 on the OOD set, against 0.148 and 0.175 for Meta OT, the strongest baseline (Fig.~\ref{fig:appE-summary-cubic-euclidean}).

\begin{figure}[!ht]
    \centering
    \includegraphics[width=0.69\linewidth]{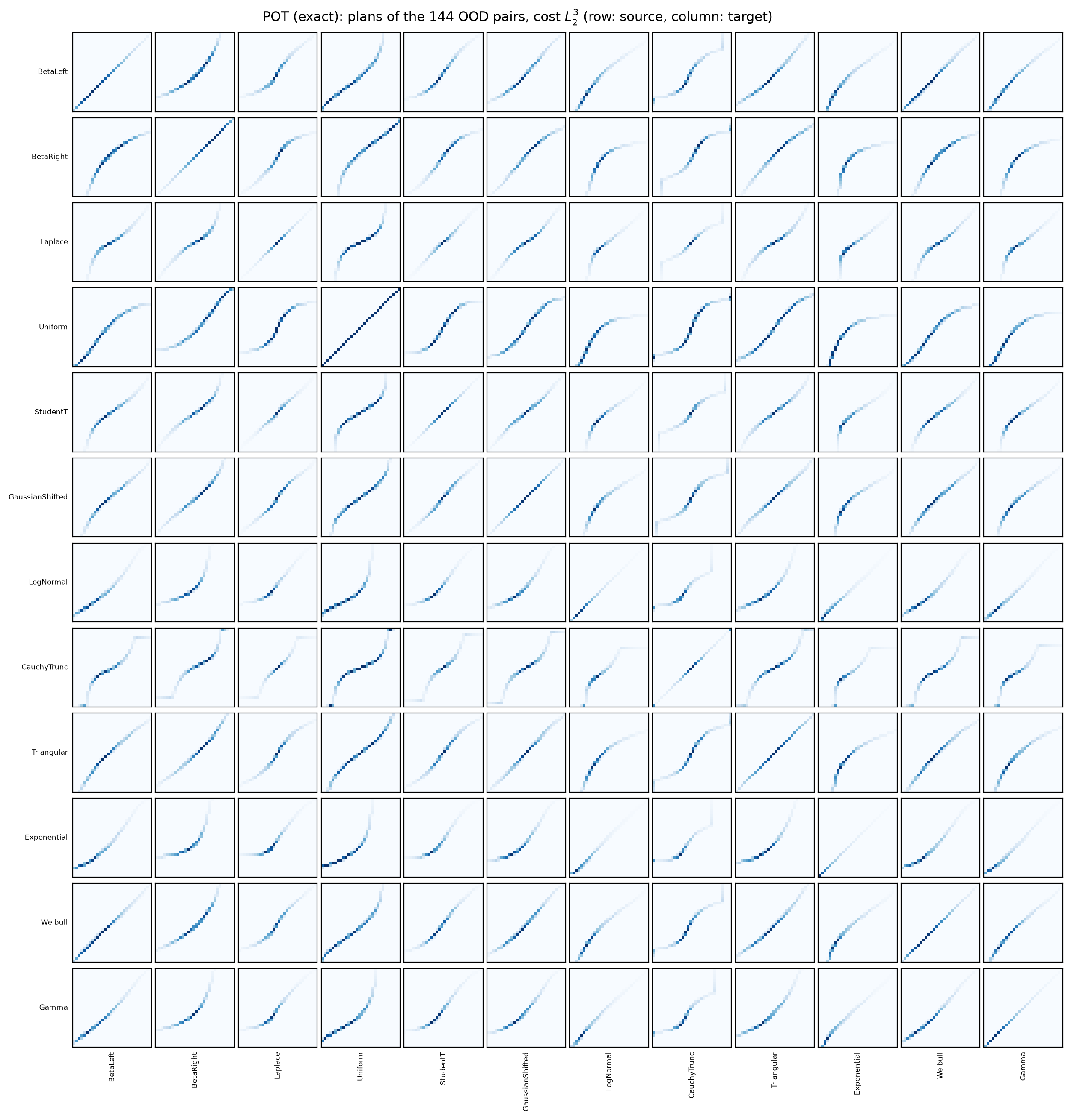}
    \caption{POT (exact): plans of the 144 OOD pairs, cost $L_2^3$.}
    \label{fig:appE-pairs-cubic-euclidean-exact}
\end{figure}

\begin{figure}[p]
    \centering
    \includegraphics[width=0.69\linewidth]{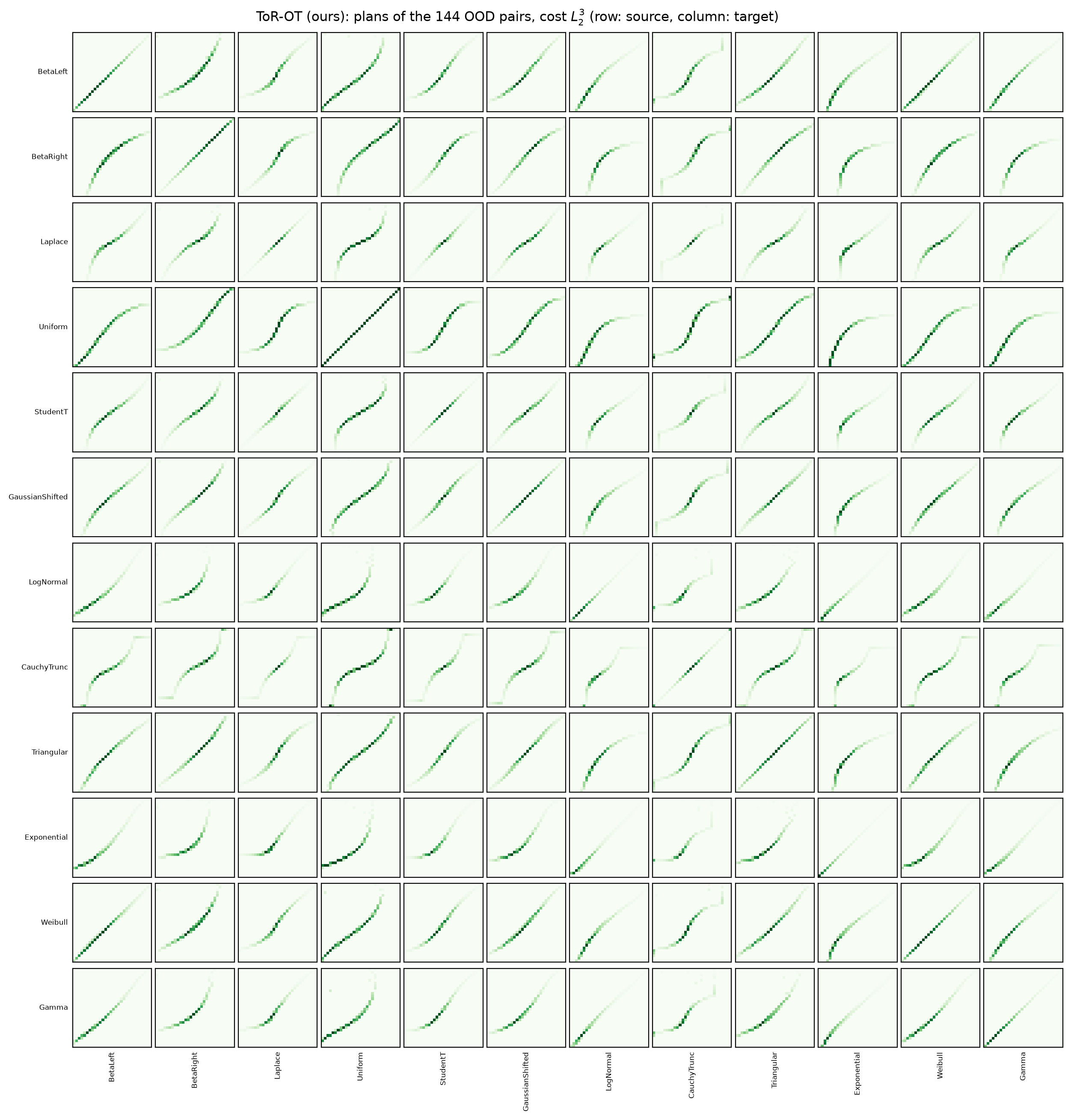}
    \caption{ToR-OT (ours): plans of the 144 OOD pairs, cost $L_2^3$.}
    \label{fig:appE-pairs-cubic-euclidean-v2row_lastrow}
    \vspace{0.5em}
    \includegraphics[width=0.69\linewidth]{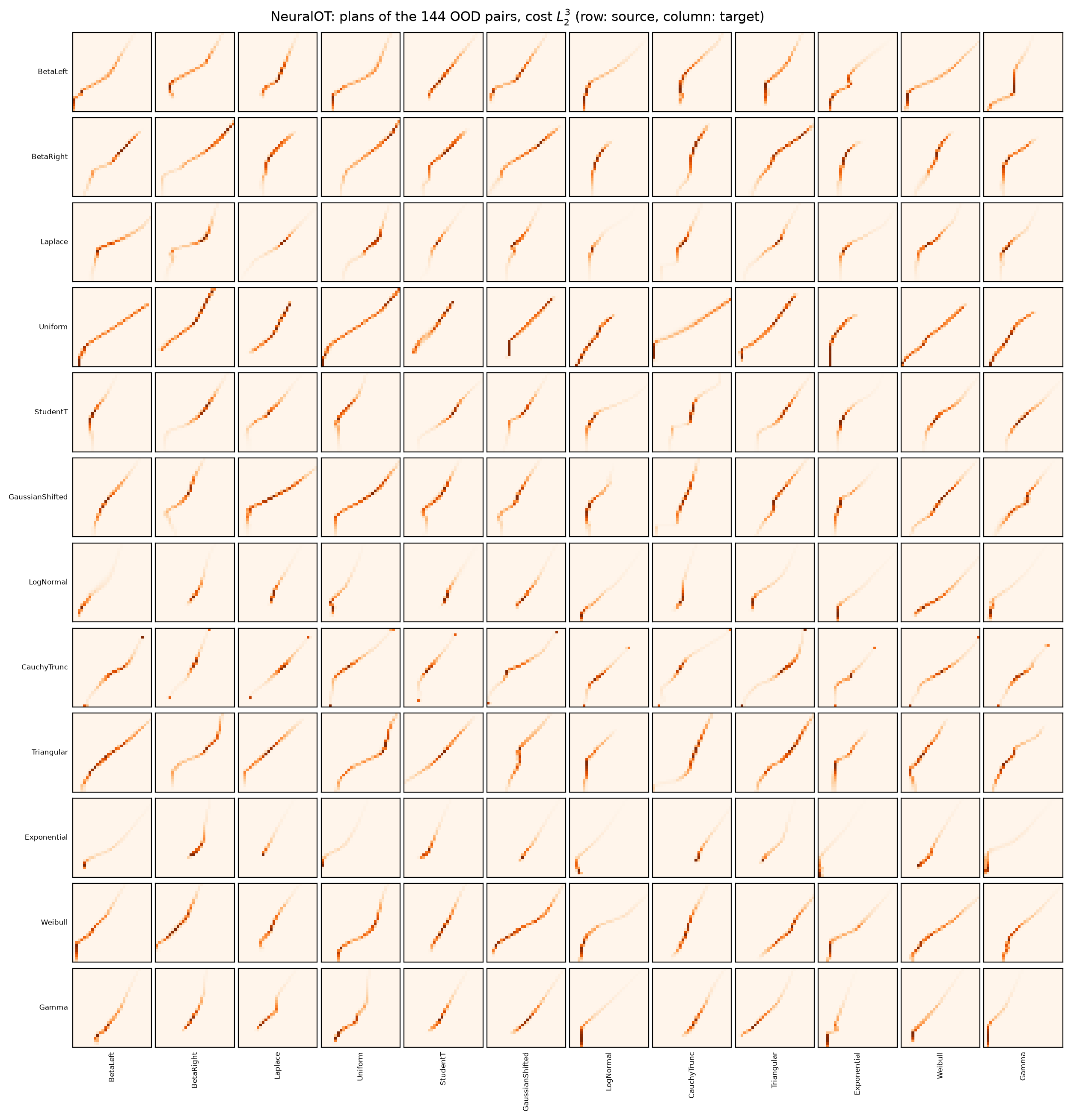}
    \caption{NeuralOT: plans of the 144 OOD pairs, cost $L_2^3$.}
    \label{fig:appE-pairs-cubic-euclidean-neuralot}
\end{figure}

\begin{figure}[p]
    \centering
    \includegraphics[width=0.69\linewidth]{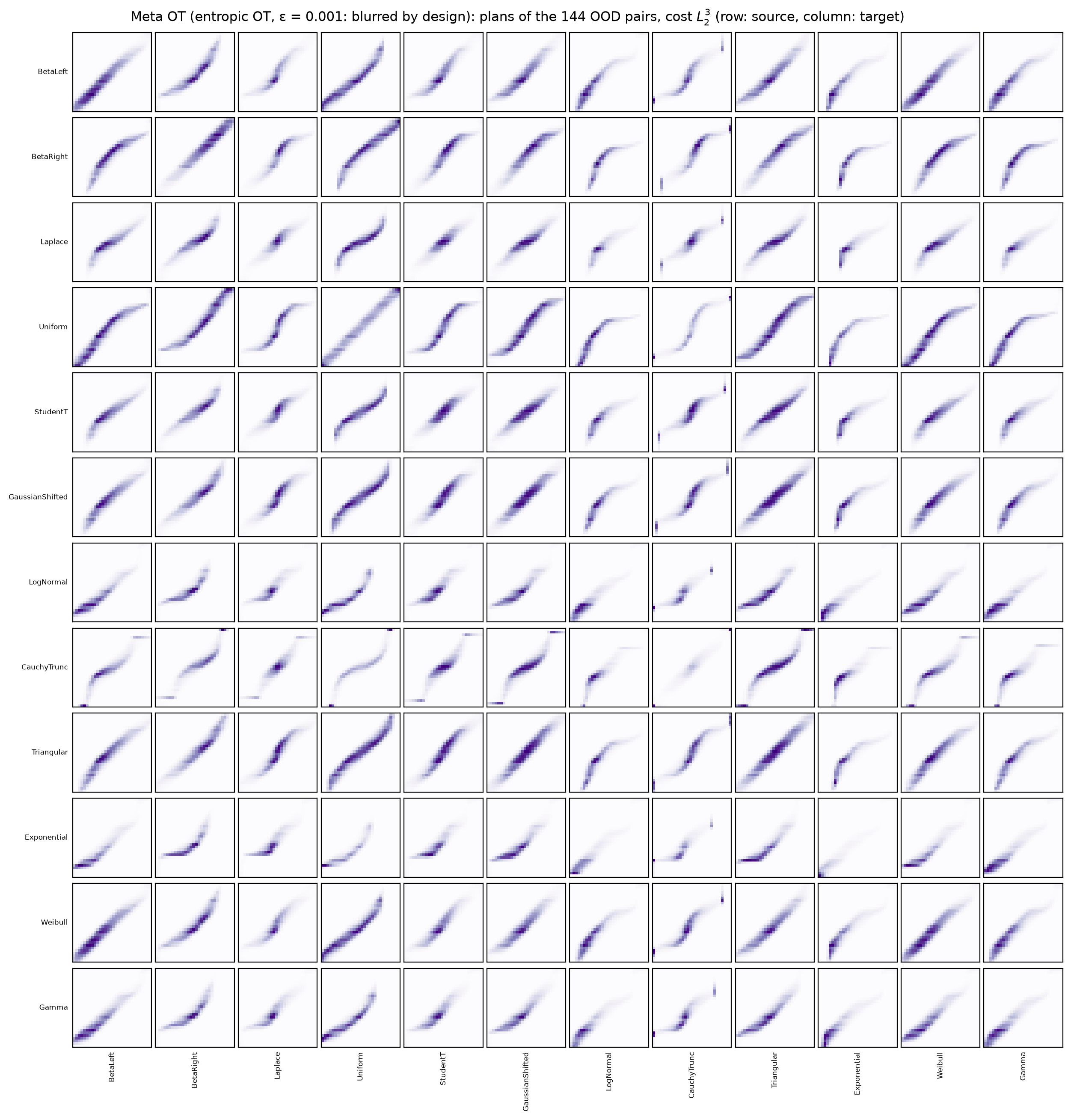}
    \caption{Meta OT (entropic, $\varepsilon = 0.001$): plans of the 144 OOD pairs, cost $L_2^3$.}
    \label{fig:appE-pairs-cubic-euclidean-metaot}
    \vspace{0.5em}
    \includegraphics[width=0.69\linewidth]{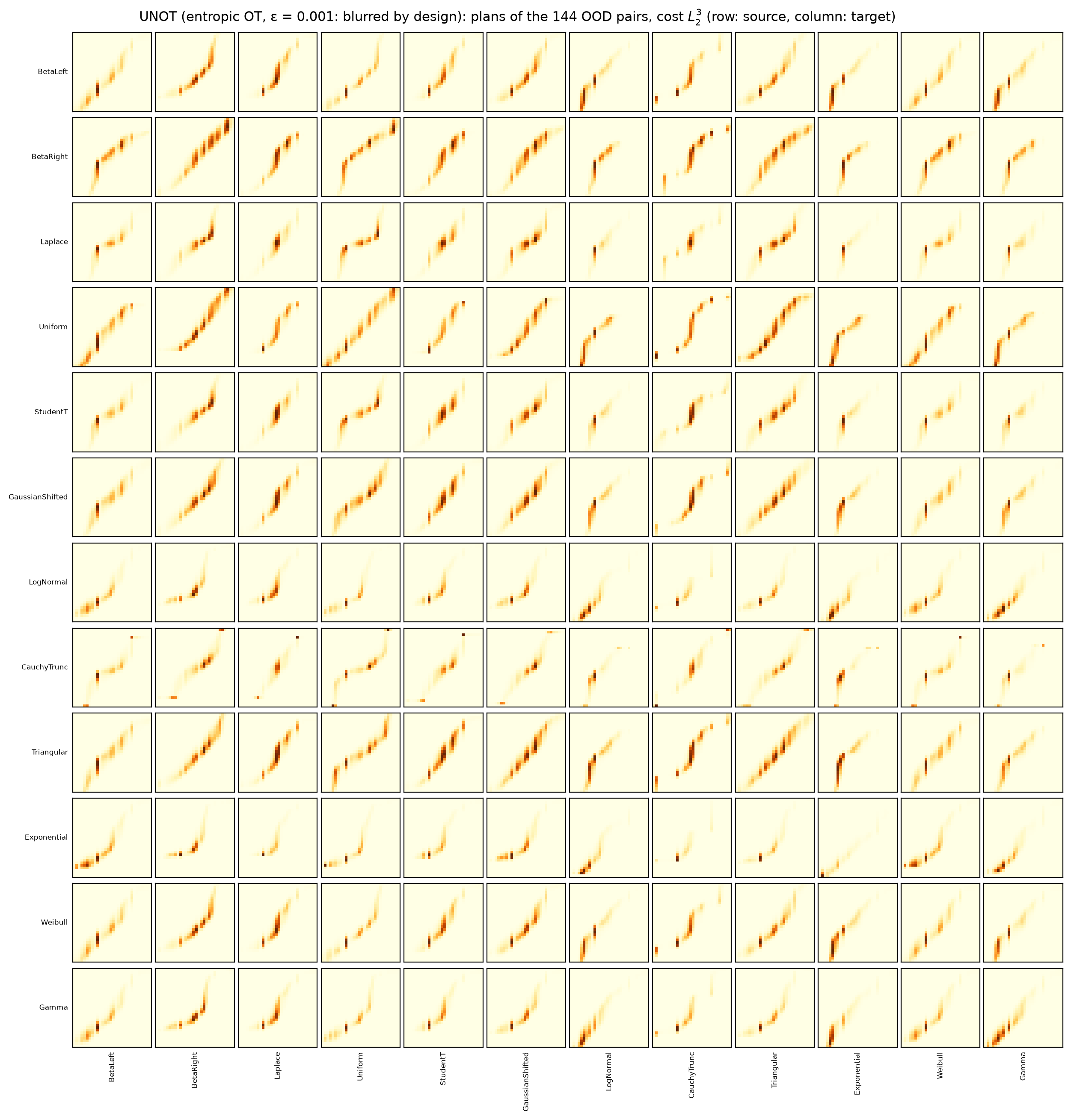}
    \caption{UNOT (entropic, $\varepsilon = 0.001$): plans of the 144 OOD pairs, cost $L_2^3$.}
    \label{fig:appE-pairs-cubic-euclidean-unot}
\end{figure}

\clearpage
\subsection{Cost type: Canberra}
\label{app:full-1d-canberra}

\begin{figure}[!ht]
    \centering
    \includegraphics[width=\linewidth]{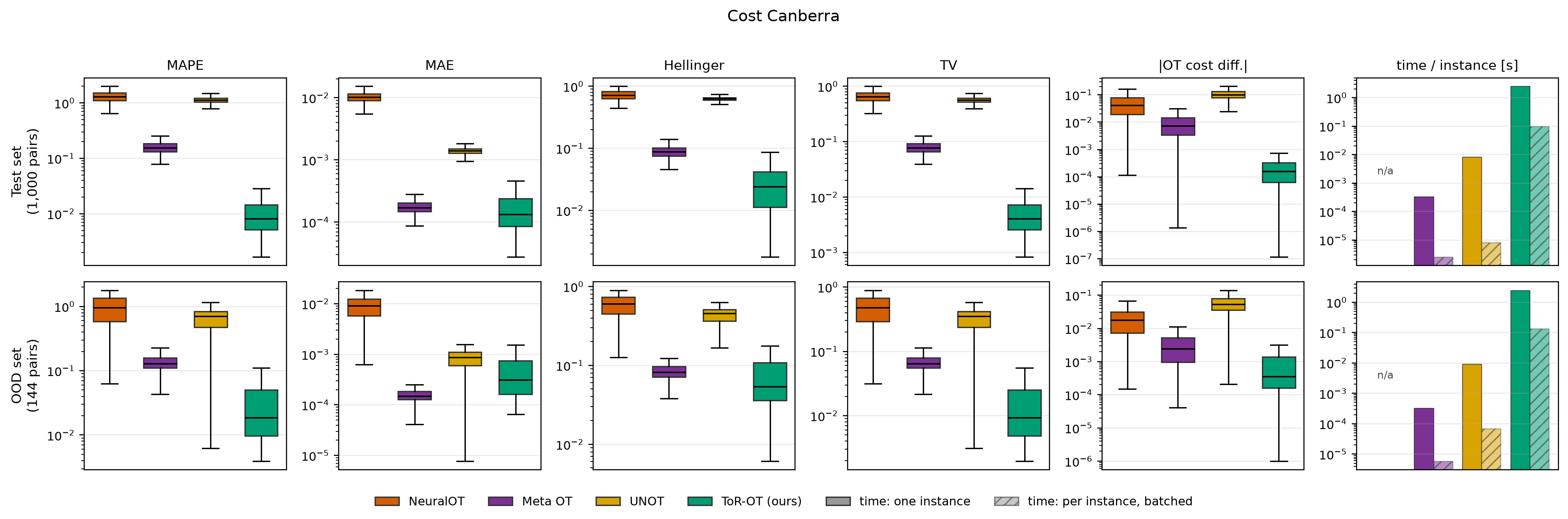}
    \caption{Cost Canberra: test set (top, 1{,}000 pairs) and OOD set (bottom, 144 pairs). Five plan metrics, each method against the solution of its own problem (box: quartiles; line: median; whiskers: 1.5 IQR; outliers not shown; log scale), and time per instance on one A100 GPU (solid: one instance; hatched: per instance, batched). NeuralOT optimizes each instance separately and its runs shared GPUs, so its times are not comparable (n/a).}
    \label{fig:appE-summary-canberra}
\end{figure}

Cost $c(x, y) = |x - y| / (|x| + |y|)$; Meta OT uses $\varepsilon = 0.01$ and UNOT $\varepsilon = 0.001$. ToR-OT's mean MAPE is 0.013 on the test set and 0.039 on the OOD set, against 0.154 and 0.127 for Meta OT, the strongest baseline (Fig.~\ref{fig:appE-summary-canberra}).

\begin{figure}[!ht]
    \centering
    \includegraphics[width=0.69\linewidth]{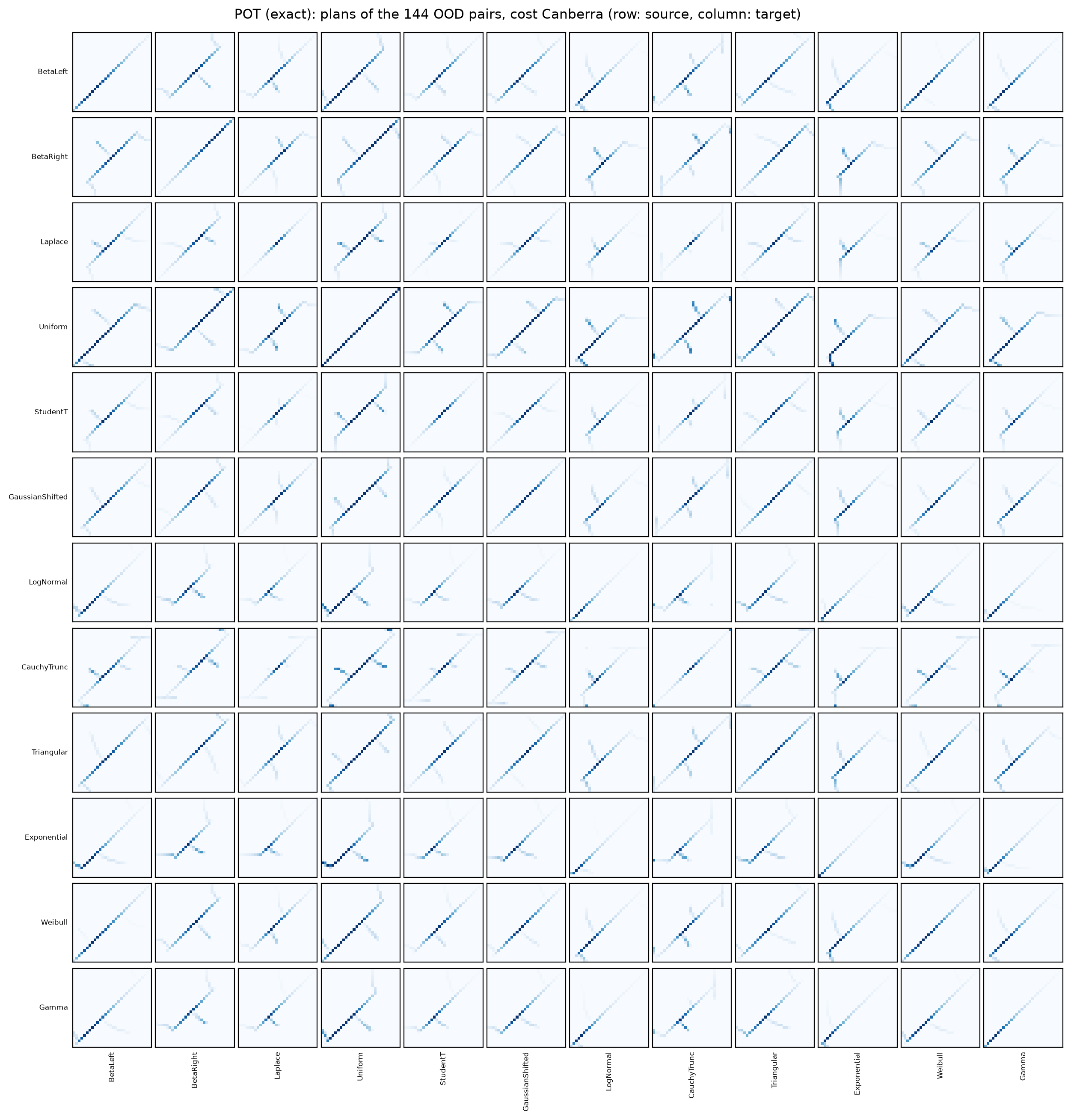}
    \caption{POT (exact): plans of the 144 OOD pairs, cost Canberra.}
    \label{fig:appE-pairs-canberra-exact}
\end{figure}

\begin{figure}[p]
    \centering
    \includegraphics[width=0.69\linewidth]{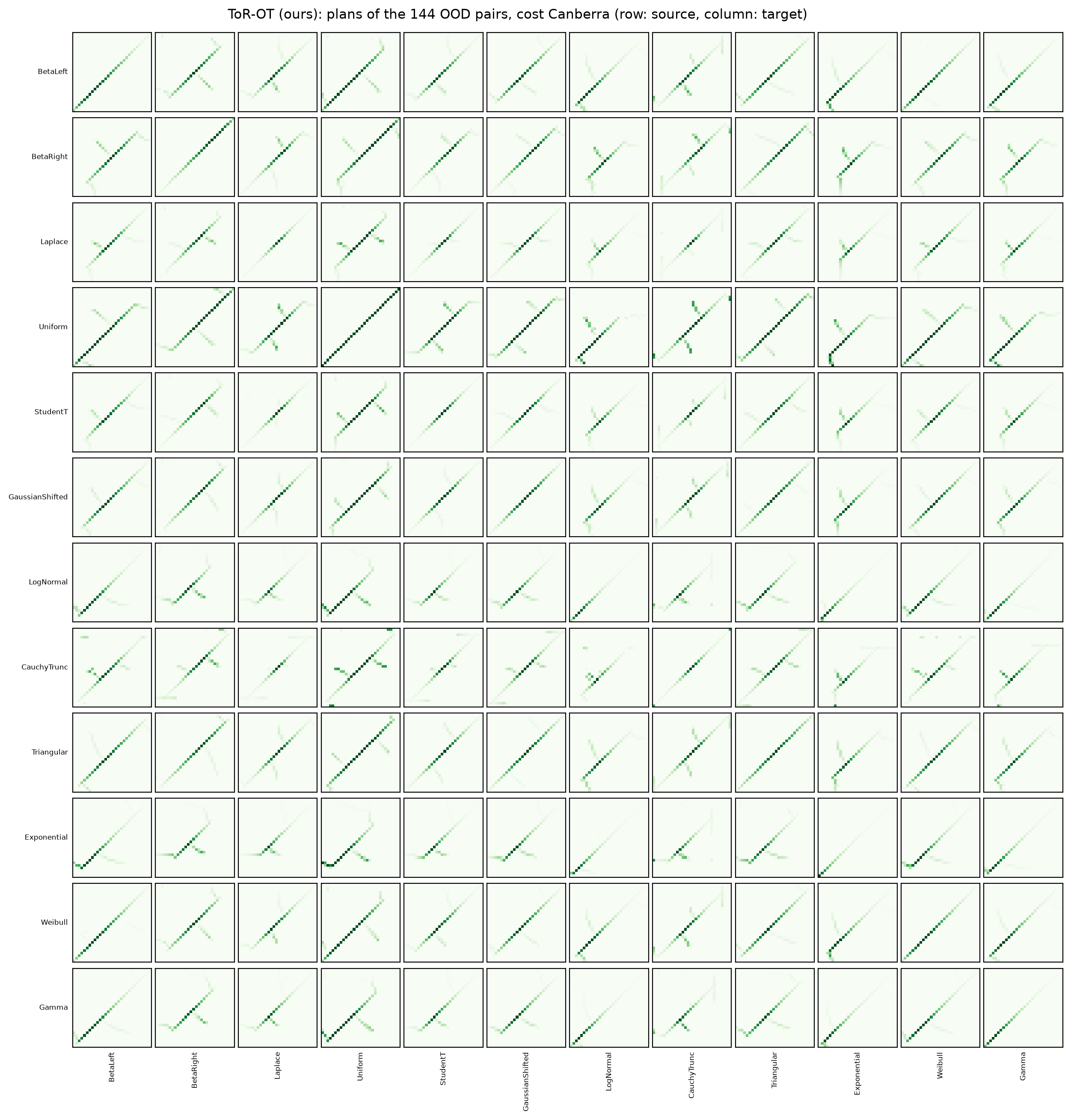}
    \caption{ToR-OT (ours): plans of the 144 OOD pairs, cost Canberra.}
    \label{fig:appE-pairs-canberra-v2row_lastrow}
    \vspace{0.5em}
    \includegraphics[width=0.69\linewidth]{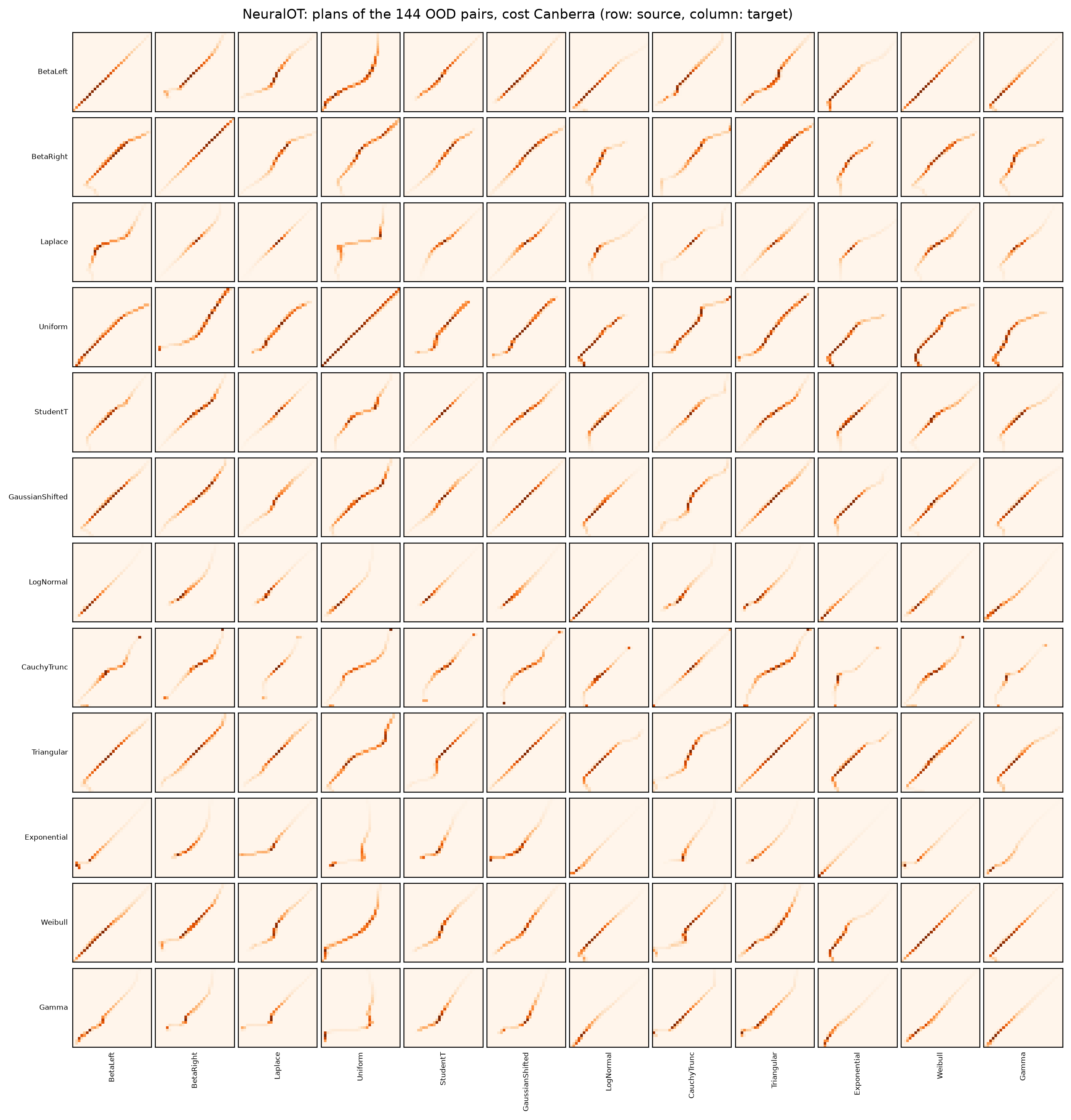}
    \caption{NeuralOT: plans of the 144 OOD pairs, cost Canberra.}
    \label{fig:appE-pairs-canberra-neuralot}
\end{figure}

\begin{figure}[p]
    \centering
    \includegraphics[width=0.69\linewidth]{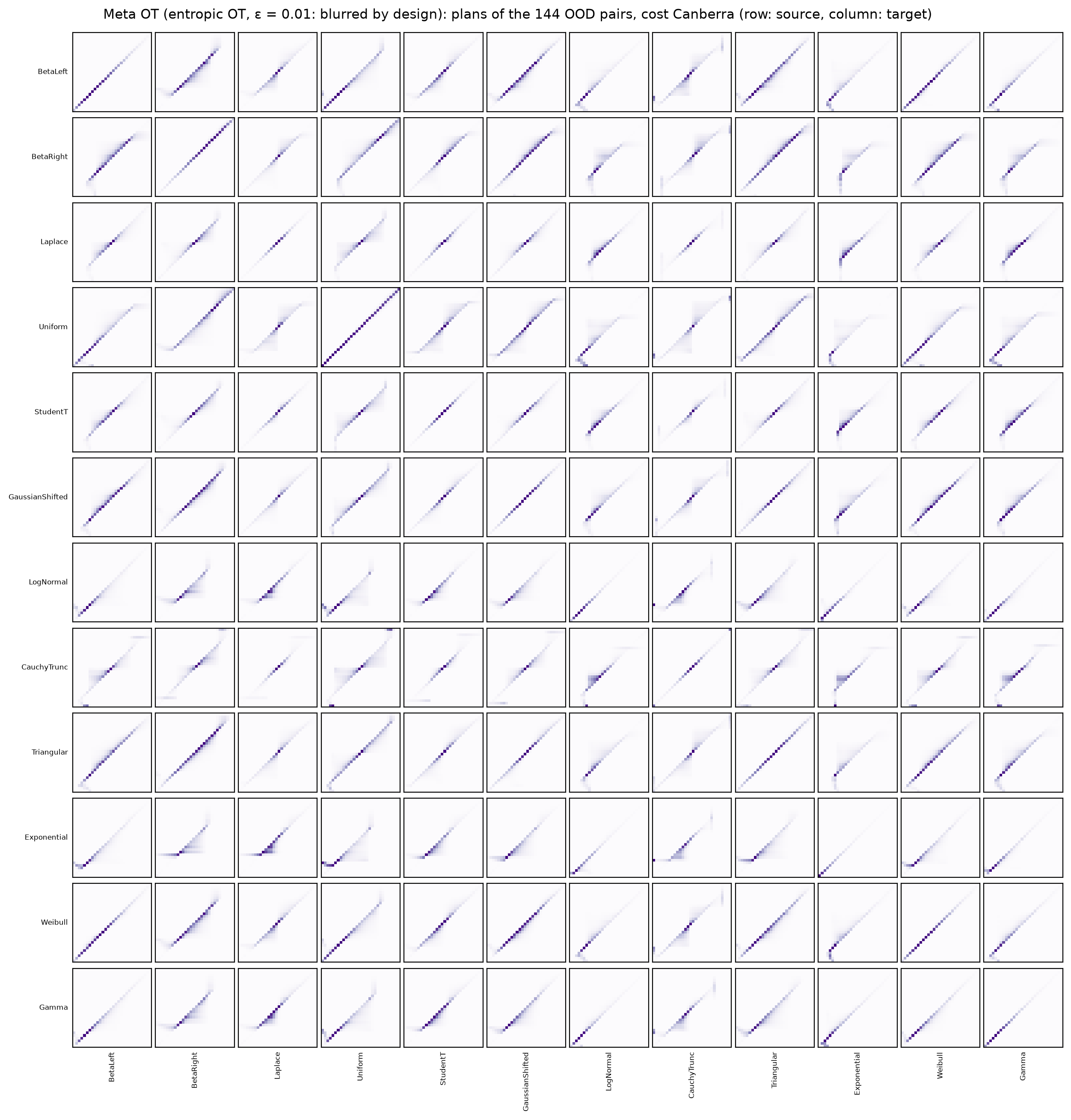}
    \caption{Meta OT (entropic, $\varepsilon = 0.01$): plans of the 144 OOD pairs, cost Canberra.}
    \label{fig:appE-pairs-canberra-metaot}
    \vspace{0.5em}
    \includegraphics[width=0.69\linewidth]{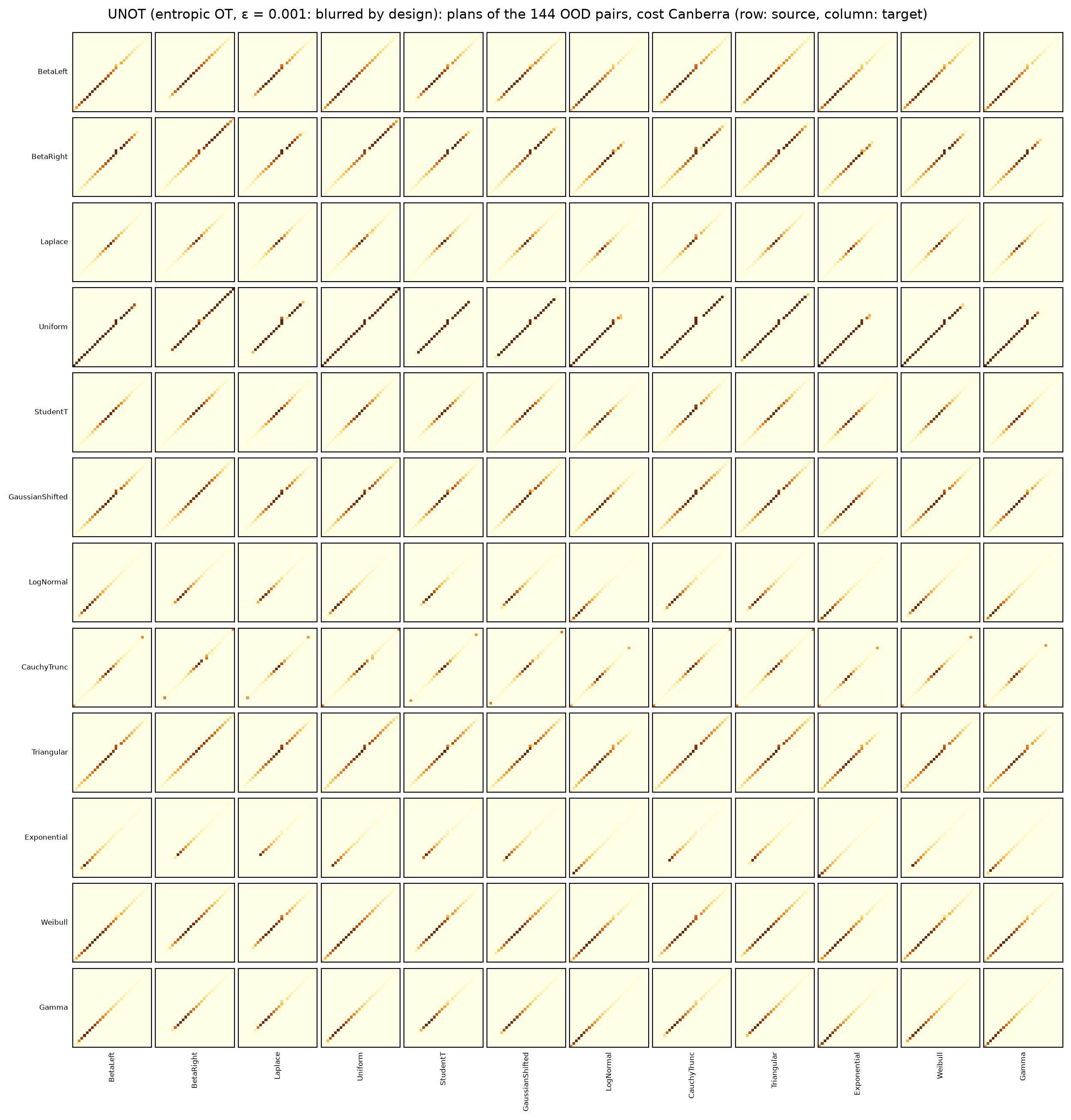}
    \caption{UNOT (entropic, $\varepsilon = 0.001$): plans of the 144 OOD pairs, cost Canberra.}
    \label{fig:appE-pairs-canberra-unot}
\end{figure}

\clearpage
\subsection{Cost type: Gaussian}
\label{app:full-1d-gaussian}

\begin{figure}[!ht]
    \centering
    \includegraphics[width=\linewidth]{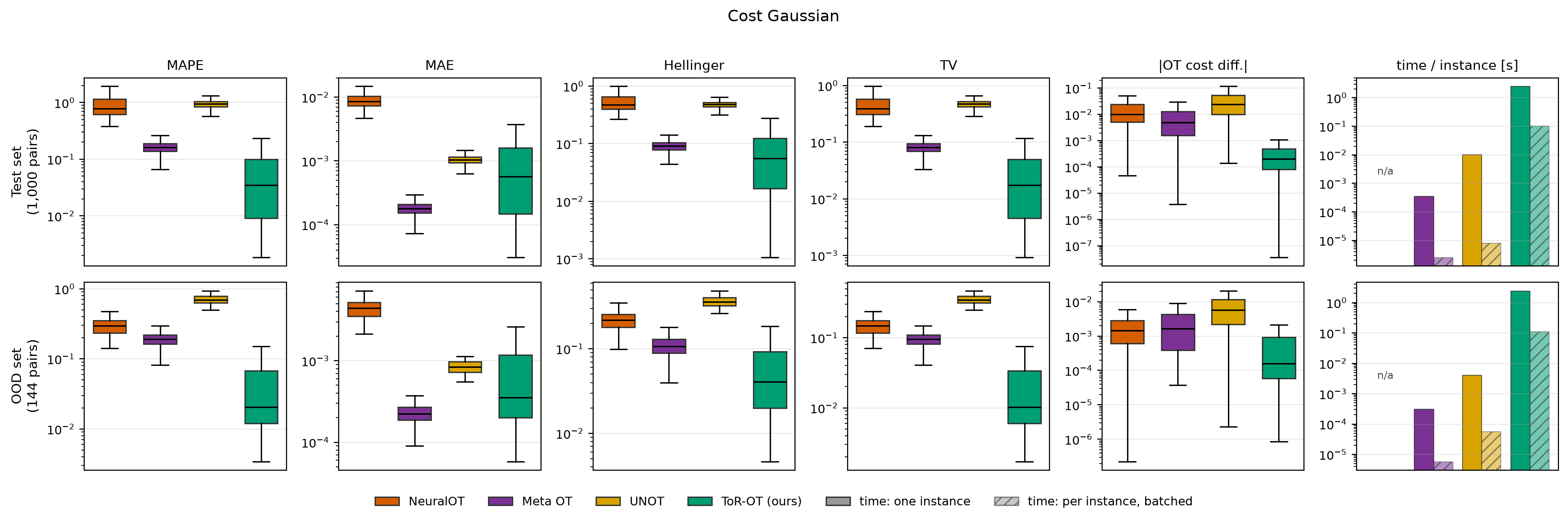}
    \caption{Cost Gaussian: test set (top, 1{,}000 pairs) and OOD set (bottom, 144 pairs). Five plan metrics, each method against the solution of its own problem (box: quartiles; line: median; whiskers: 1.5 IQR; outliers not shown; log scale), and time per instance on one A100 GPU (solid: one instance; hatched: per instance, batched). NeuralOT optimizes each instance separately and its runs shared GPUs, so its times are not comparable (n/a).}
    \label{fig:appE-summary-gaussian}
\end{figure}

Cost $c(x, y) = 1 - \exp(-|x - y|^2 / 2\sigma^2)$, $\sigma = 0.2$; Meta OT uses $\varepsilon = 0.01$ and UNOT $\varepsilon = 0.01$. ToR-OT's mean MAPE is 0.064 on the test set and 0.062 on the OOD set, against 0.176 and 0.200 for Meta OT, the strongest baseline (Fig.~\ref{fig:appE-summary-gaussian}).

\begin{figure}[!ht]
    \centering
    \includegraphics[width=0.69\linewidth]{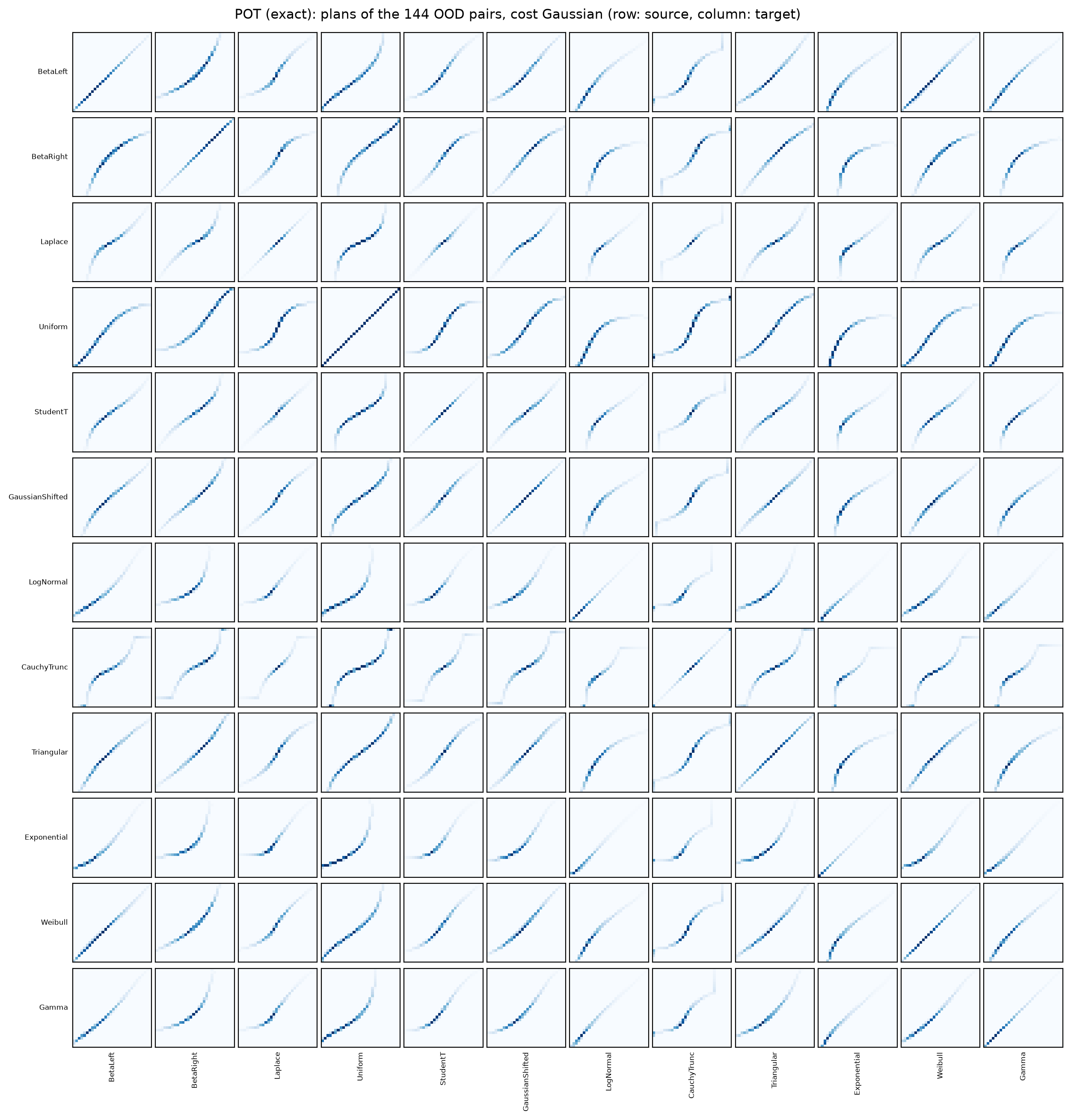}
    \caption{POT (exact): plans of the 144 OOD pairs, cost Gaussian.}
    \label{fig:appE-pairs-gaussian-exact}
\end{figure}

\begin{figure}[p]
    \centering
    \includegraphics[width=0.69\linewidth]{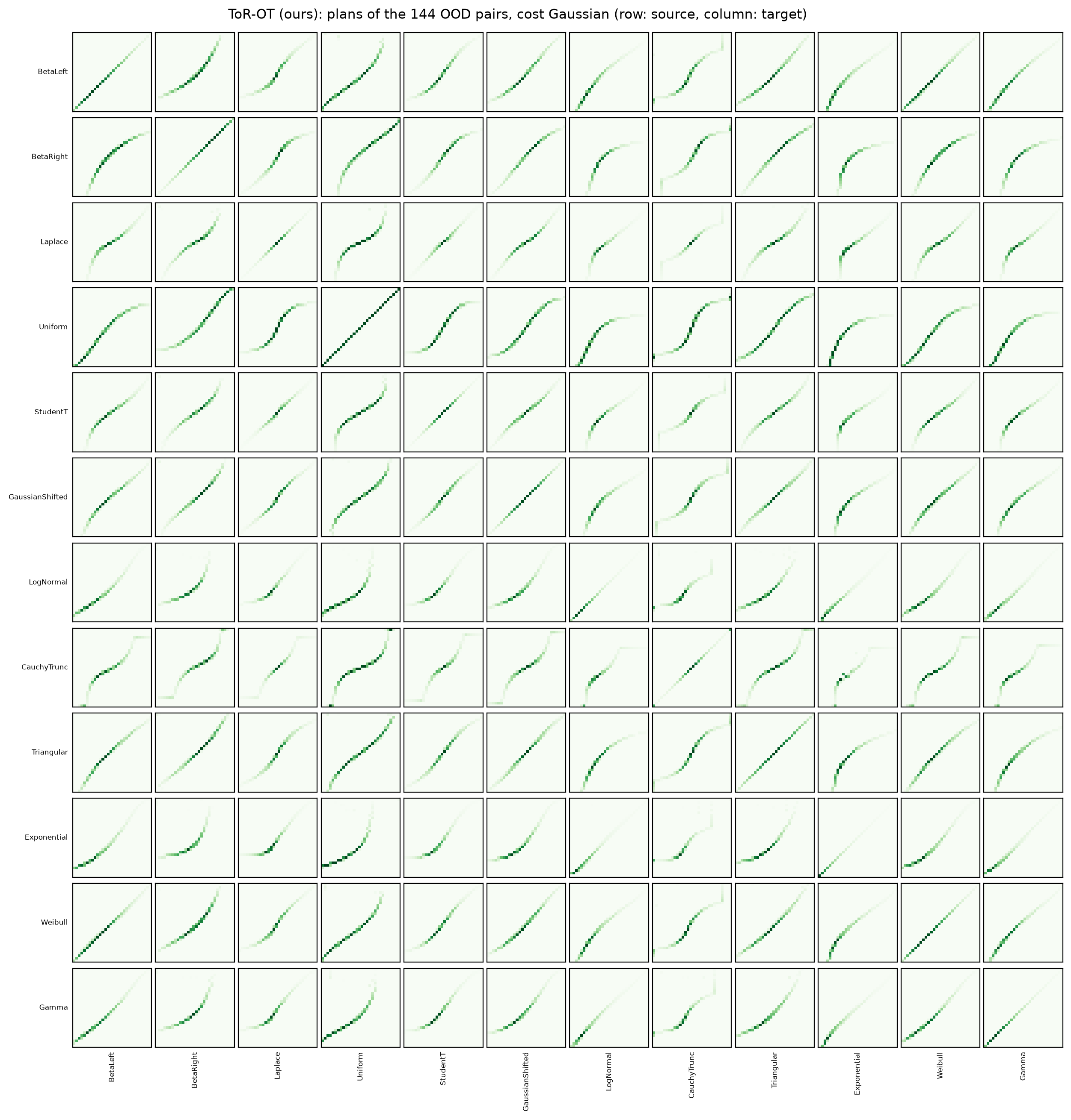}
    \caption{ToR-OT (ours): plans of the 144 OOD pairs, cost Gaussian.}
    \label{fig:appE-pairs-gaussian-v2row_lastrow}
    \vspace{0.5em}
    \includegraphics[width=0.69\linewidth]{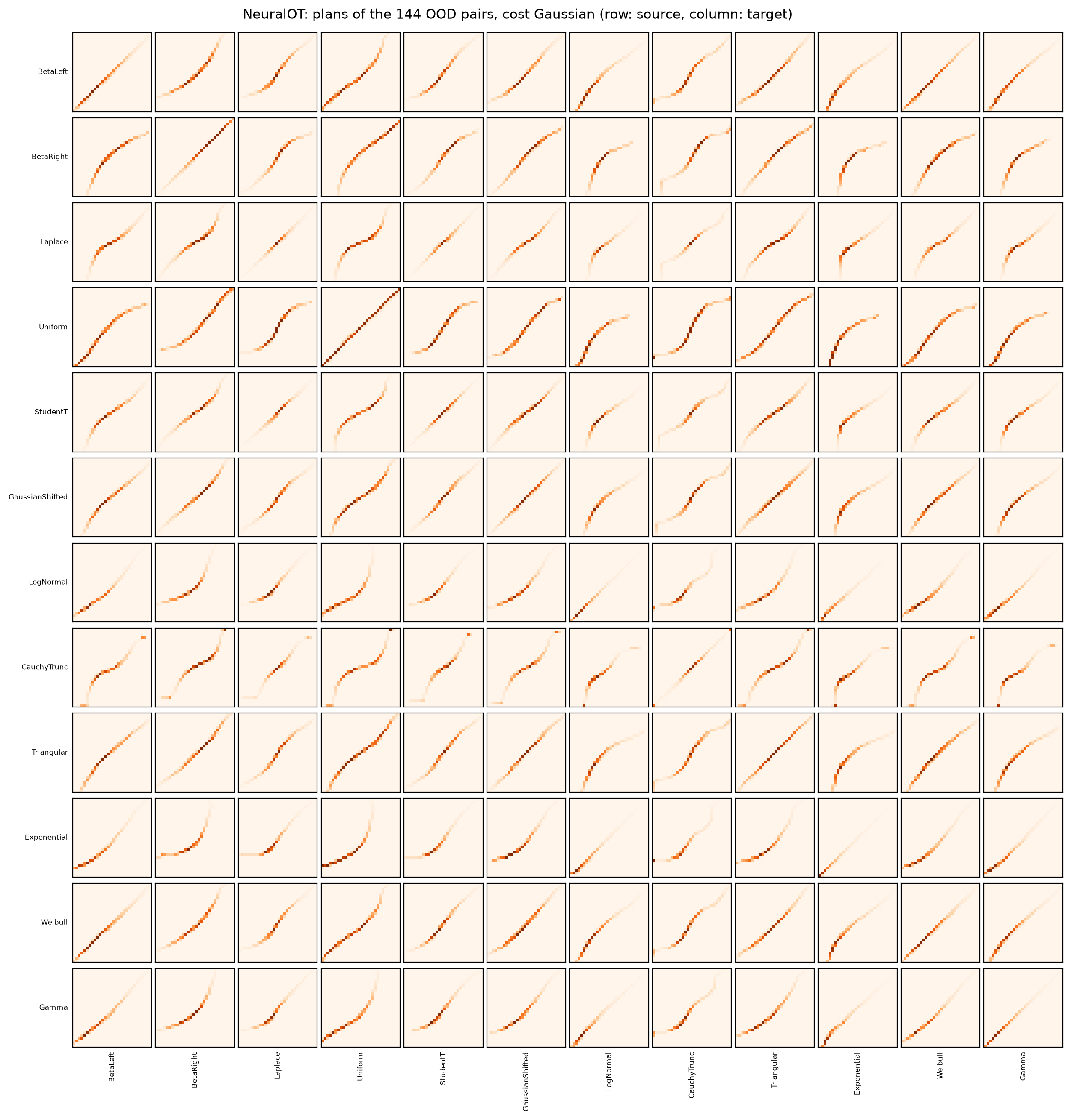}
    \caption{NeuralOT: plans of the 144 OOD pairs, cost Gaussian.}
    \label{fig:appE-pairs-gaussian-neuralot}
\end{figure}

\begin{figure}[p]
    \centering
    \includegraphics[width=0.69\linewidth]{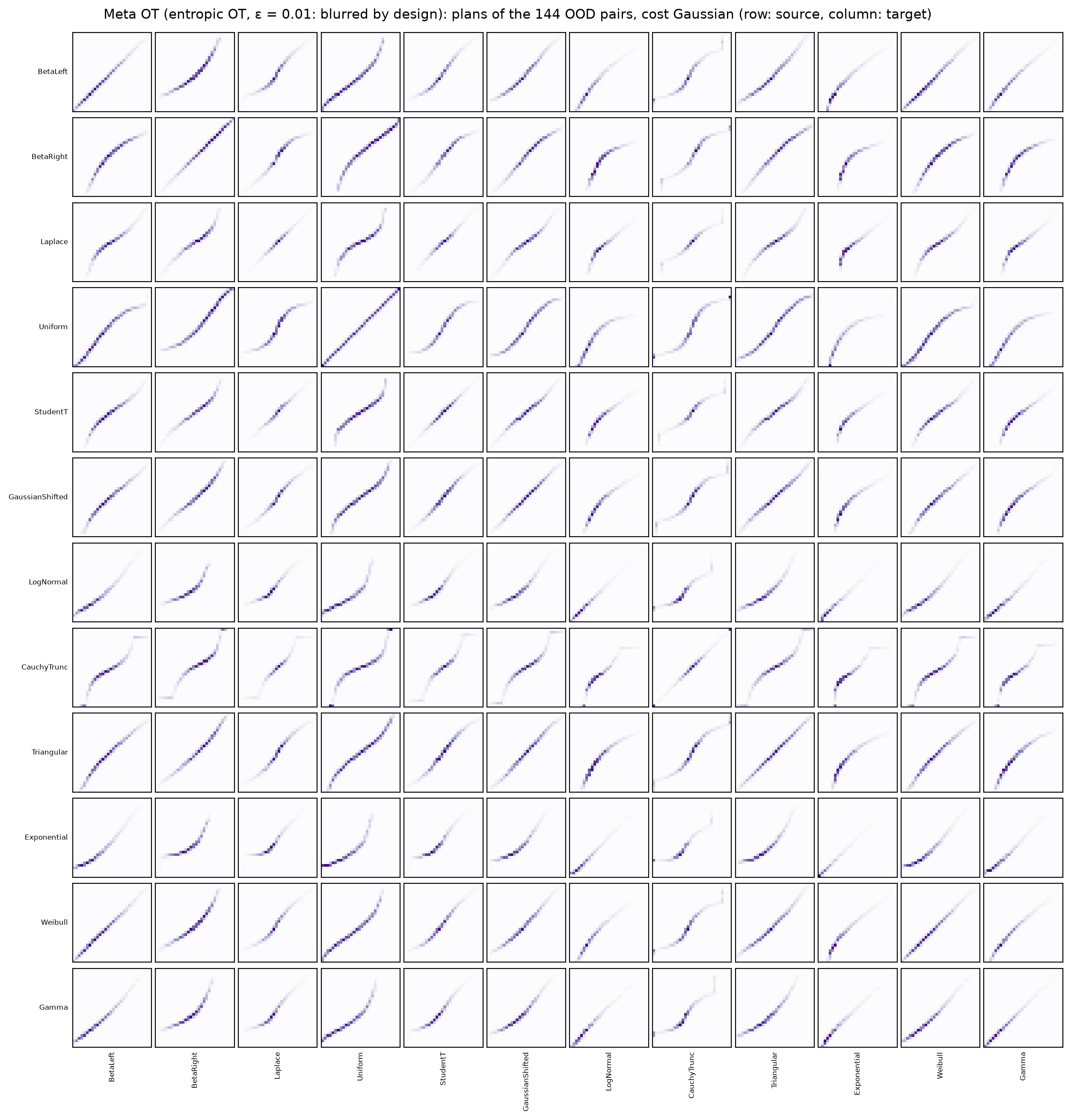}
    \caption{Meta OT (entropic, $\varepsilon = 0.01$): plans of the 144 OOD pairs, cost Gaussian.}
    \label{fig:appE-pairs-gaussian-metaot}
    \vspace{0.5em}
    \includegraphics[width=0.69\linewidth]{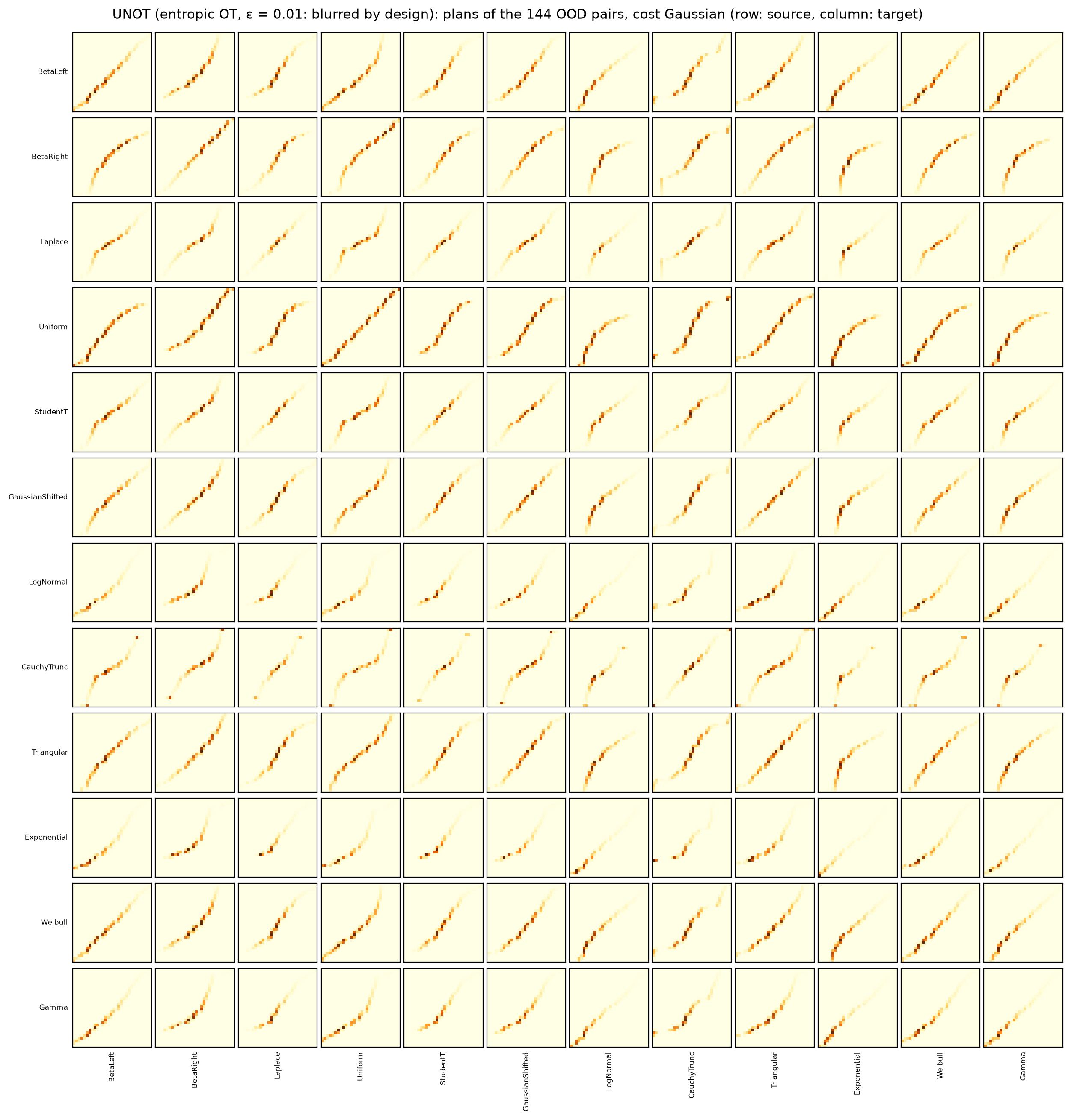}
    \caption{UNOT (entropic, $\varepsilon = 0.01$): plans of the 144 OOD pairs, cost Gaussian.}
    \label{fig:appE-pairs-gaussian-unot}
\end{figure}

\clearpage
\subsection{All six cost types}
\label{app:full-1d-all}

\begin{figure}[!ht]
    \centering
    \includegraphics[width=\linewidth]{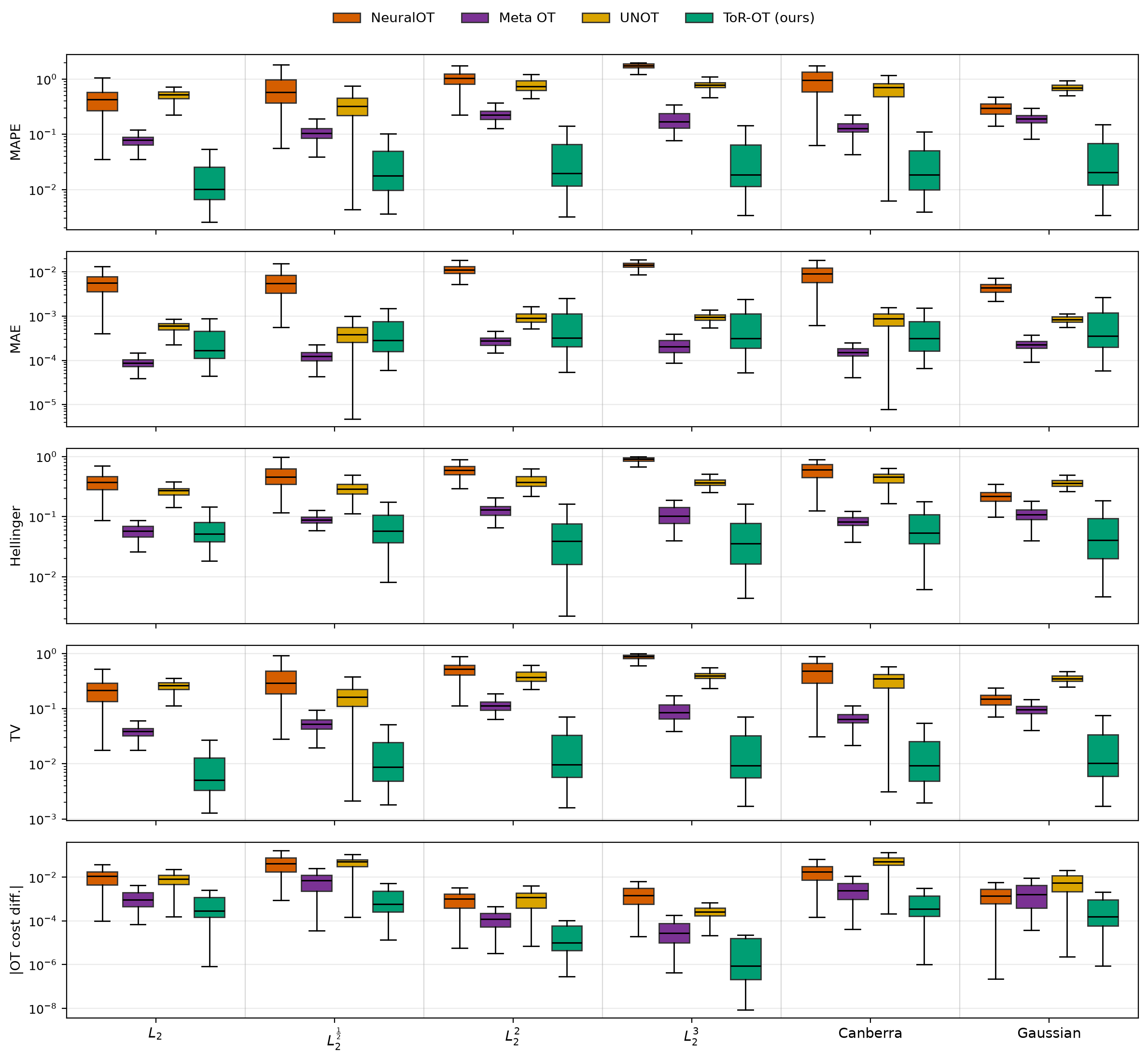}
    \caption{All six costs, OOD set (144 pairs per cost): the five plan metrics, each method against the solution of
    its own problem (box: quartiles; line: median; whiskers: 1.5 IQR; outliers not shown; log scale).}
    \label{fig:appE-allcosts-ood}
\end{figure}

\begin{figure}[!ht]
    \centering
    \includegraphics[width=\linewidth]{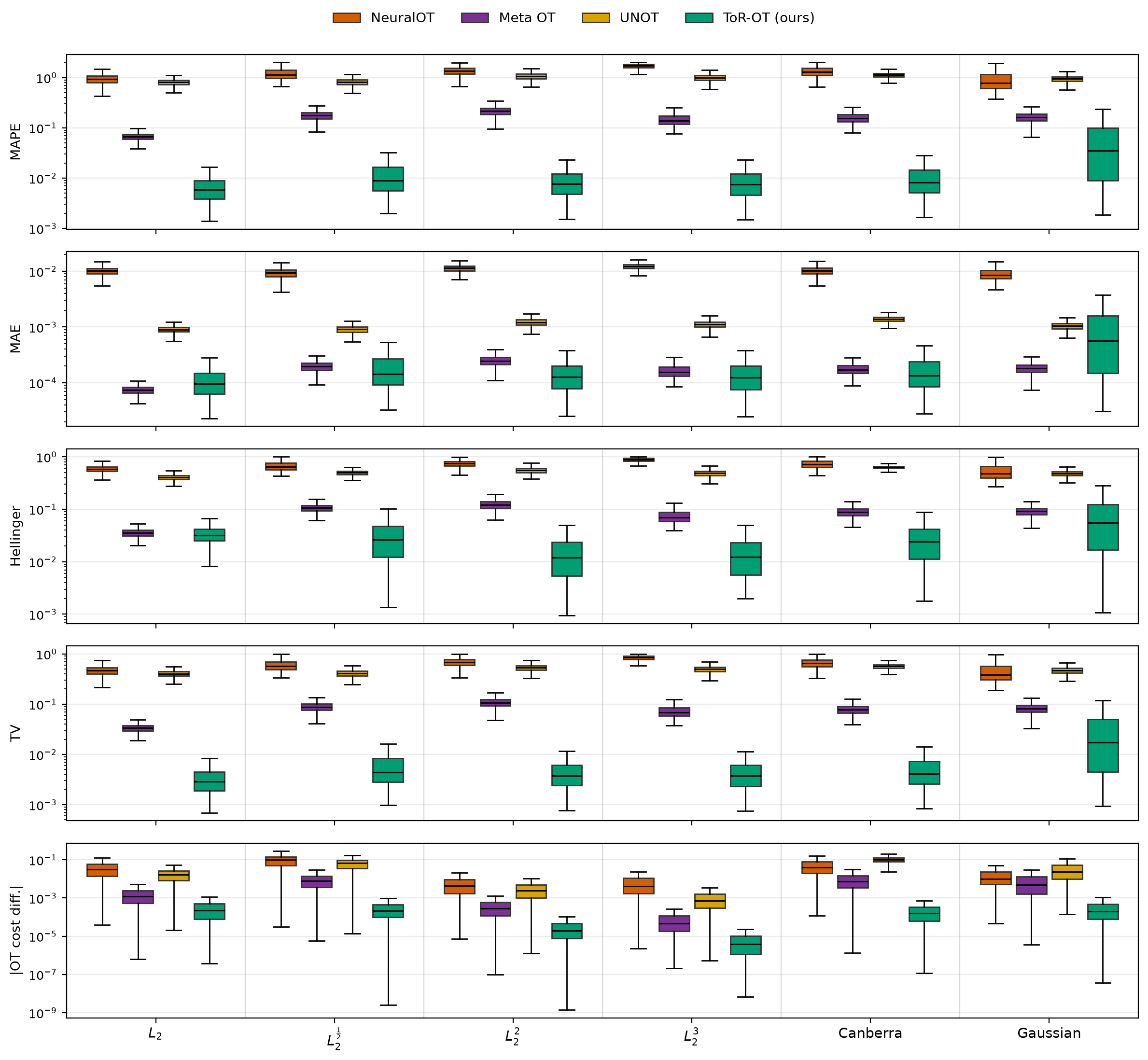}
    \caption{All six costs, test set (1{,}000 pairs per cost), as Fig.~\ref{fig:appE-allcosts-ood}.}
    \label{fig:appE-allcosts-test}
\end{figure}

\begin{figure}[ht]
    \centering
    \includegraphics[width=\linewidth]{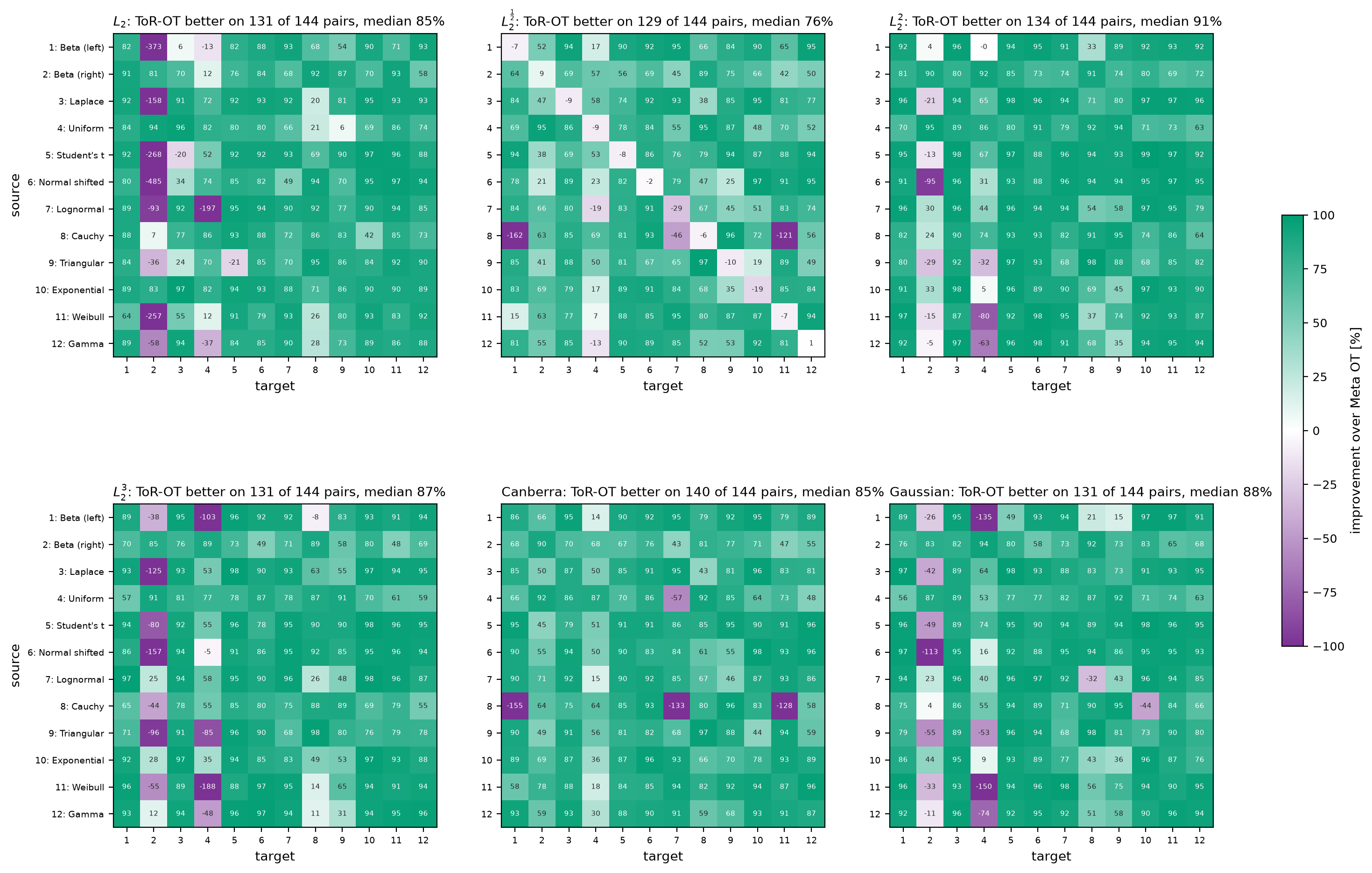}
    \caption{Relative MAPE improvement of ToR-OT over Meta OT, $100\,(\mathrm{MAPE}_{\text{Meta OT}} -
    \mathrm{MAPE}_{\text{ToR-OT}}) / \mathrm{MAPE}_{\text{Meta OT}}$, on each of the 144 OOD pairs (row: source;
    column: target; distributions numbered as in the first column), for each cost; each method is scored against the
    solution of its own problem. Green: ToR-OT better; purple: Meta OT better; the color scale is clipped at
    $\pm 100\%$. Panel titles: the number of pairs on which ToR-OT is better and the median improvement.}
    \label{fig:appE-improvement}
\end{figure}

Figures~\ref{fig:appE-allcosts-ood} and~\ref{fig:appE-allcosts-test} show every method on every cost, one panel per
metric. Fig.~\ref{fig:appE-improvement} shows, per OOD pair, how much ToR-OT improves over Meta OT, the strongest
baseline on every cost: ToR-OT is better on 129 to 140 of the 144 pairs, with median improvements from 76\% to 91\%.

\newpage
\clearpage
\makeatletter
\ifx\Ginput@path\@undefined\def\Ginput@path{}\fi
\edef\Ginput@path{\unexpanded\expandafter{\Ginput@path}{./}{figures_cameraready/}{figures_cameraready/appendix_F/}{appendix_F/}}
\makeatother

\section{Full results on 2D distributions}
\label{app:full-2d}

\providecommand{\appEtocline}[2]{\noindent\hspace*{1em}\makebox[2.2em][l]{\ref{#1}}#2\ \dotfill\ \pageref{#1}\par}
\noindent\textbf{Contents}\par\smallskip
\appEtocline{app:full-2d-euclidean}{Cost type: $L_2$}
\appEtocline{app:full-2d-sqrt-euclidean}{Cost type: $L_2^{1/2}$}
\appEtocline{app:full-2d-sqeuclidean}{Cost type: $L_2^2$}
\appEtocline{app:full-2d-cubic-euclidean}{Cost type: $L_2^3$}
\appEtocline{app:full-2d-canberra}{Cost type: Canberra}
\appEtocline{app:full-2d-gaussian}{Cost type: Gaussian}
\appEtocline{app:full-2d-all}{All six cost types}
\medskip

Each cost-type subsection has a summary of the five plan metrics and the run time (test and OOD sets), followed by
the plans of the 36 OOD pairs for POT, ToR-OT, Meta OT and UNOT (and NeuralOT for $L_2^2$ and $L_2$).

\begin{figure}[!ht]
    \centering
    \includegraphics[width=\linewidth]{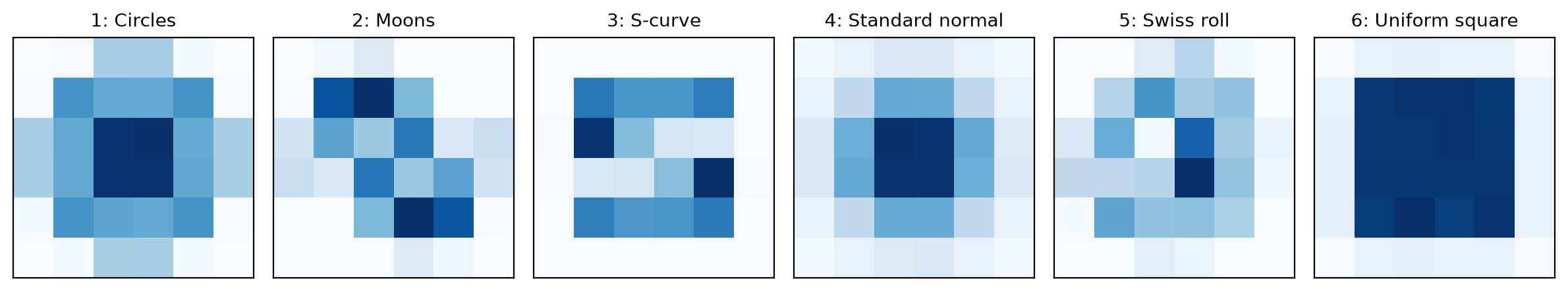}
    \caption{The 6 distributions of the 2D OOD set on the $6 \times 6$ grid, numbered as in the plan grids and the
    improvement heatmaps.}
    \label{fig:appF-distributions}
\end{figure}

This appendix repeats App.~\ref{app:full-1d} for two-dimensional distributions, with the 2D model of App.~\ref{app:training}.
Supports are the centres of a $6 \times 6$ grid on $[0, 1]^2$ ($n = m = 36$); the $36 \times 36$ plans list the bins
with the second coordinate varying fastest. The six costs are those of the 1D model with $\lVert x - y \rVert_2$ in
place of $|x - y|$ (definitions in each subsection). The two sets:
\begin{itemize}
    \item \textbf{Test set:} 1{,}000 random pairs per cost, marginals uniform on the simplex (the training distribution).
    \item \textbf{OOD set:} the 36 ordered pairs of 6 named distributions (Fig.~\ref{fig:appF-distributions}): the
          2D OOD set of the submitted version without 8 Gaussians, which has 24 empty bins of 36, a case the 2D
          training data never contains (its Dirichlet marginals have no empty bins).
\end{itemize}
Scoring, metrics and plots are as in App.~\ref{app:full-1d}: each method is compared with the solution of the problem
it solves (exact OT: the optimal plan closest to the prediction; entropic OT: the converged Sinkhorn plan at the
method's $\varepsilon$). NeuralOT's 2D implementation supports only $L_2^2$ and $L_2$, and it optimizes each pair
separately, so it was run on those two costs: all OOD pairs, and 1{,}000 ($L_2^2$) and 200 ($L_2$, spread over the
set) test pairs. The box plots use training seed 0 of the learned baselines; the numbers in the
text are means over their three seeds.

\subsection{Cost type: $L_2$}
\label{app:full-2d-euclidean}

\begin{figure}[!ht]
    \centering
    \includegraphics[width=\linewidth]{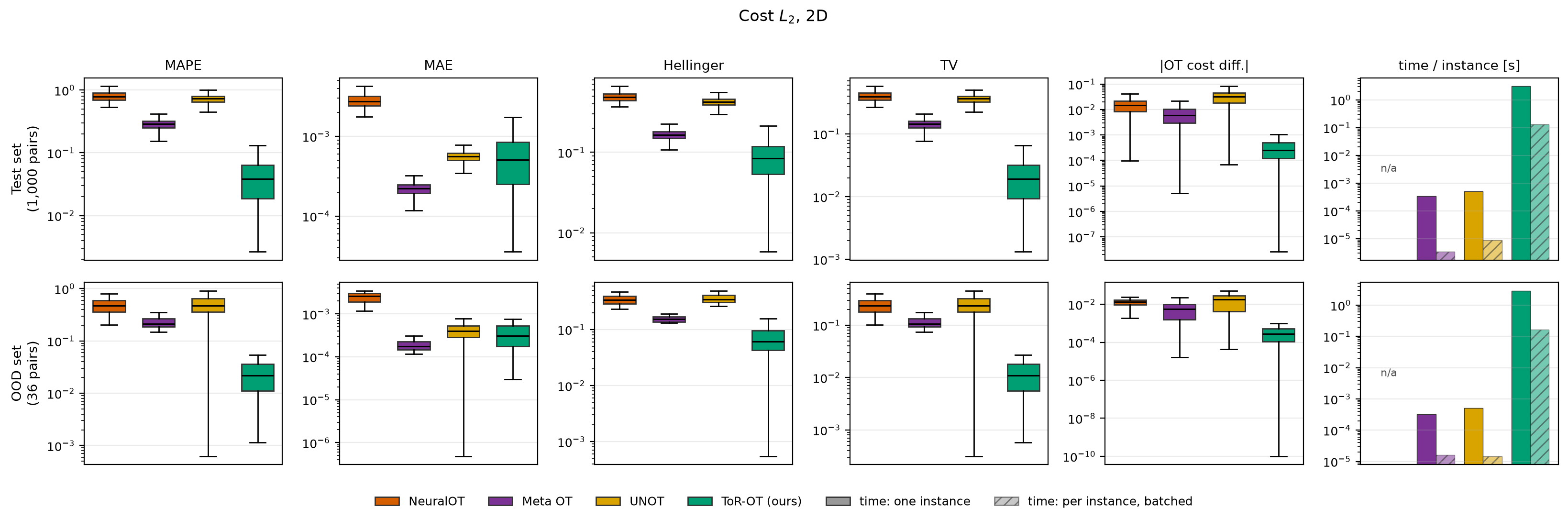}
    \caption{Cost $L_2$, 2D: test set (top, 1{,}000 pairs) and OOD set (bottom, 36 pairs). Five plan metrics, each method against the solution of its own problem (box: quartiles; line: median; whiskers: 1.5 IQR; outliers not shown; log scale), and time per instance on one A100 GPU (solid: one instance; hatched: per instance, batched). NeuralOT optimizes each instance separately and its runs shared GPUs, so its times are not comparable (n/a). NeuralOT: 200 of the 1{,}000 test pairs.}
    \label{fig:appF-summary-euclidean}
\end{figure}

\begin{figure}[!ht]
    \centering
    \includegraphics[width=0.69\linewidth]{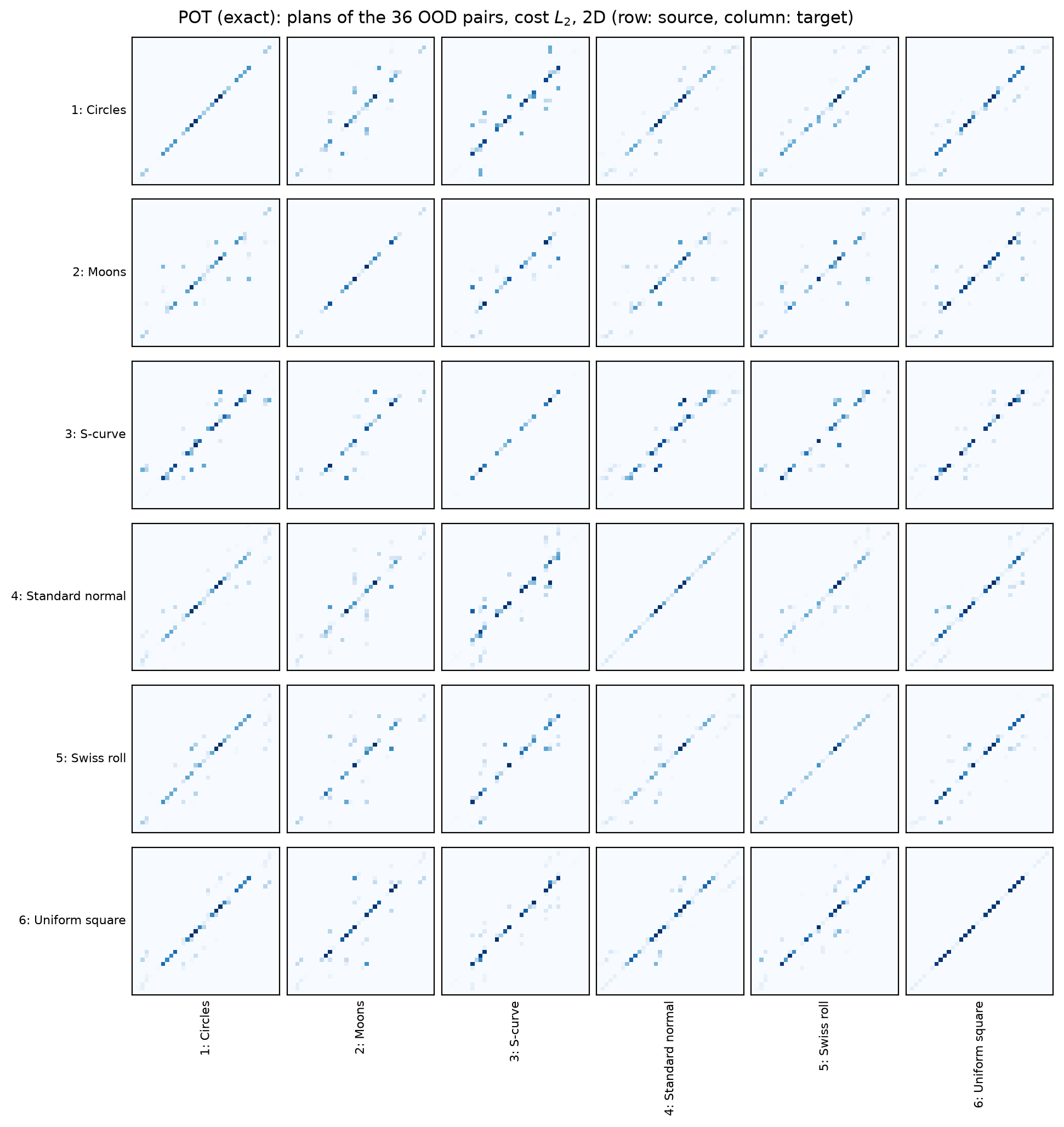}
    \caption{POT (exact): plans of the 36 OOD pairs, cost $L_2$, 2D.}
    \label{fig:appF-pairs-euclidean-exact}
\end{figure}

Cost $c(x, y) = \lVert x - y \rVert_2$; Meta OT and UNOT use $\varepsilon = 0.01$. ToR-OT's mean MAPE is 0.047 on the test set and 0.036 on the OOD set, against 0.286 and 0.202 for Meta OT, the strongest baseline; ToR-OT is better on 29 of the 36 OOD pairs (Fig.~\ref{fig:appF-summary-euclidean}). As in 1D, the optimal plan for $L_2$ is often not unique (POT's plan, Fig.~\ref{fig:appF-pairs-euclidean-exact}, is one of several), and ToR-OT and NeuralOT are scored against the optimal plan closest to their prediction.

\begin{figure}[p]
    \centering
    \includegraphics[width=0.69\linewidth]{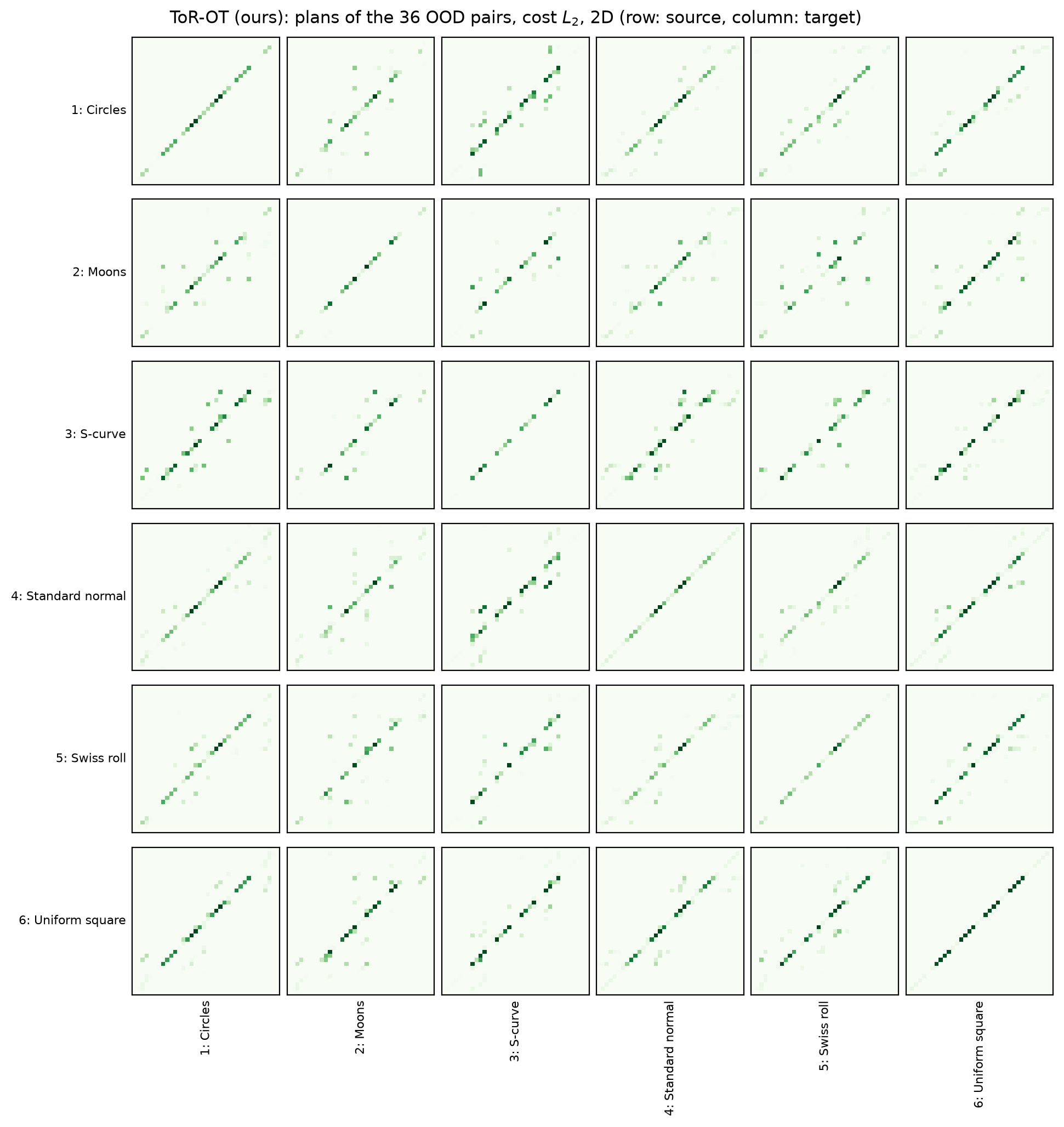}
    \caption{ToR-OT (ours): plans of the 36 OOD pairs, cost $L_2$, 2D.}
    \label{fig:appF-pairs-euclidean-v2row_lastrow}
    \vspace{0.5em}
    \includegraphics[width=0.69\linewidth]{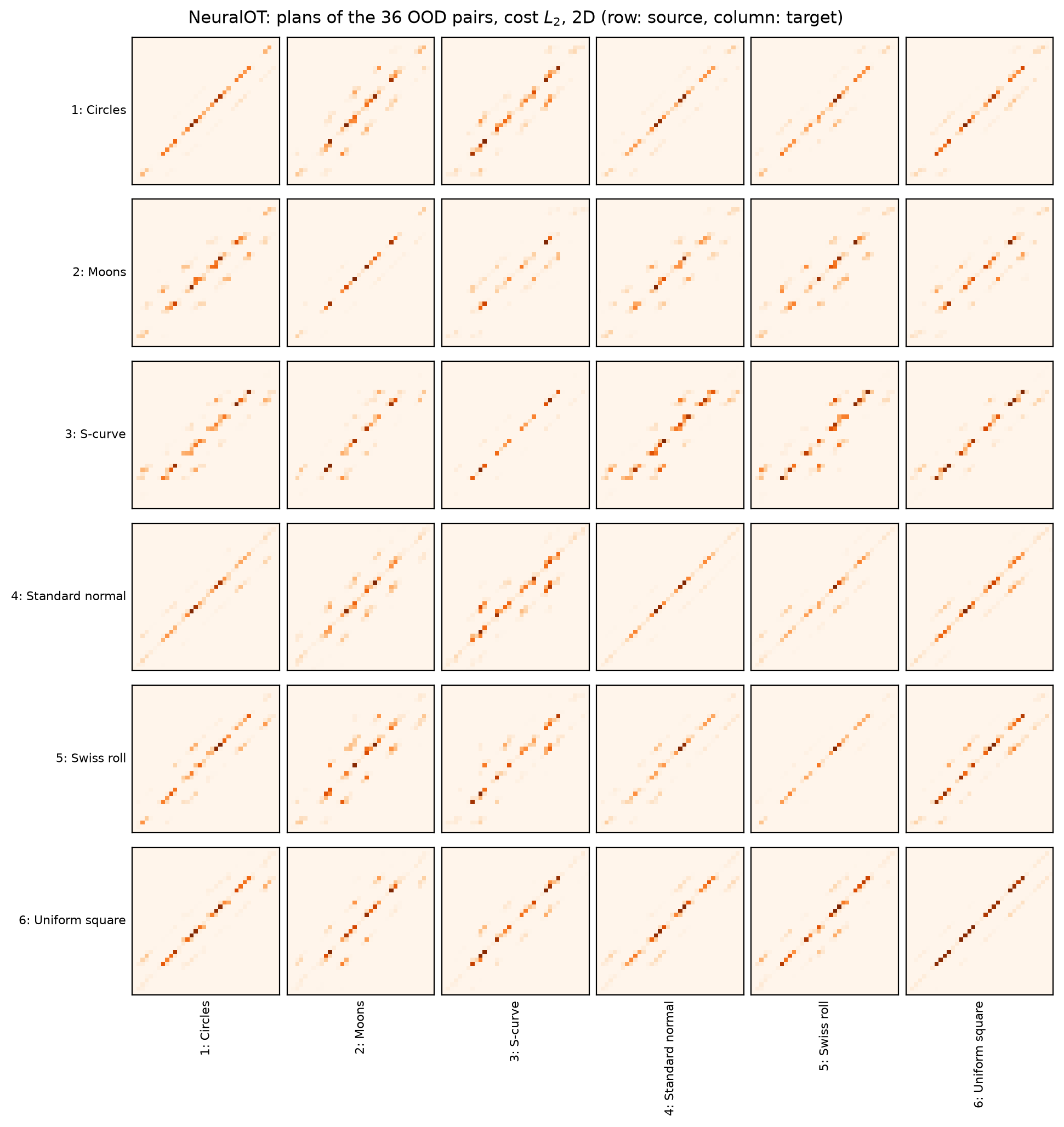}
    \caption{NeuralOT: plans of the 36 OOD pairs, cost $L_2$, 2D.}
    \label{fig:appF-pairs-euclidean-neuralot}
\end{figure}

\begin{figure}[p]
    \centering
    \includegraphics[width=0.69\linewidth]{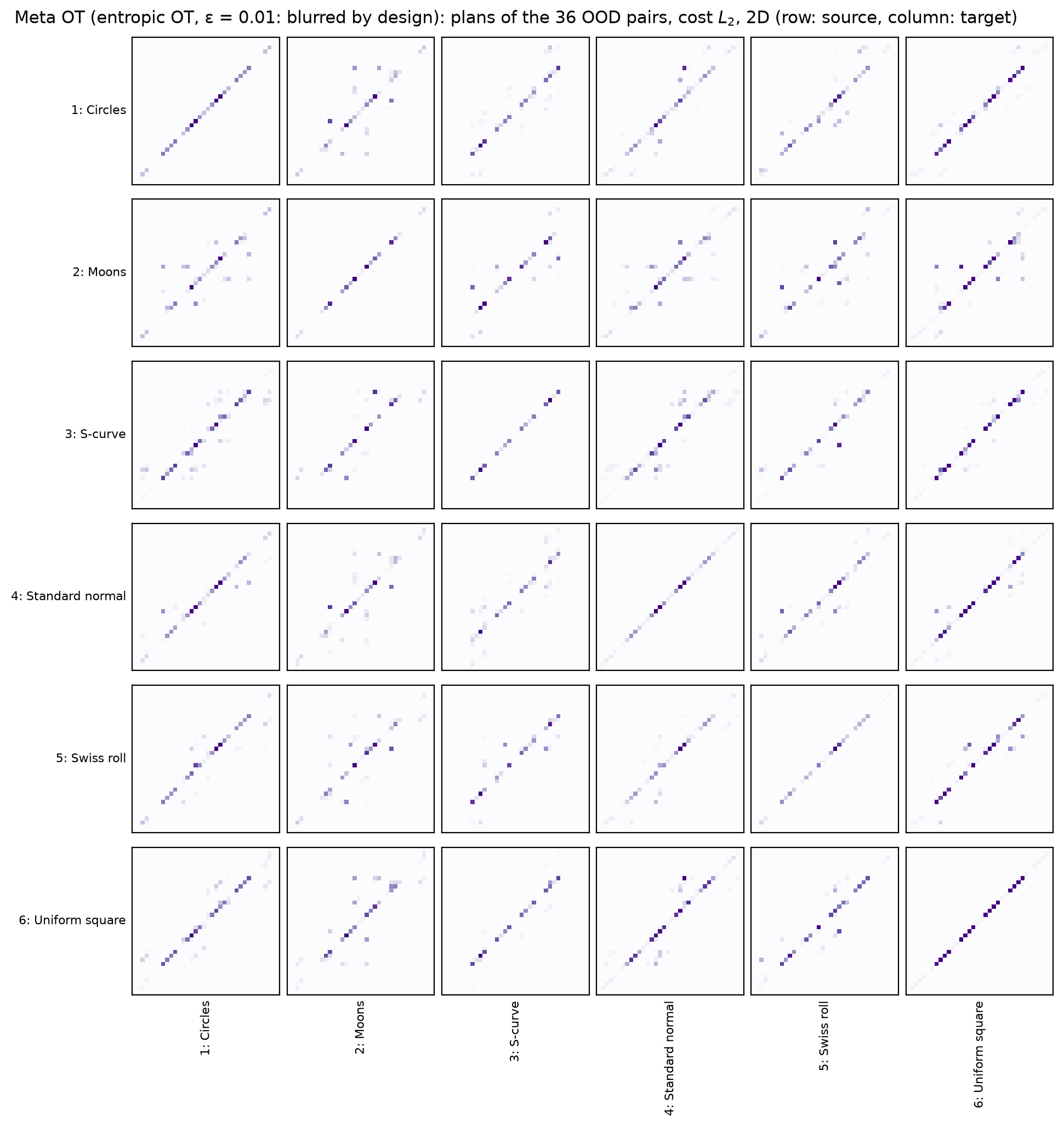}
    \caption{Meta OT (entropic, $\varepsilon = 0.01$): plans of the 36 OOD pairs, cost $L_2$, 2D.}
    \label{fig:appF-pairs-euclidean-metaot}
    \vspace{0.5em}
    \includegraphics[width=0.69\linewidth]{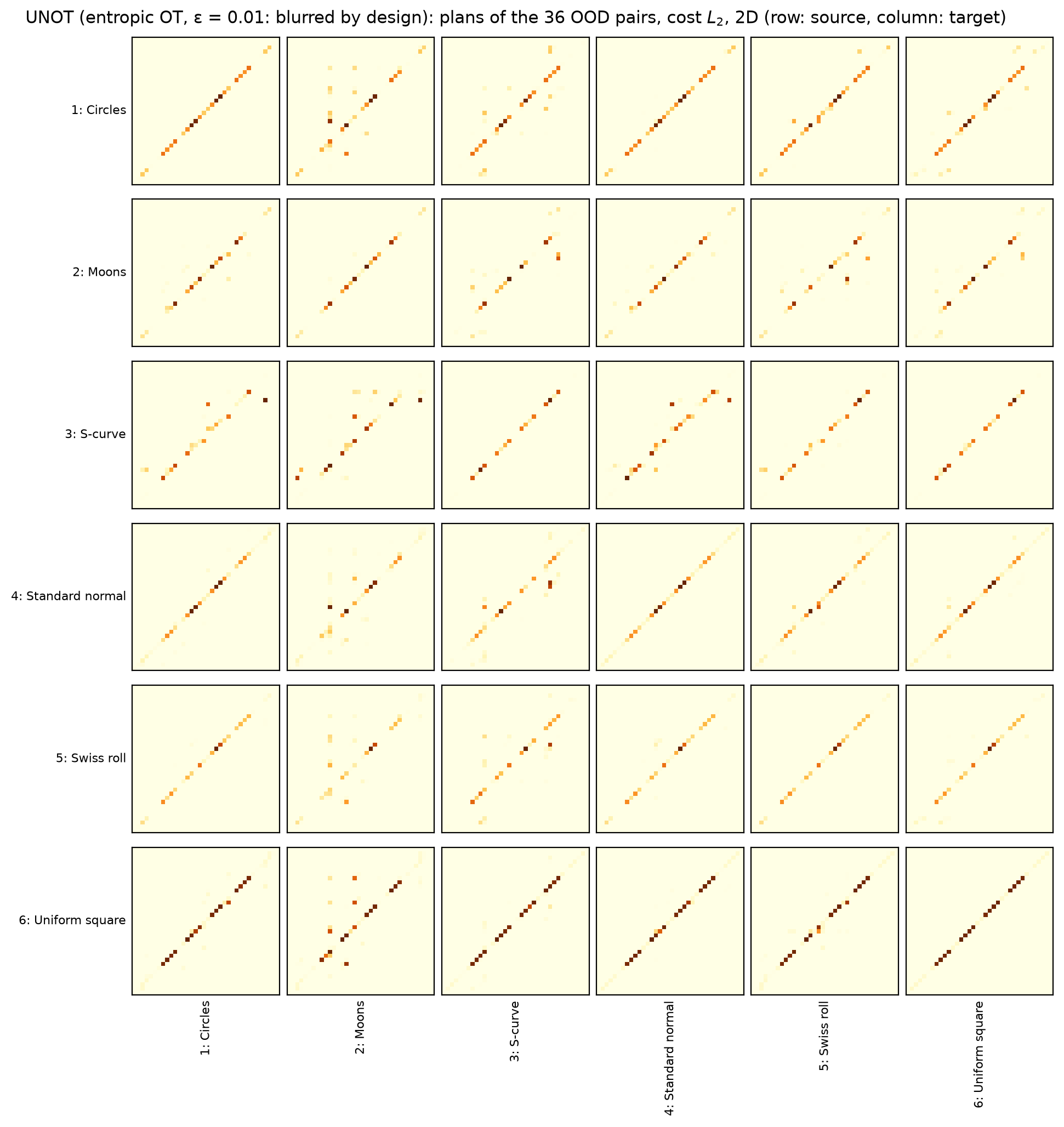}
    \caption{UNOT (entropic, $\varepsilon = 0.01$): plans of the 36 OOD pairs, cost $L_2$, 2D.}
    \label{fig:appF-pairs-euclidean-unot}
\end{figure}

\clearpage
\subsection{Cost type: $L_2^{1/2}$}
\label{app:full-2d-sqrt-euclidean}

\begin{figure}[!ht]
    \centering
    \includegraphics[width=\linewidth]{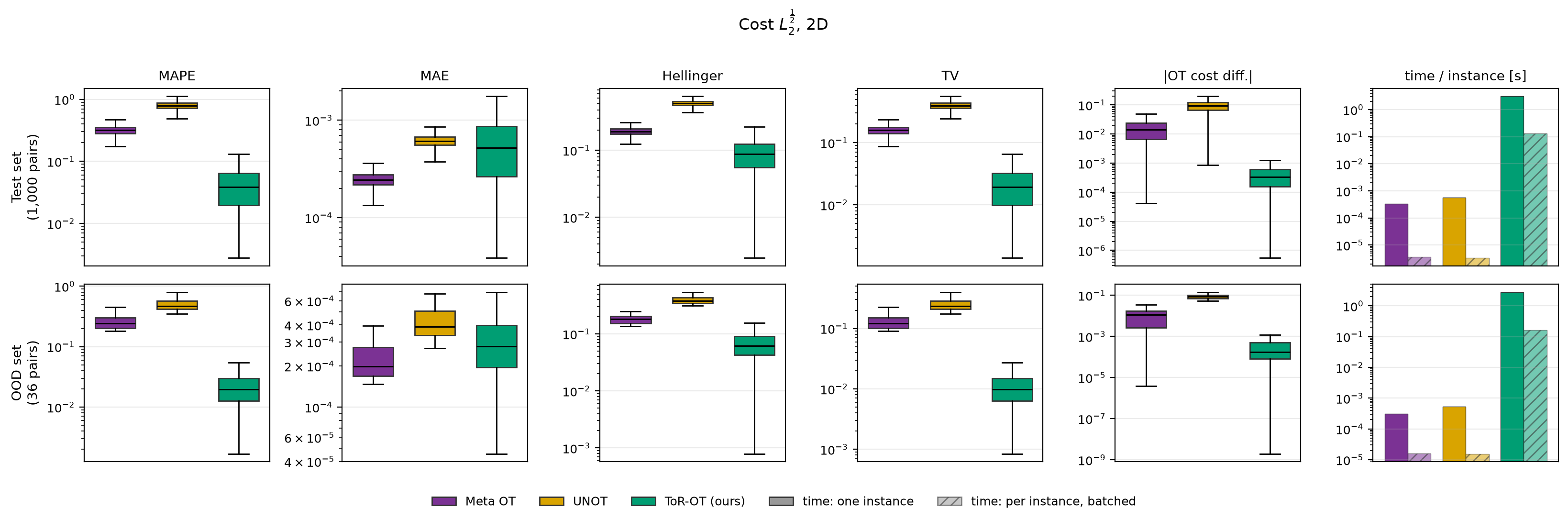}
    \caption{Cost $L_2^{1/2}$, 2D: test set (top, 1{,}000 pairs) and OOD set (bottom, 36 pairs). Five plan metrics, each method against the solution of its own problem (box: quartiles; line: median; whiskers: 1.5 IQR; outliers not shown; log scale), and time per instance on one A100 GPU (solid: one instance; hatched: per instance, batched).}
    \label{fig:appF-summary-sqrt-euclidean}
\end{figure}

Cost $c(x, y) = \lVert x - y \rVert_2^{1/2}$ (concave); Meta OT and UNOT use $\varepsilon = 0.01$. ToR-OT's mean MAPE is 0.049 on the test set and 0.032 on the OOD set, against 0.320 and 0.225 for Meta OT, the strongest baseline; ToR-OT is better on 30 of the 36 OOD pairs (Fig.~\ref{fig:appF-summary-sqrt-euclidean}).

\begin{figure}[!ht]
    \centering
    \includegraphics[width=0.69\linewidth]{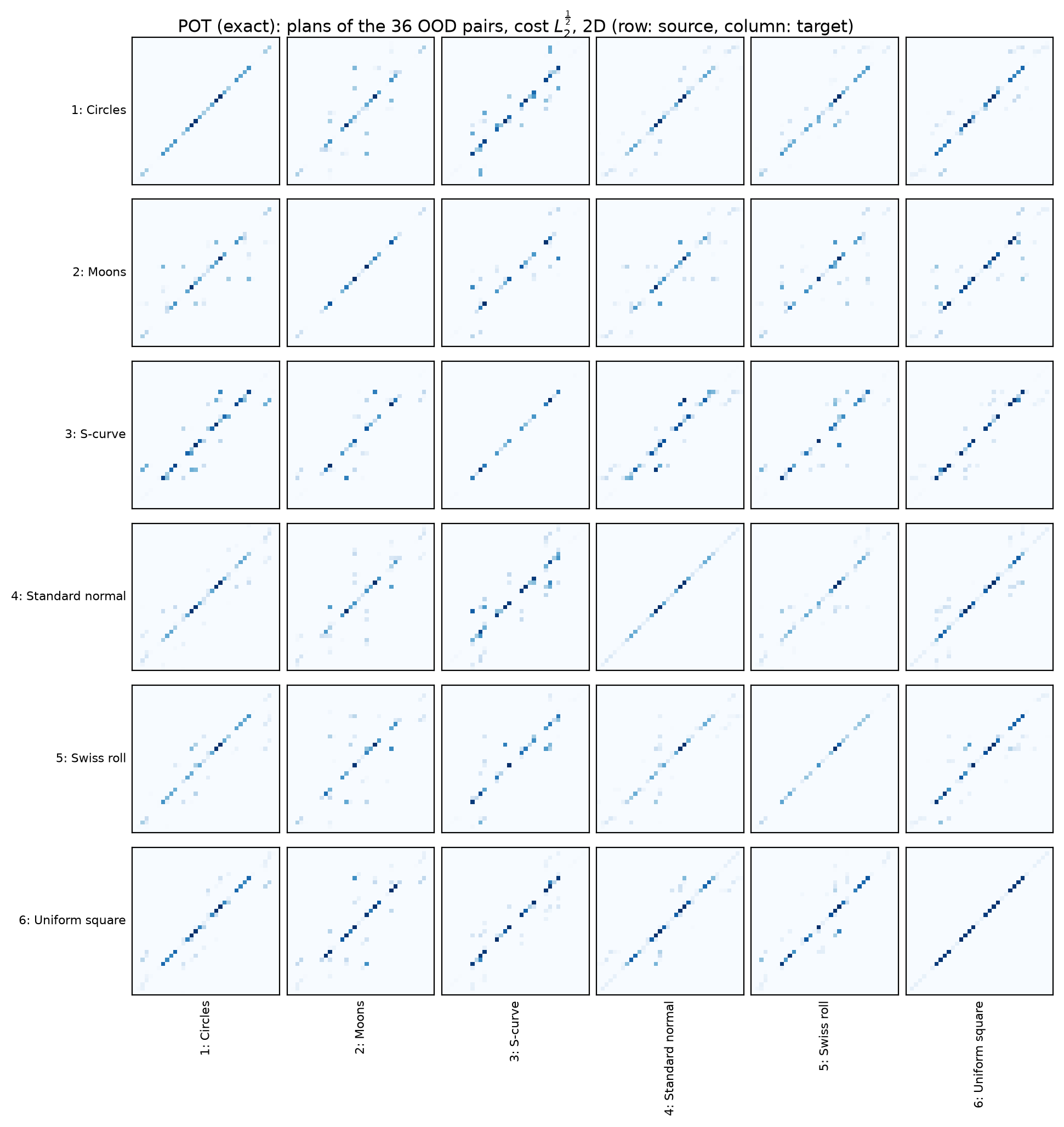}
    \caption{POT (exact): plans of the 36 OOD pairs, cost $L_2^{1/2}$, 2D.}
    \label{fig:appF-pairs-sqrt-euclidean-exact}
\end{figure}

\begin{figure}[p]
    \centering
    \includegraphics[width=0.69\linewidth]{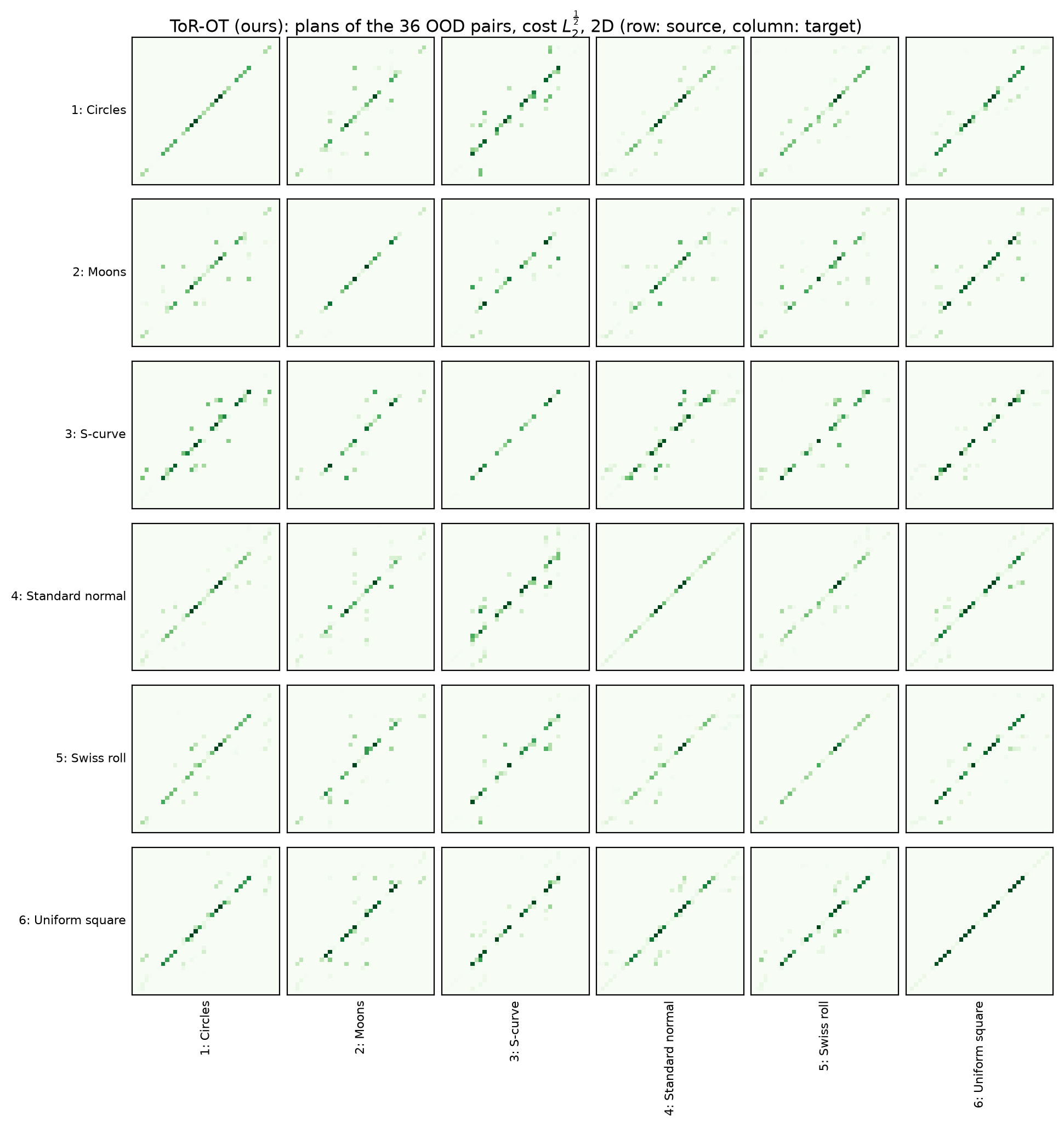}
    \caption{ToR-OT (ours): plans of the 36 OOD pairs, cost $L_2^{1/2}$, 2D.}
    \label{fig:appF-pairs-sqrt-euclidean-v2row_lastrow}
    \vspace{0.5em}
    \includegraphics[width=0.69\linewidth]{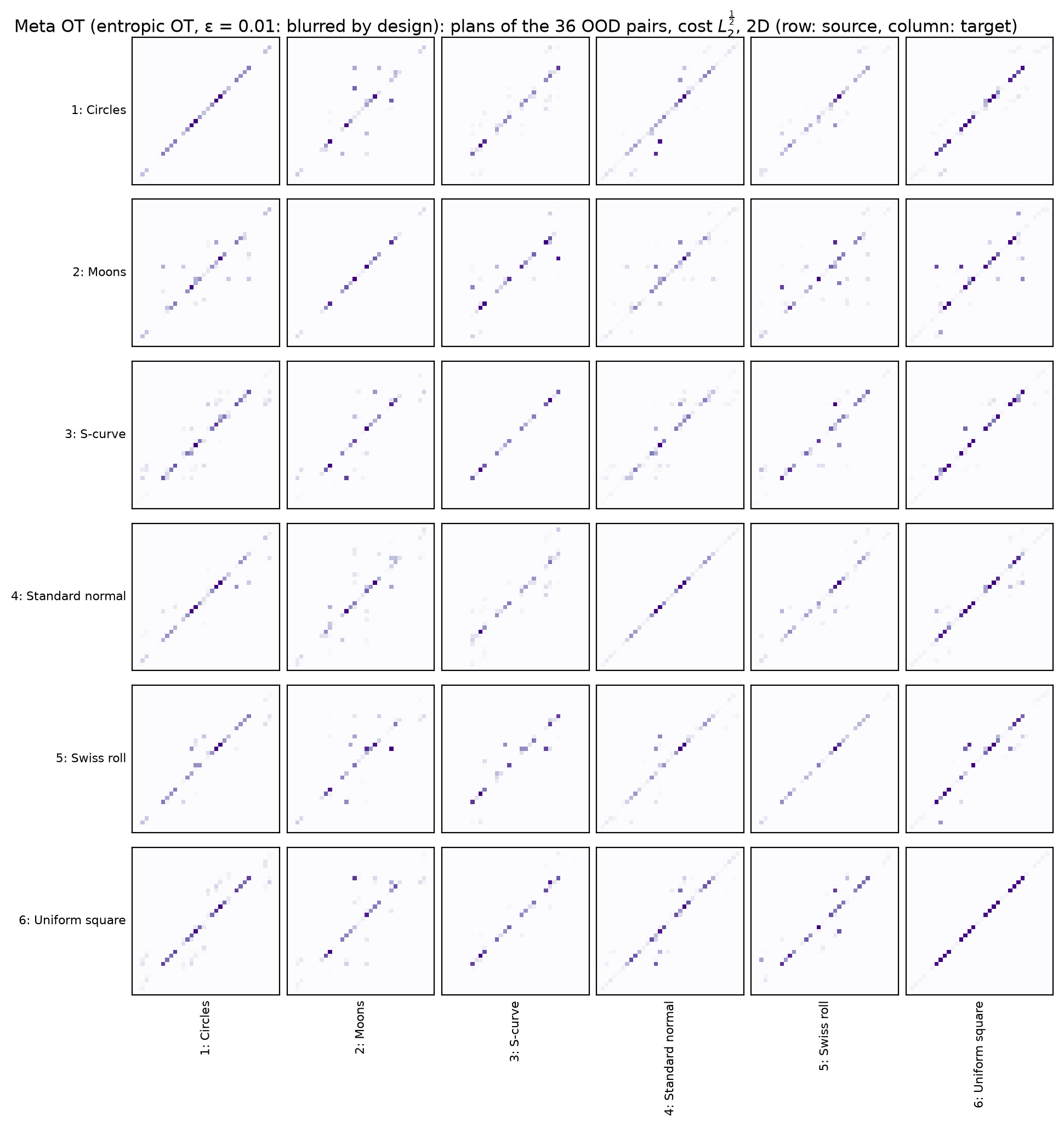}
    \caption{Meta OT (entropic, $\varepsilon = 0.01$): plans of the 36 OOD pairs, cost $L_2^{1/2}$, 2D.}
    \label{fig:appF-pairs-sqrt-euclidean-metaot}
\end{figure}

\begin{figure}[p]
    \centering
    \includegraphics[width=0.69\linewidth]{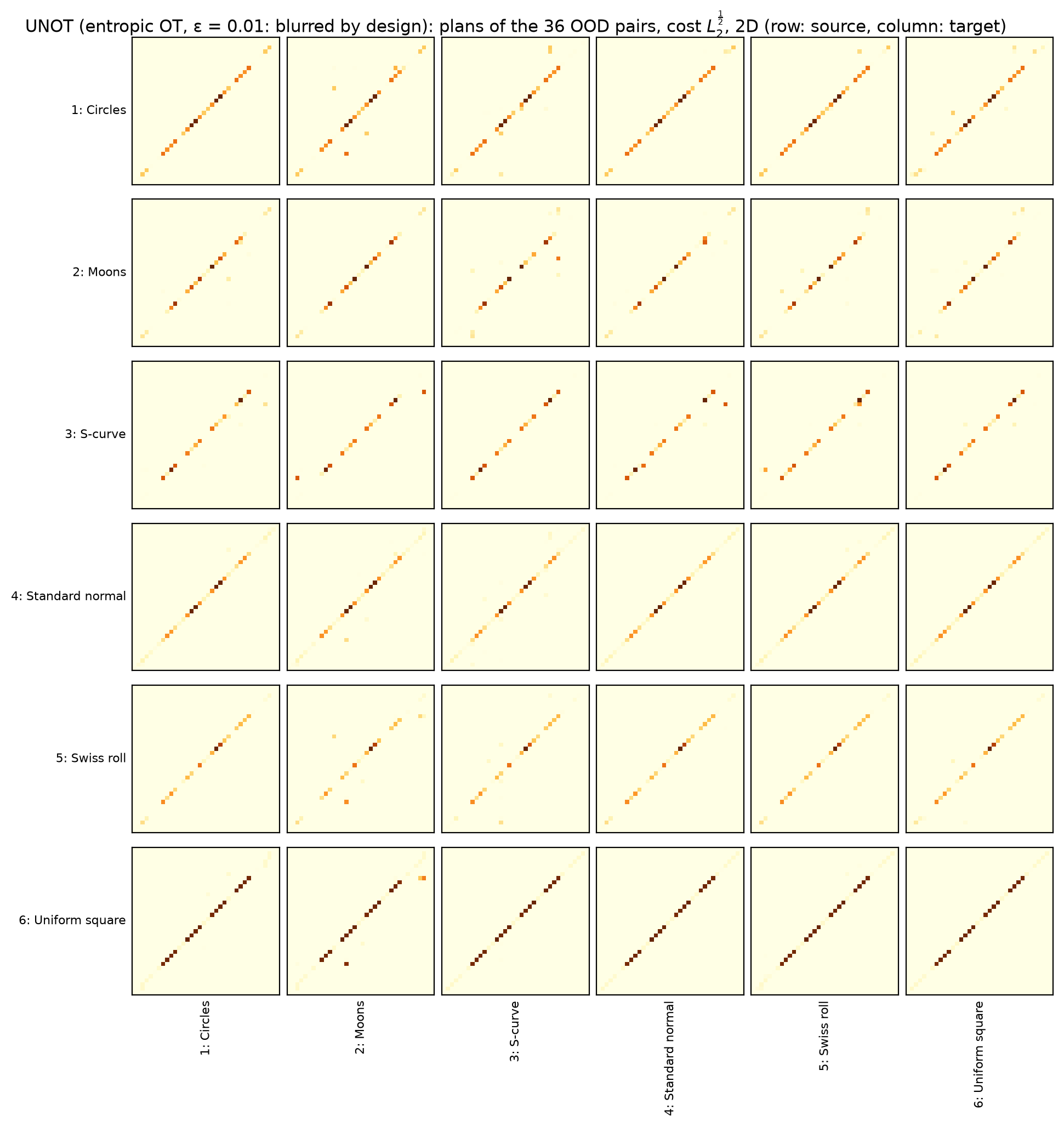}
    \caption{UNOT (entropic, $\varepsilon = 0.01$): plans of the 36 OOD pairs, cost $L_2^{1/2}$, 2D.}
    \label{fig:appF-pairs-sqrt-euclidean-unot}
\end{figure}

\clearpage
\subsection{Cost type: $L_2^2$}
\label{app:full-2d-sqeuclidean}

\begin{figure}[!ht]
    \centering
    \includegraphics[width=\linewidth]{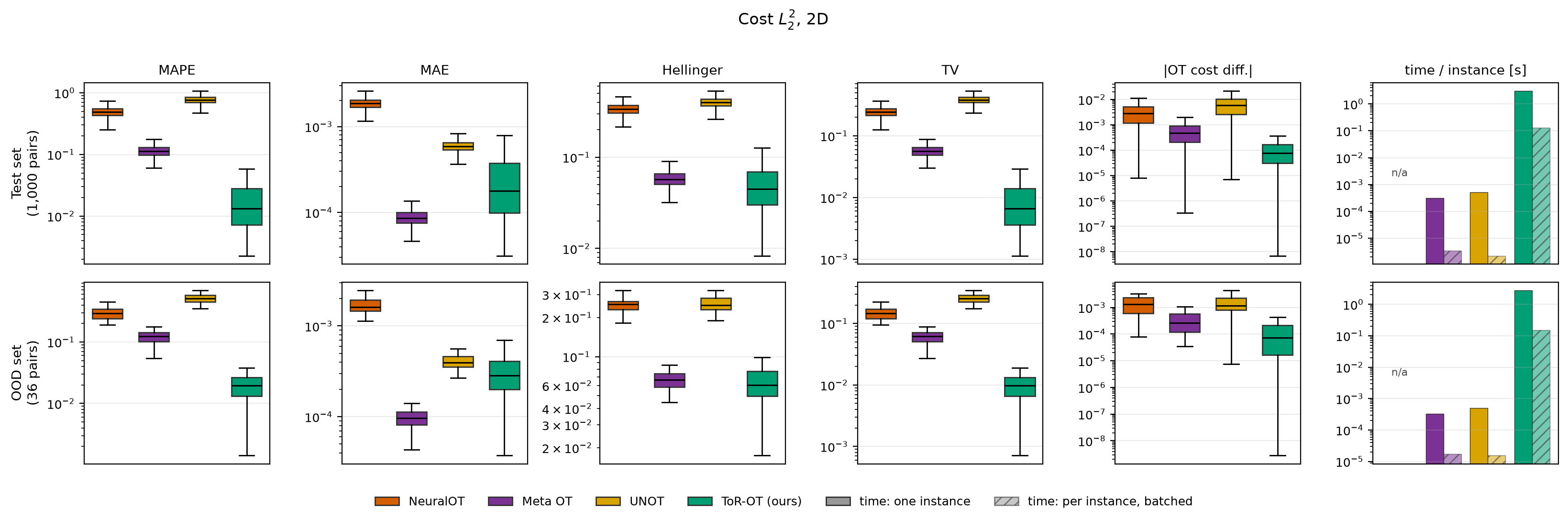}
    \caption{Cost $L_2^2$, 2D: test set (top, 1{,}000 pairs) and OOD set (bottom, 36 pairs). Five plan metrics, each method against the solution of its own problem (box: quartiles; line: median; whiskers: 1.5 IQR; outliers not shown; log scale), and time per instance on one A100 GPU (solid: one instance; hatched: per instance, batched). NeuralOT optimizes each instance separately and its runs shared GPUs, so its times are not comparable (n/a).}
    \label{fig:appF-summary-sqeuclidean}
\end{figure}

Cost $c(x, y) = \lVert x - y \rVert_2^2$; Meta OT and UNOT use $\varepsilon = 0.01$. ToR-OT's mean MAPE is 0.023 on the test set and 0.029 on the OOD set, against 0.115 and 0.109 for Meta OT, the strongest baseline; ToR-OT is better on 34 of the 36 OOD pairs (Fig.~\ref{fig:appF-summary-sqeuclidean}).

\begin{figure}[!ht]
    \centering
    \includegraphics[width=0.69\linewidth]{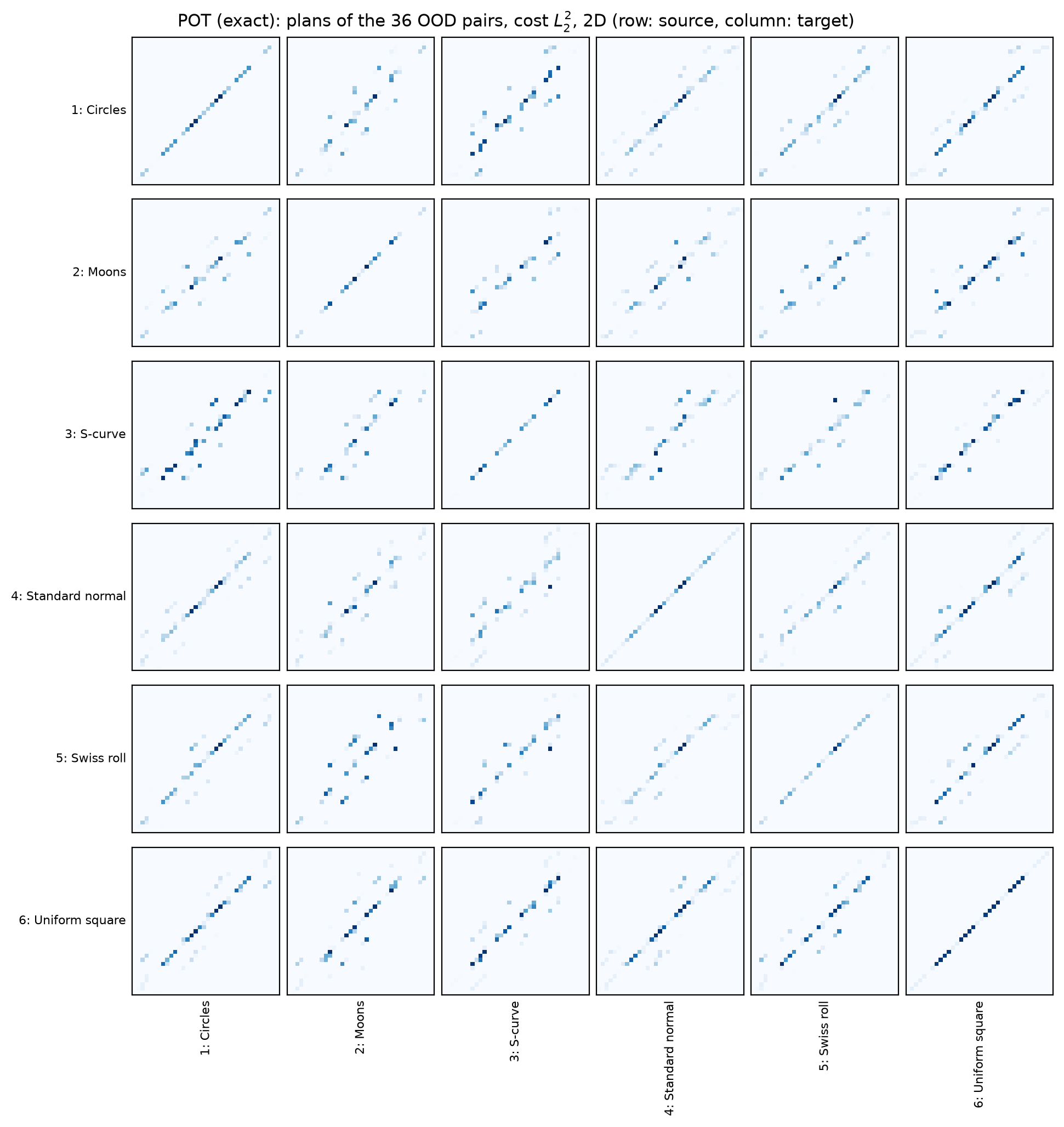}
    \caption{POT (exact): plans of the 36 OOD pairs, cost $L_2^2$, 2D.}
    \label{fig:appF-pairs-sqeuclidean-exact}
\end{figure}

\begin{figure}[p]
    \centering
    \includegraphics[width=0.69\linewidth]{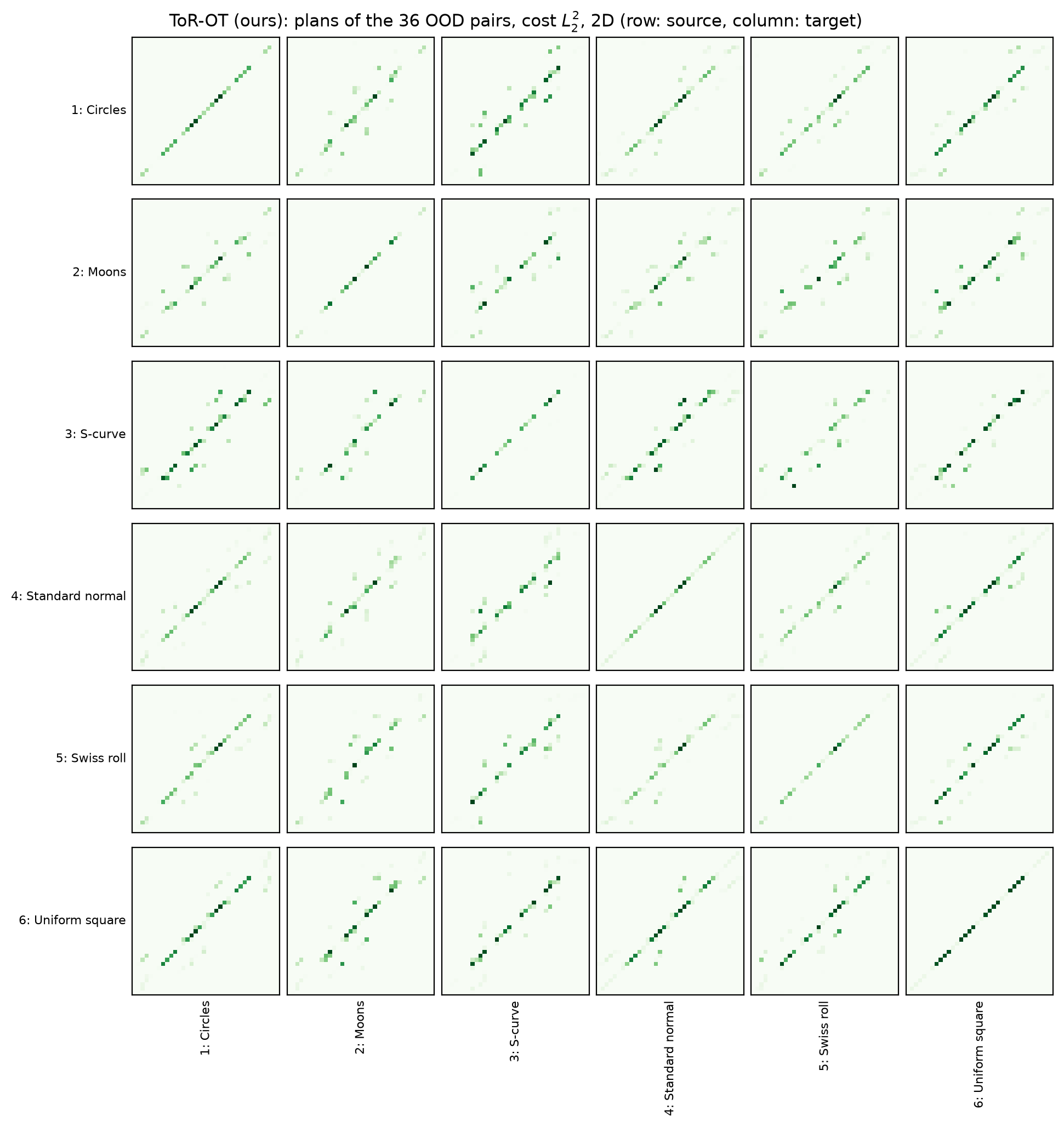}
    \caption{ToR-OT (ours): plans of the 36 OOD pairs, cost $L_2^2$, 2D.}
    \label{fig:appF-pairs-sqeuclidean-v2row_lastrow}
    \vspace{0.5em}
    \includegraphics[width=0.69\linewidth]{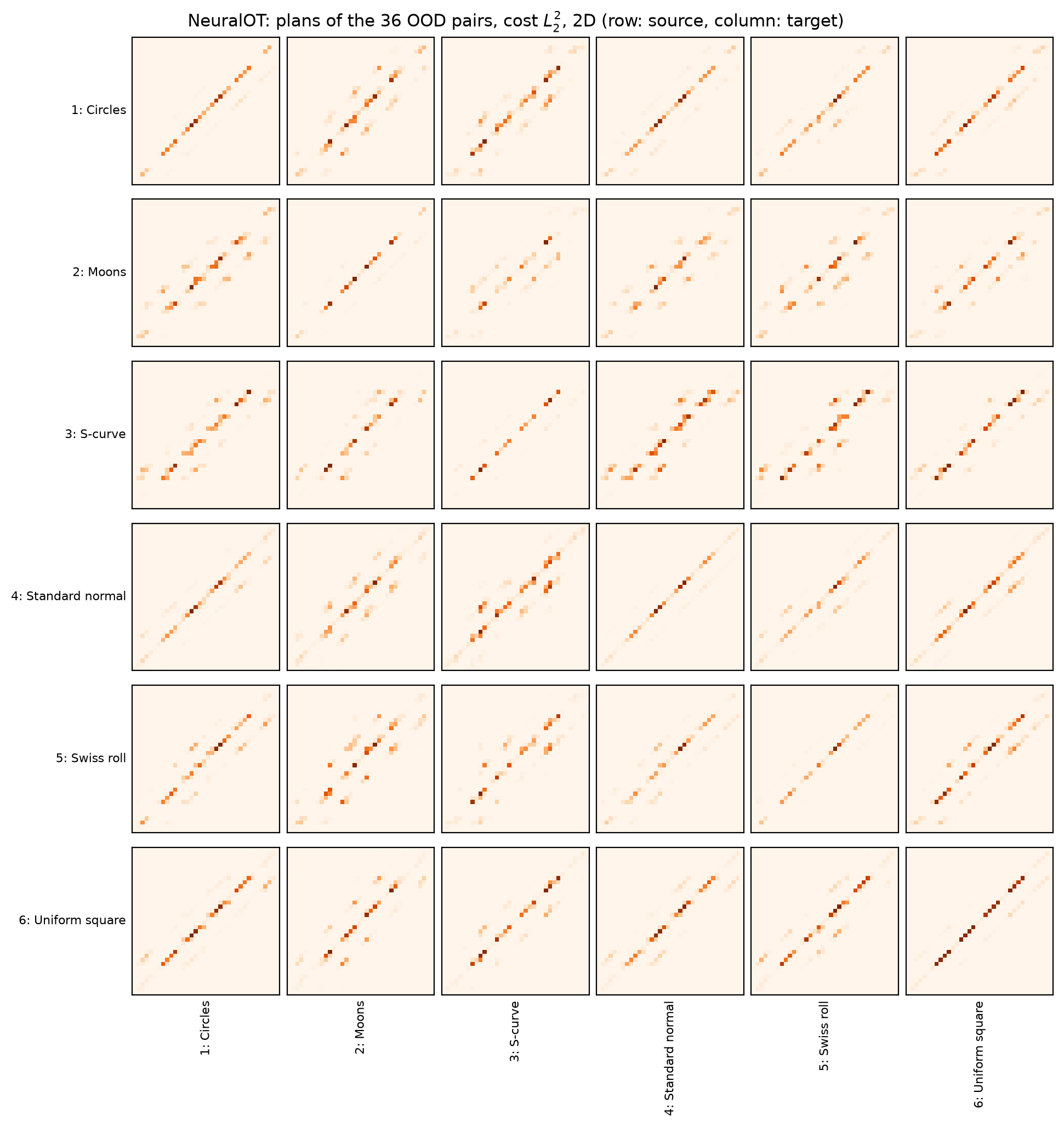}
    \caption{NeuralOT: plans of the 36 OOD pairs, cost $L_2^2$, 2D.}
    \label{fig:appF-pairs-sqeuclidean-neuralot}
\end{figure}

\begin{figure}[p]
    \centering
    \includegraphics[width=0.69\linewidth]{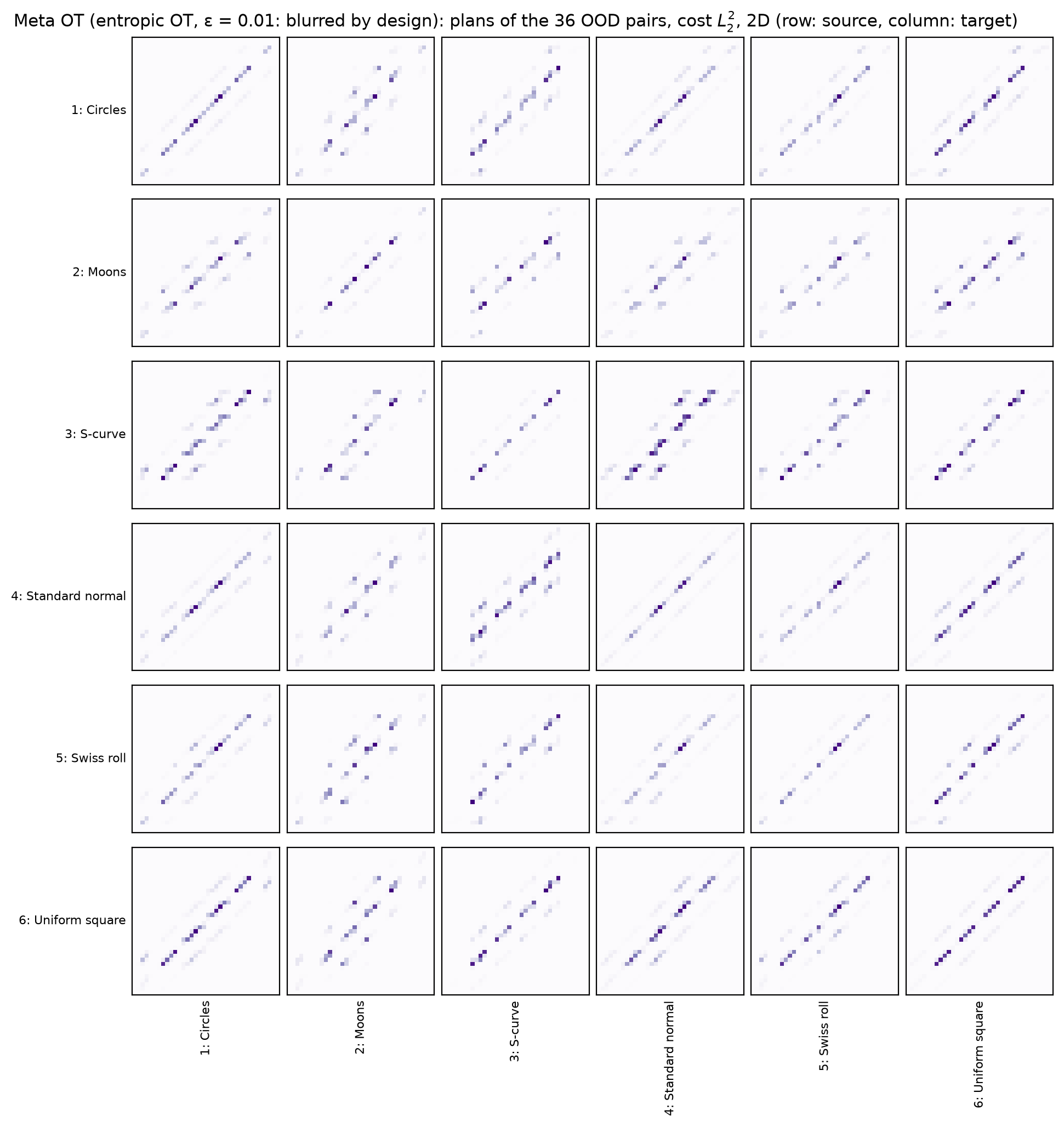}
    \caption{Meta OT (entropic, $\varepsilon = 0.01$): plans of the 36 OOD pairs, cost $L_2^2$, 2D.}
    \label{fig:appF-pairs-sqeuclidean-metaot}
    \vspace{0.5em}
    \includegraphics[width=0.69\linewidth]{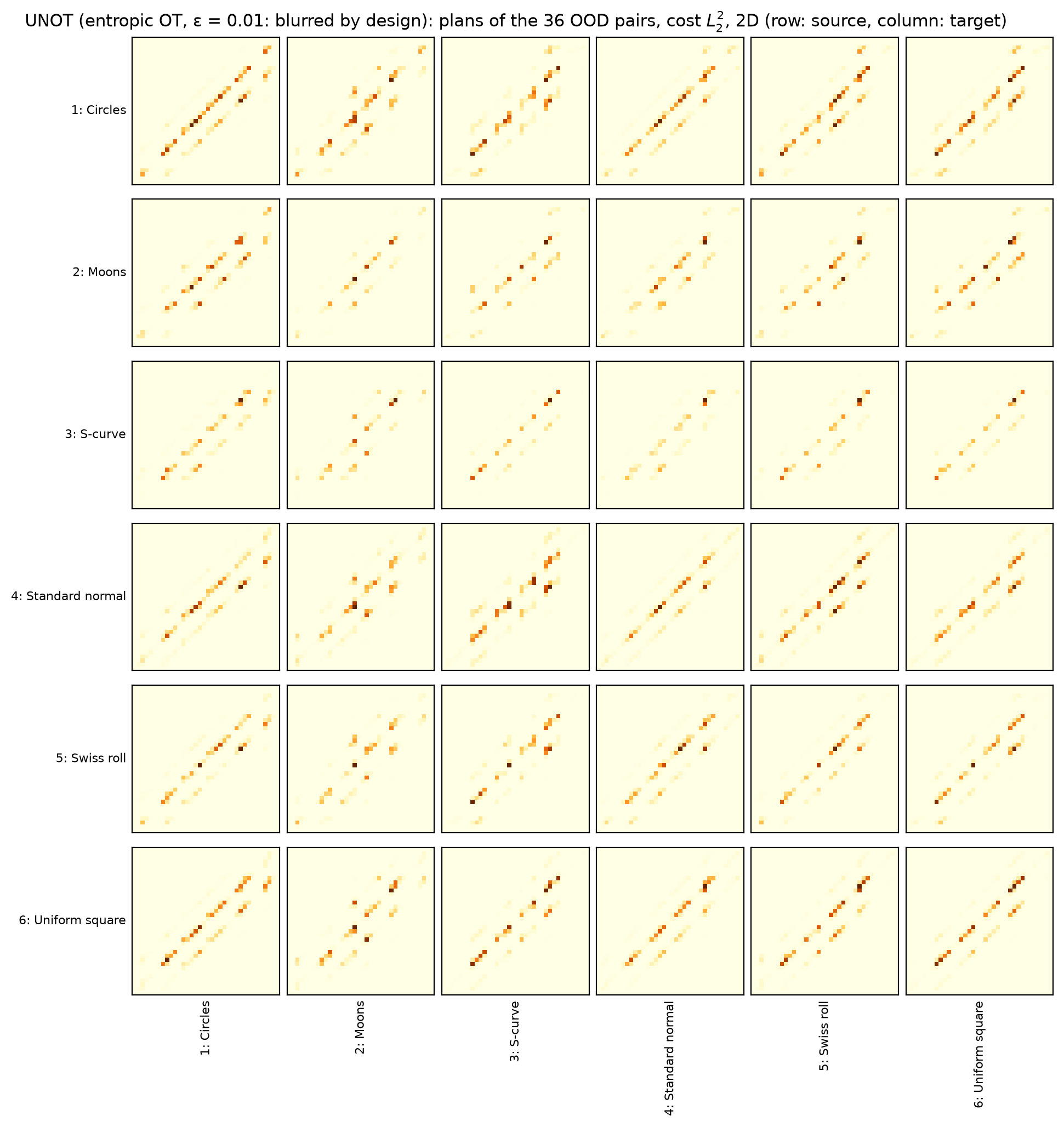}
    \caption{UNOT (entropic, $\varepsilon = 0.01$): plans of the 36 OOD pairs, cost $L_2^2$, 2D.}
    \label{fig:appF-pairs-sqeuclidean-unot}
\end{figure}

\clearpage
\subsection{Cost type: $L_2^3$}
\label{app:full-2d-cubic-euclidean}

\begin{figure}[!ht]
    \centering
    \includegraphics[width=\linewidth]{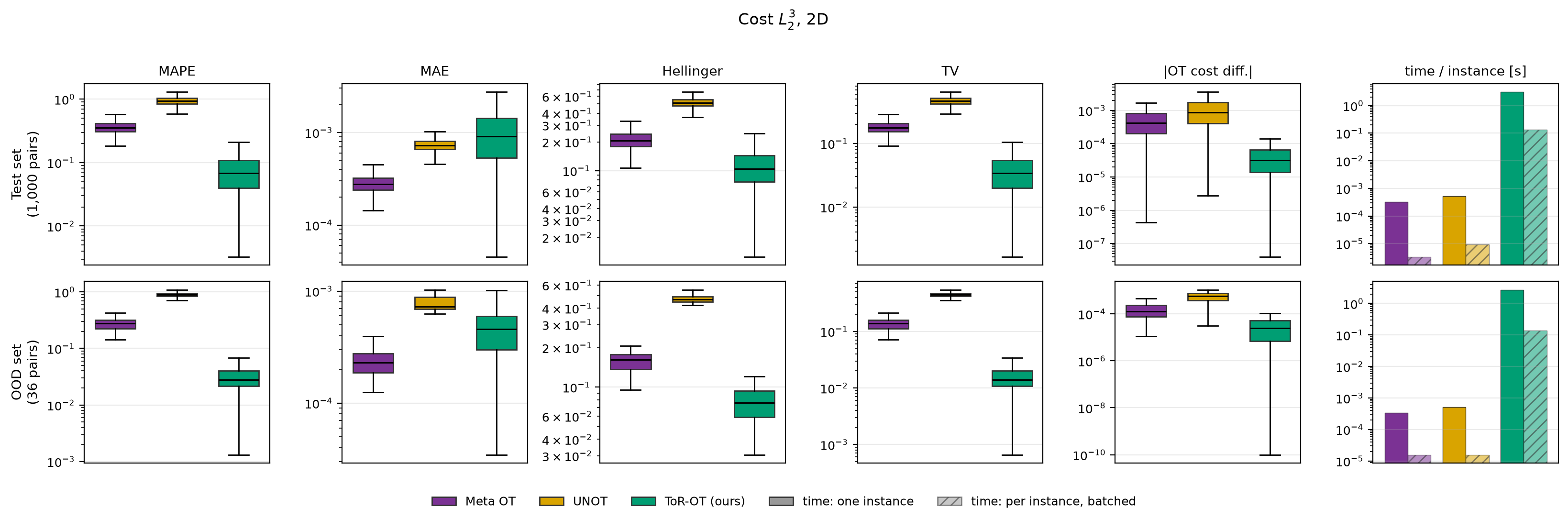}
    \caption{Cost $L_2^3$, 2D: test set (top, 1{,}000 pairs) and OOD set (bottom, 36 pairs). Five plan metrics, each method against the solution of its own problem (box: quartiles; line: median; whiskers: 1.5 IQR; outliers not shown; log scale), and time per instance on one A100 GPU (solid: one instance; hatched: per instance, batched).}
    \label{fig:appF-summary-cubic-euclidean}
\end{figure}

Cost $c(x, y) = \lVert x - y \rVert_2^3$; Meta OT and UNOT use $\varepsilon = 0.001$. ToR-OT's mean MAPE is 0.080 on the test set and 0.043 on the OOD set, against 0.369 and 0.256 for Meta OT, the strongest baseline; ToR-OT is better on 35 of the 36 OOD pairs (Fig.~\ref{fig:appF-summary-cubic-euclidean}).

\begin{figure}[!ht]
    \centering
    \includegraphics[width=0.69\linewidth]{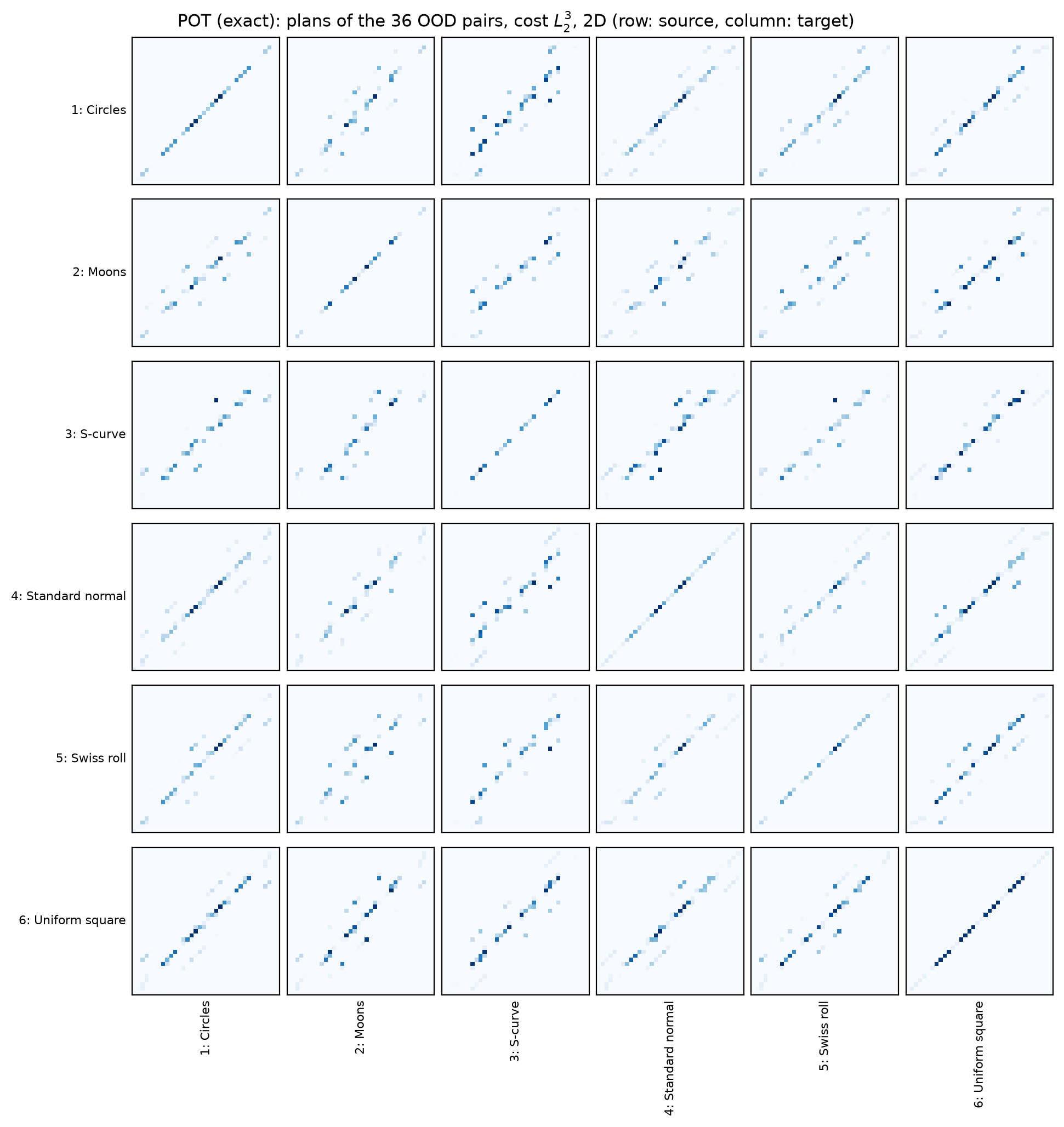}
    \caption{POT (exact): plans of the 36 OOD pairs, cost $L_2^3$, 2D.}
    \label{fig:appF-pairs-cubic-euclidean-exact}
\end{figure}

\begin{figure}[p]
    \centering
    \includegraphics[width=0.69\linewidth]{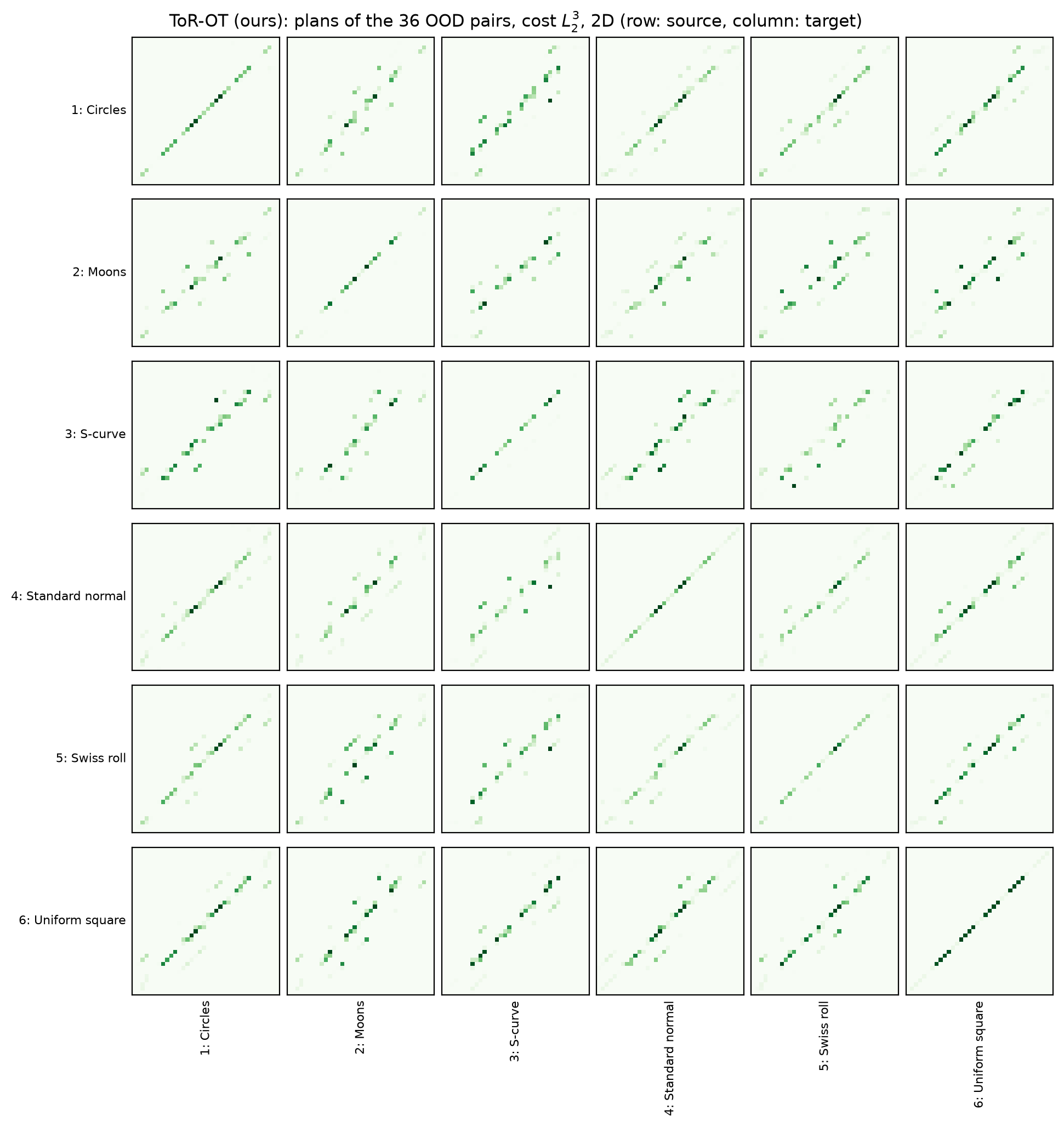}
    \caption{ToR-OT (ours): plans of the 36 OOD pairs, cost $L_2^3$, 2D.}
    \label{fig:appF-pairs-cubic-euclidean-v2row_lastrow}
    \vspace{0.5em}
    \includegraphics[width=0.69\linewidth]{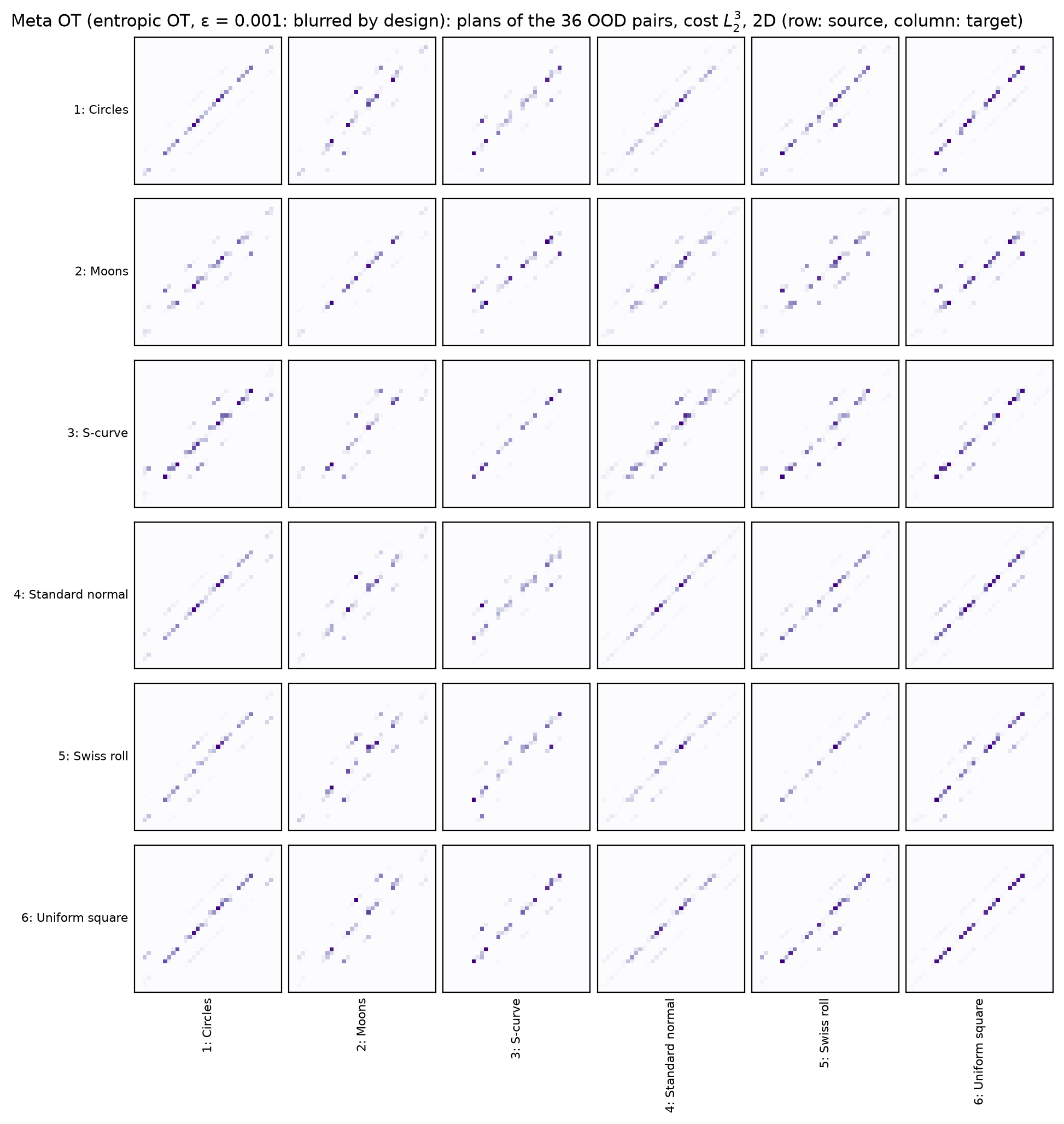}
    \caption{Meta OT (entropic, $\varepsilon = 0.001$): plans of the 36 OOD pairs, cost $L_2^3$, 2D.}
    \label{fig:appF-pairs-cubic-euclidean-metaot}
\end{figure}

\begin{figure}[p]
    \centering
    \includegraphics[width=0.69\linewidth]{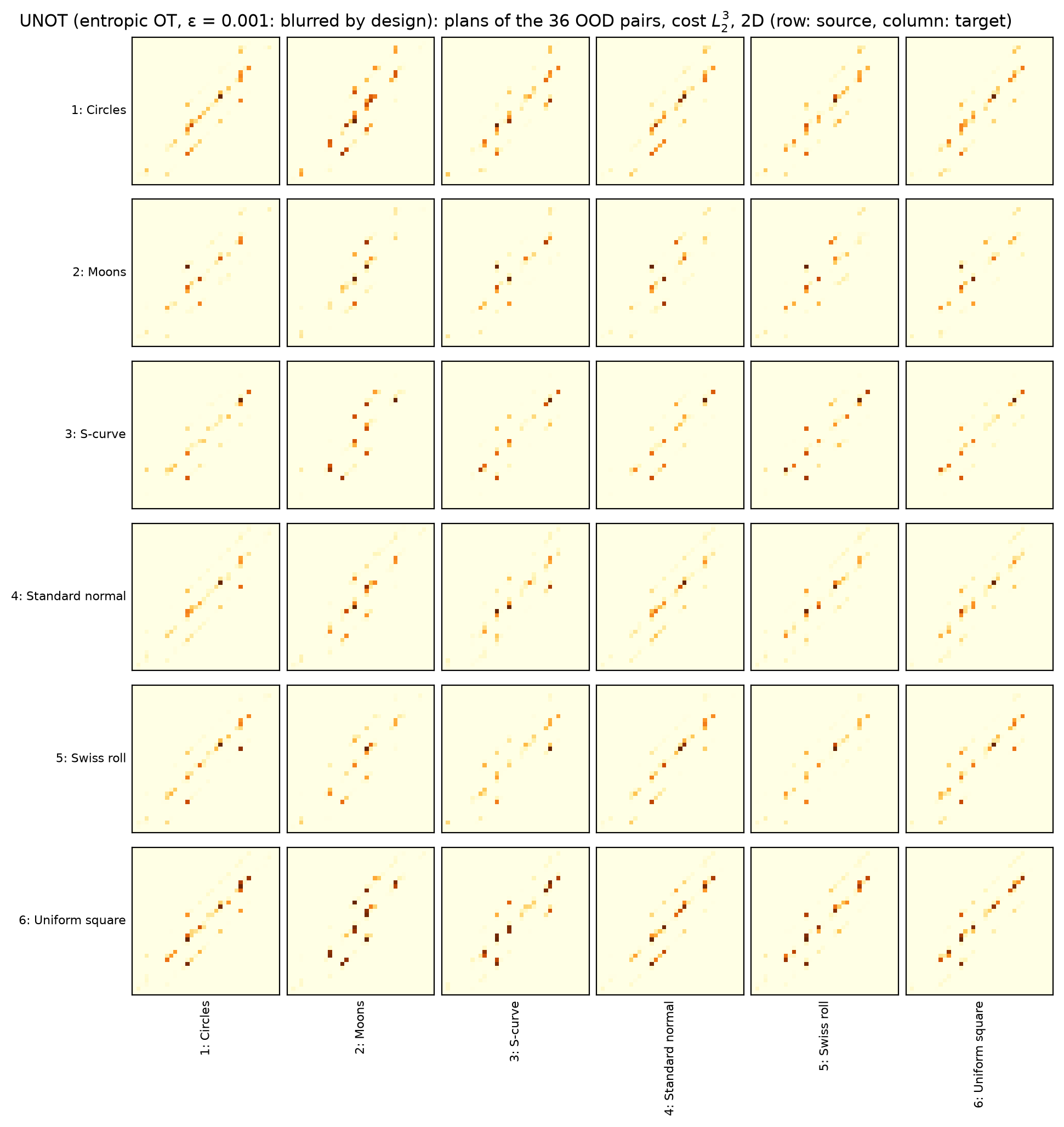}
    \caption{UNOT (entropic, $\varepsilon = 0.001$): plans of the 36 OOD pairs, cost $L_2^3$, 2D.}
    \label{fig:appF-pairs-cubic-euclidean-unot}
\end{figure}

\clearpage
\subsection{Cost type: Canberra}
\label{app:full-2d-canberra}

\begin{figure}[!ht]
    \centering
    \includegraphics[width=\linewidth]{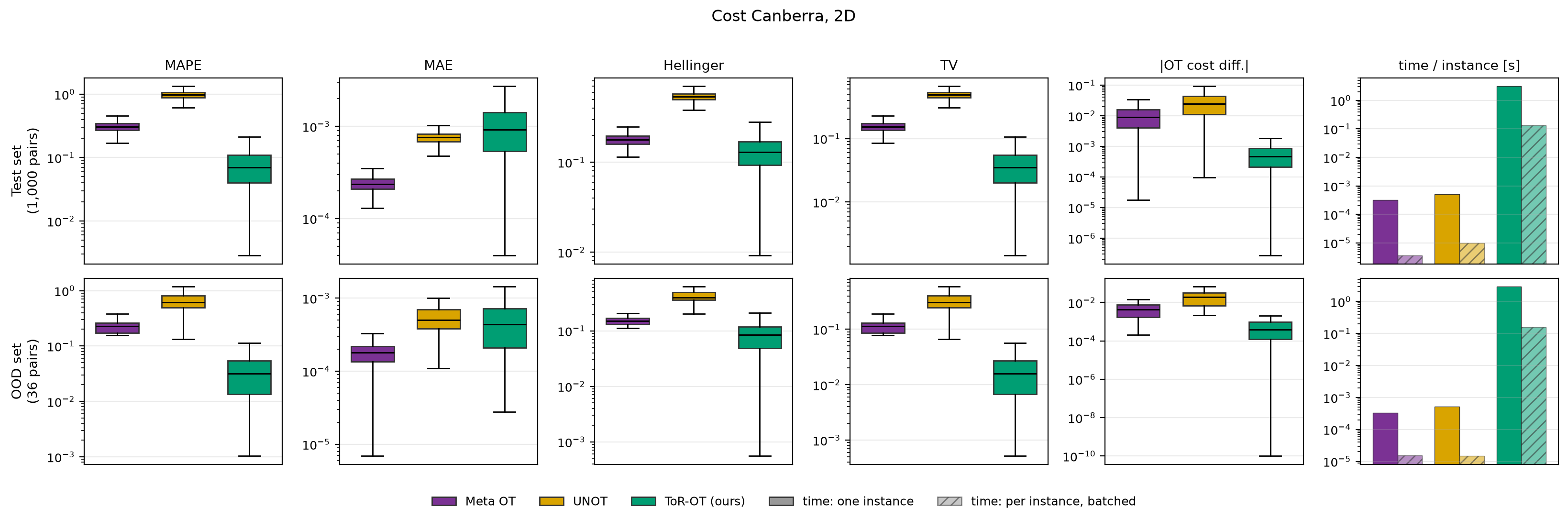}
    \caption{Cost Canberra, 2D: test set (top, 1{,}000 pairs) and OOD set (bottom, 36 pairs). Five plan metrics, each method against the solution of its own problem (box: quartiles; line: median; whiskers: 1.5 IQR; outliers not shown; log scale), and time per instance on one A100 GPU (solid: one instance; hatched: per instance, batched).}
    \label{fig:appF-summary-canberra}
\end{figure}

Cost $c(x, y) = \sum_d |x_d - y_d| / (|x_d| + |y_d|)$; Meta OT and UNOT use $\varepsilon = 0.01$. ToR-OT's mean MAPE is 0.081 on the test set and 0.047 on the OOD set, against 0.310 and 0.204 for Meta OT, the strongest baseline; ToR-OT is better on 34 of the 36 OOD pairs (Fig.~\ref{fig:appF-summary-canberra}).

\begin{figure}[!ht]
    \centering
    \includegraphics[width=0.69\linewidth]{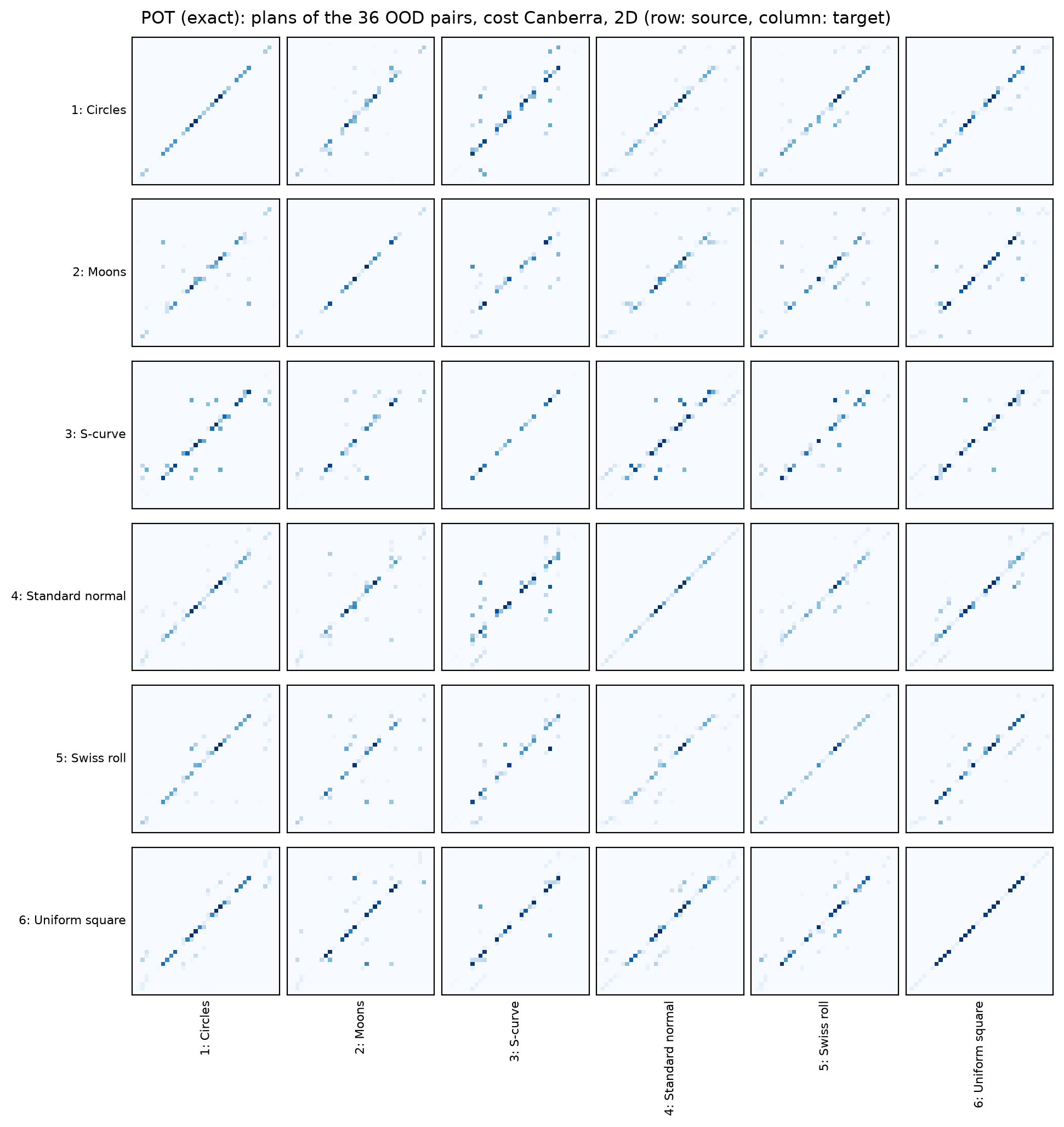}
    \caption{POT (exact): plans of the 36 OOD pairs, cost Canberra, 2D.}
    \label{fig:appF-pairs-canberra-exact}
\end{figure}

\begin{figure}[p]
    \centering
    \includegraphics[width=0.69\linewidth]{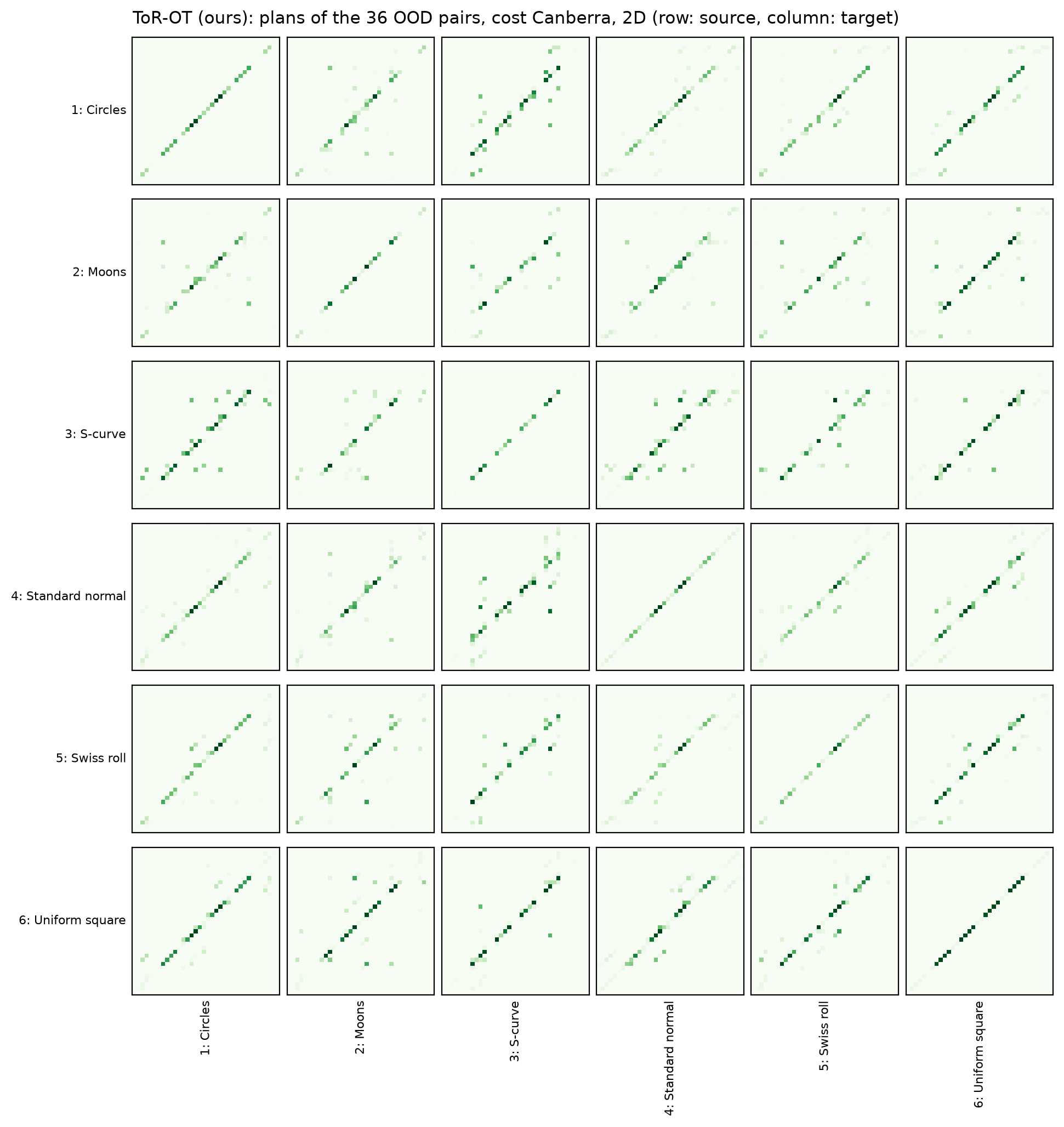}
    \caption{ToR-OT (ours): plans of the 36 OOD pairs, cost Canberra, 2D.}
    \label{fig:appF-pairs-canberra-v2row_lastrow}
    \vspace{0.5em}
    \includegraphics[width=0.69\linewidth]{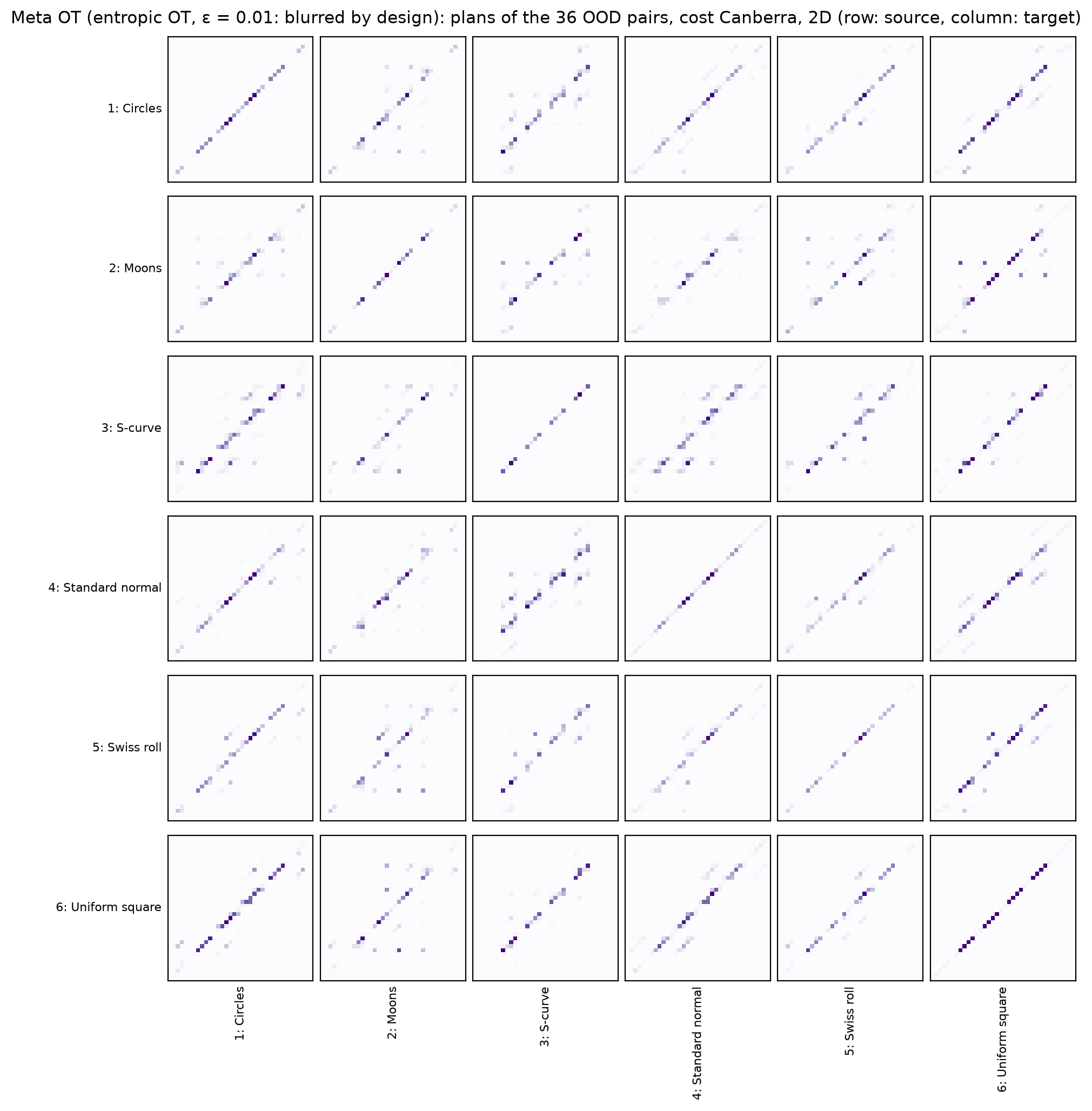}
    \caption{Meta OT (entropic, $\varepsilon = 0.01$): plans of the 36 OOD pairs, cost Canberra, 2D.}
    \label{fig:appF-pairs-canberra-metaot}
\end{figure}

\begin{figure}[p]
    \centering
    \includegraphics[width=0.69\linewidth]{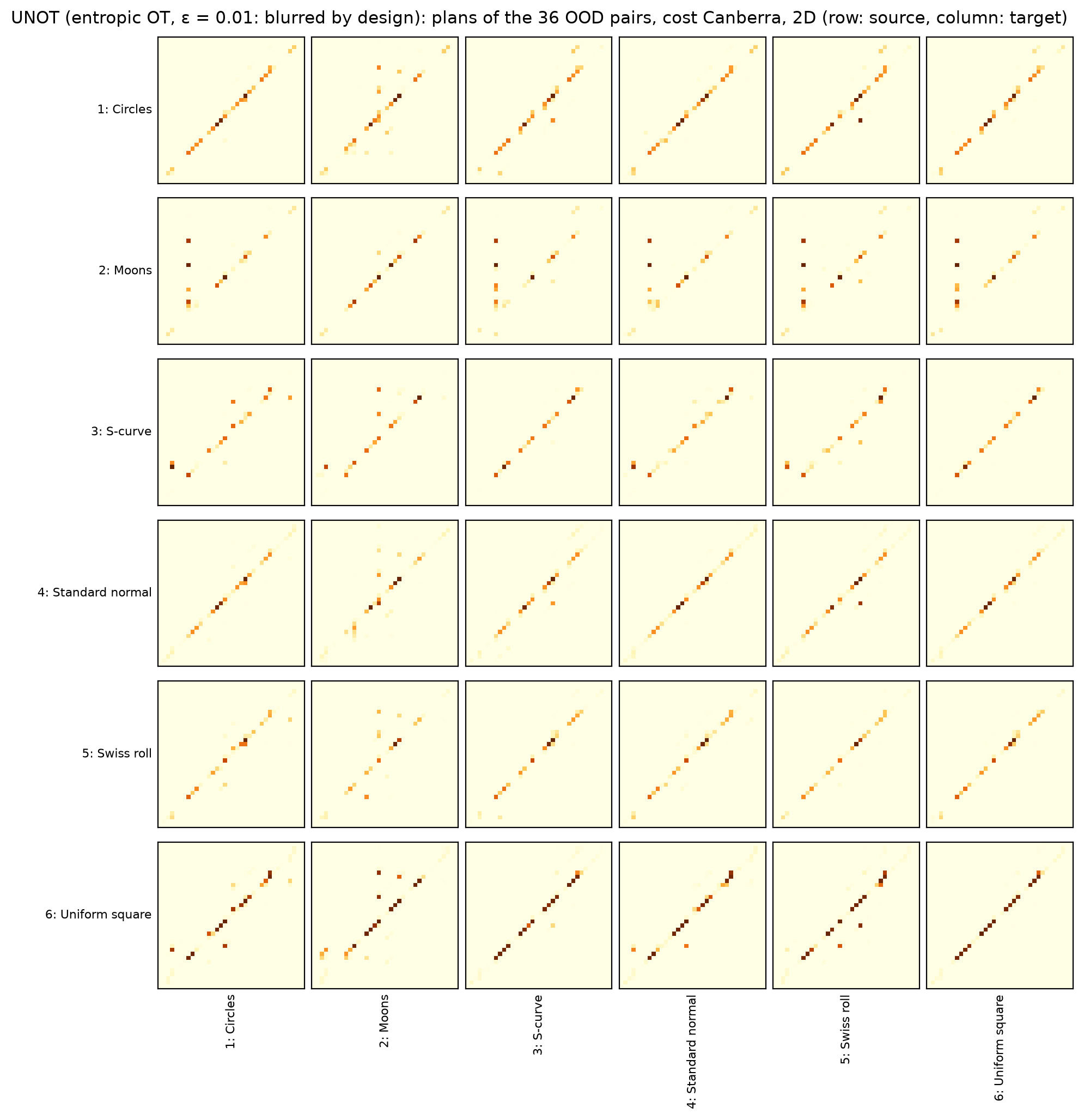}
    \caption{UNOT (entropic, $\varepsilon = 0.01$): plans of the 36 OOD pairs, cost Canberra, 2D.}
    \label{fig:appF-pairs-canberra-unot}
\end{figure}

\clearpage
\subsection{Cost type: Gaussian}
\label{app:full-2d-gaussian}

\begin{figure}[!ht]
    \centering
    \includegraphics[width=\linewidth]{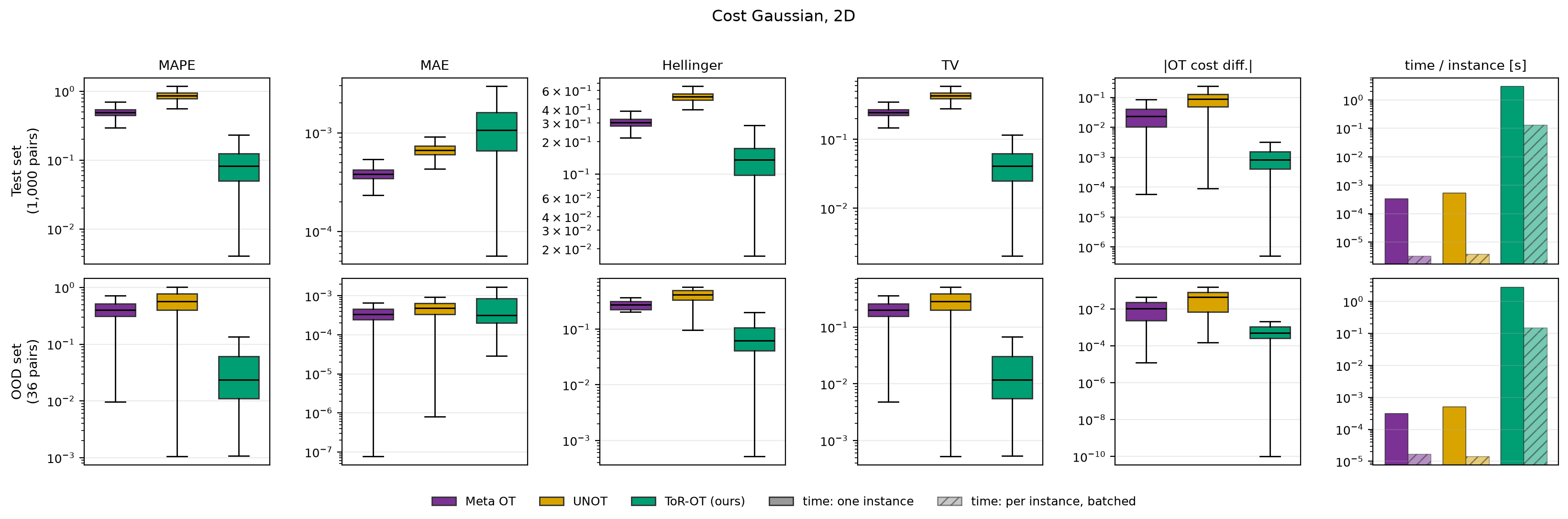}
    \caption{Cost Gaussian, 2D: test set (top, 1{,}000 pairs) and OOD set (bottom, 36 pairs). Five plan metrics, each method against the solution of its own problem (box: quartiles; line: median; whiskers: 1.5 IQR; outliers not shown; log scale), and time per instance on one A100 GPU (solid: one instance; hatched: per instance, batched).}
    \label{fig:appF-summary-gaussian}
\end{figure}

Cost $c(x, y) = 1 - \exp(-\lVert x - y \rVert_2^2 / 2\sigma^2)$, $\sigma = 0.2$; Meta OT and UNOT use $\varepsilon = 0.01$. ToR-OT's mean MAPE is 0.092 on the test set and 0.055 on the OOD set, against 0.500 and 0.371 for Meta OT, the strongest baseline; ToR-OT is better on 32 of the 36 OOD pairs (Fig.~\ref{fig:appF-summary-gaussian}).

\begin{figure}[!ht]
    \centering
    \includegraphics[width=0.69\linewidth]{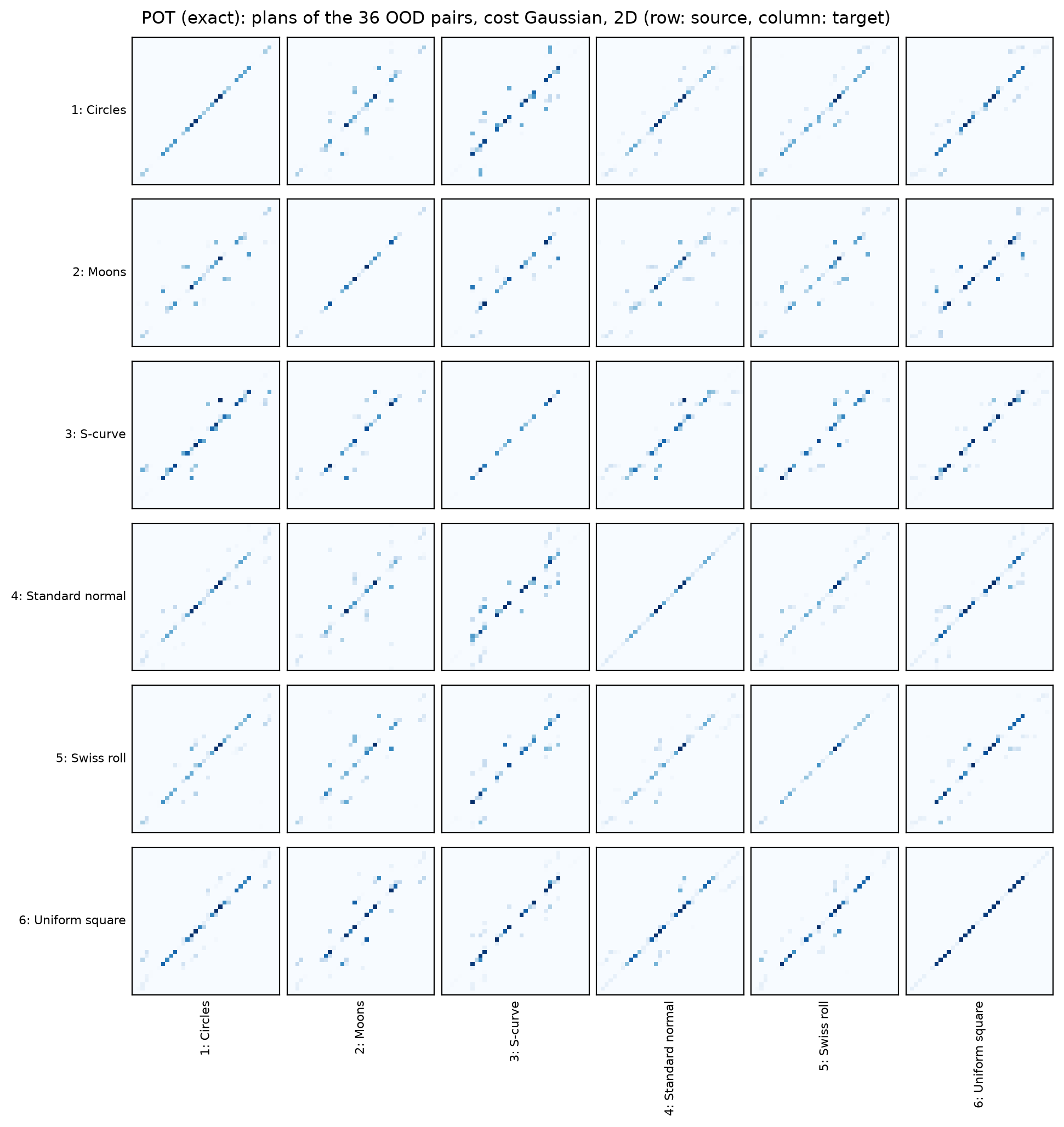}
    \caption{POT (exact): plans of the 36 OOD pairs, cost Gaussian, 2D.}
    \label{fig:appF-pairs-gaussian-exact}
\end{figure}

\begin{figure}[p]
    \centering
    \includegraphics[width=0.69\linewidth]{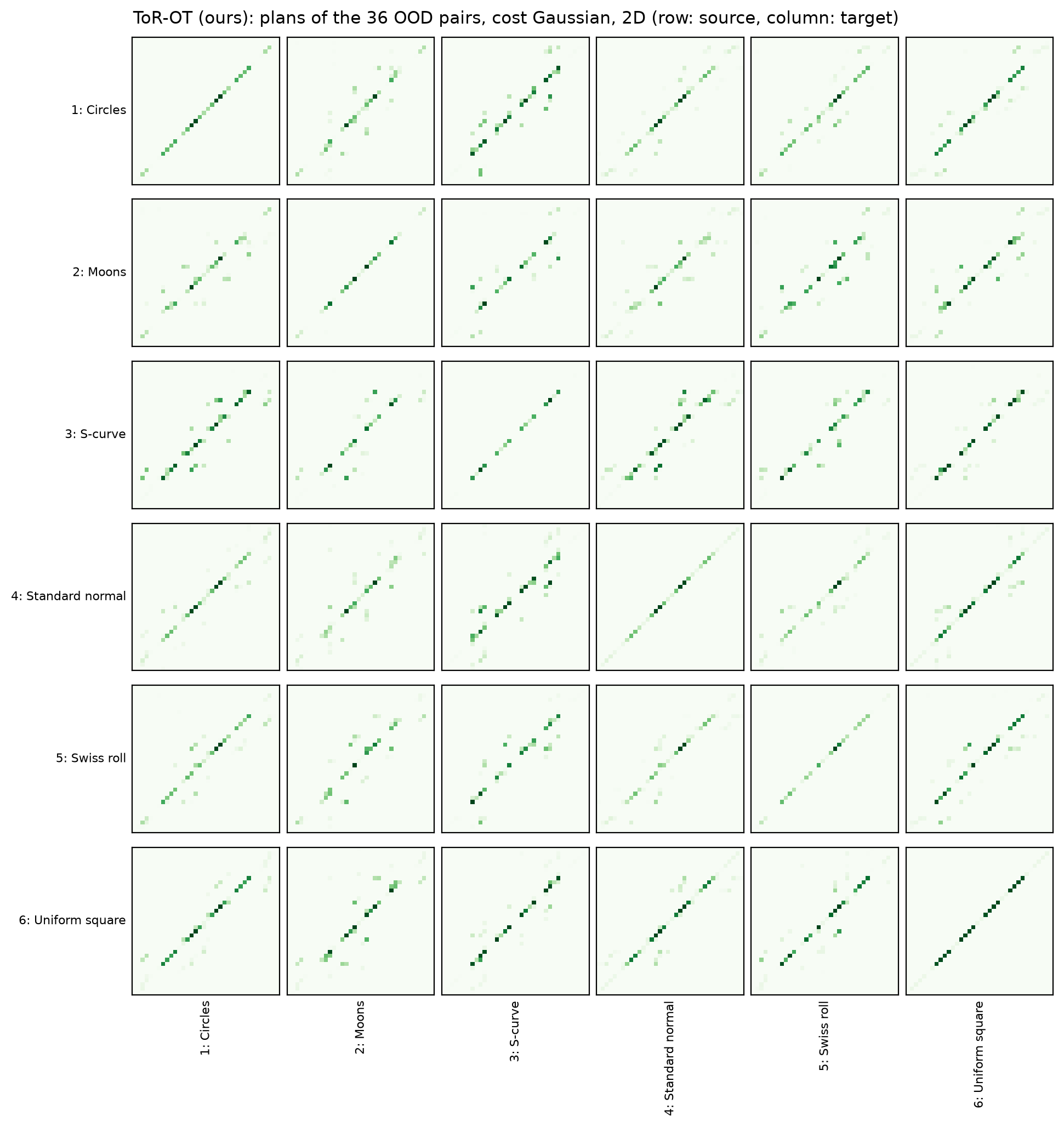}
    \caption{ToR-OT (ours): plans of the 36 OOD pairs, cost Gaussian, 2D.}
    \label{fig:appF-pairs-gaussian-v2row_lastrow}
    \vspace{0.5em}
    \includegraphics[width=0.69\linewidth]{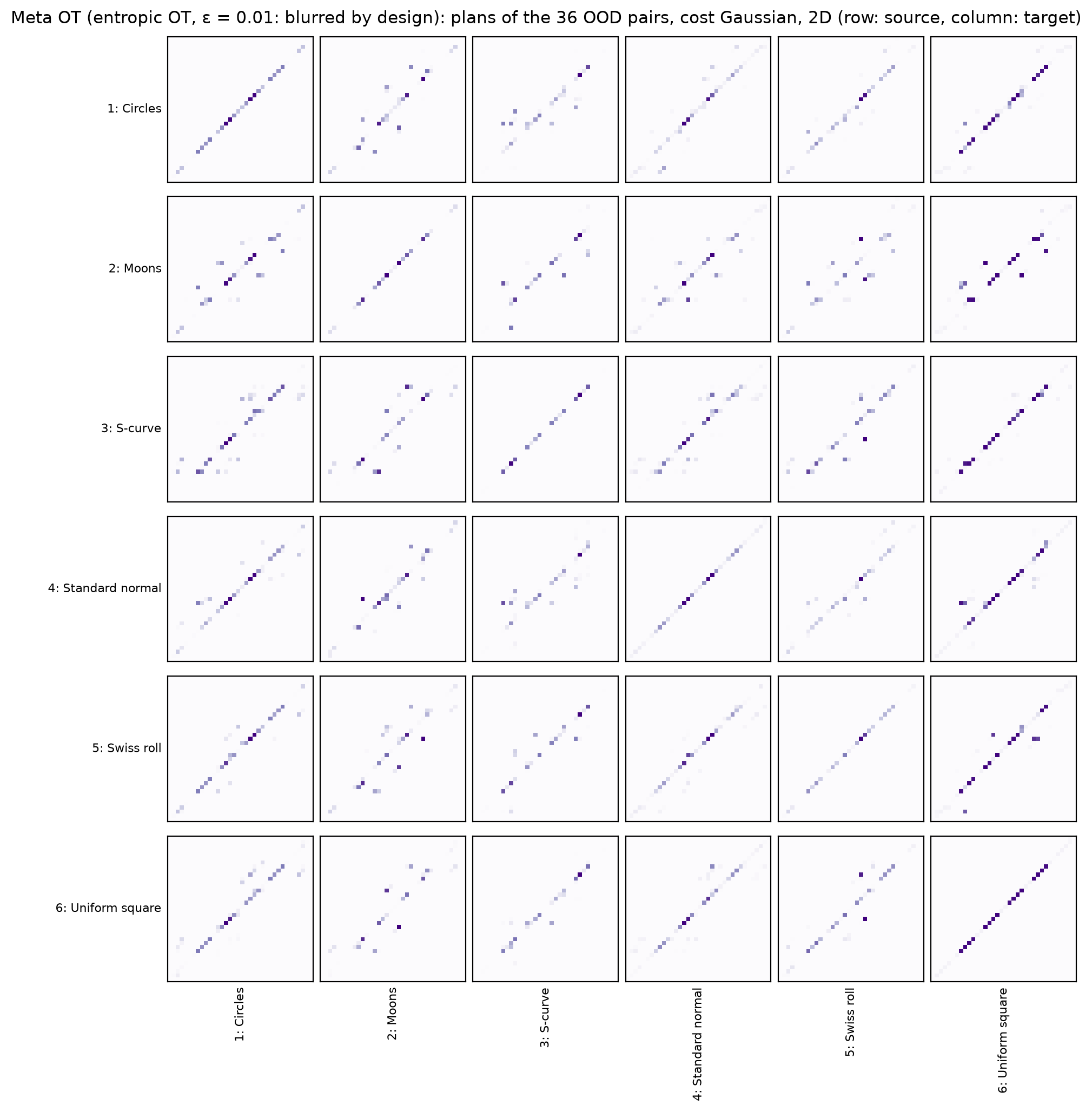}
    \caption{Meta OT (entropic, $\varepsilon = 0.01$): plans of the 36 OOD pairs, cost Gaussian, 2D.}
    \label{fig:appF-pairs-gaussian-metaot}
\end{figure}

\begin{figure}[p]
    \centering
    \includegraphics[width=0.69\linewidth]{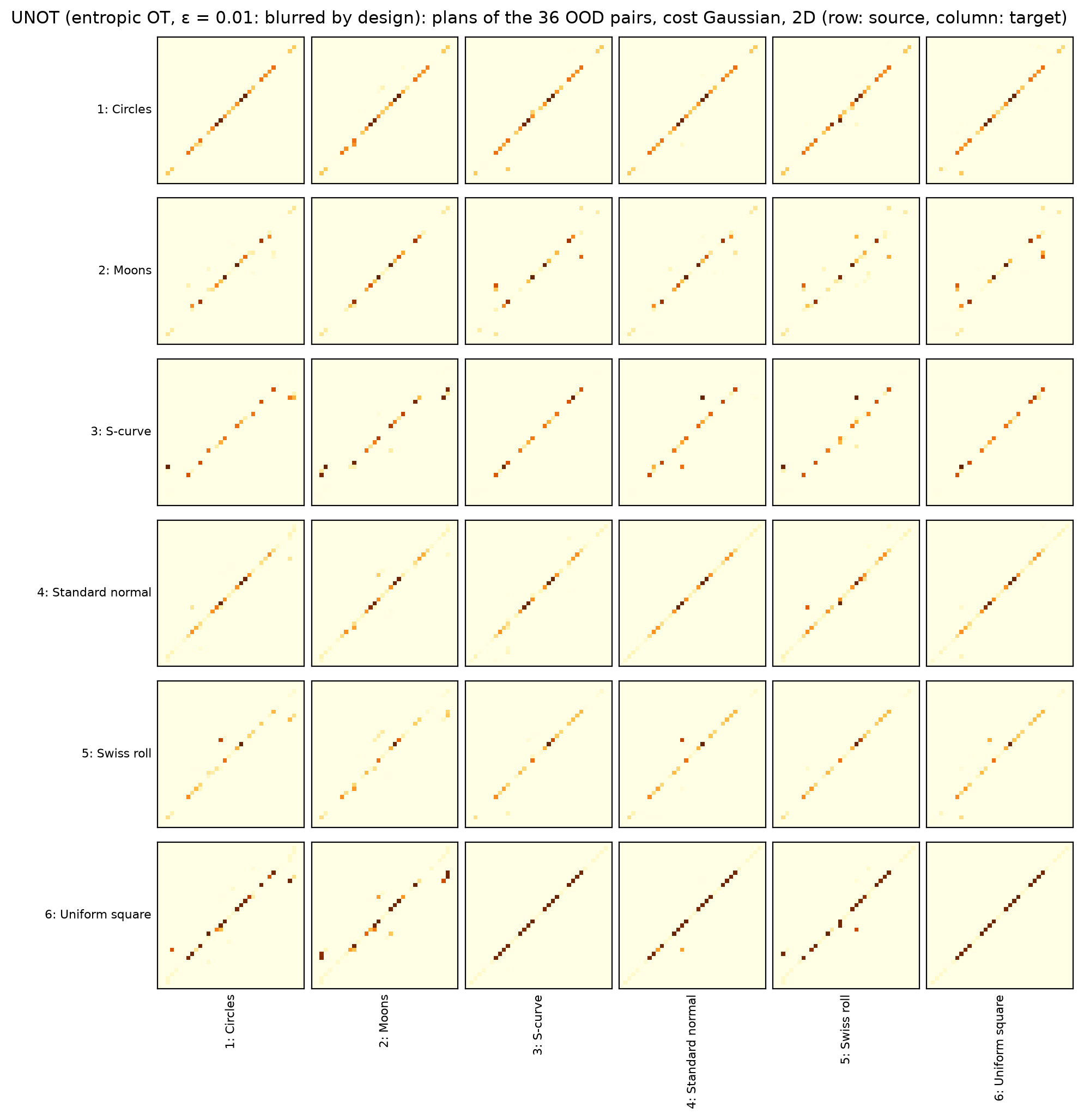}
    \caption{UNOT (entropic, $\varepsilon = 0.01$): plans of the 36 OOD pairs, cost Gaussian, 2D.}
    \label{fig:appF-pairs-gaussian-unot}
\end{figure}

\clearpage
\subsection{All six cost types}
\label{app:full-2d-all}

\begin{figure}[!ht]
    \centering
    \includegraphics[width=\linewidth]{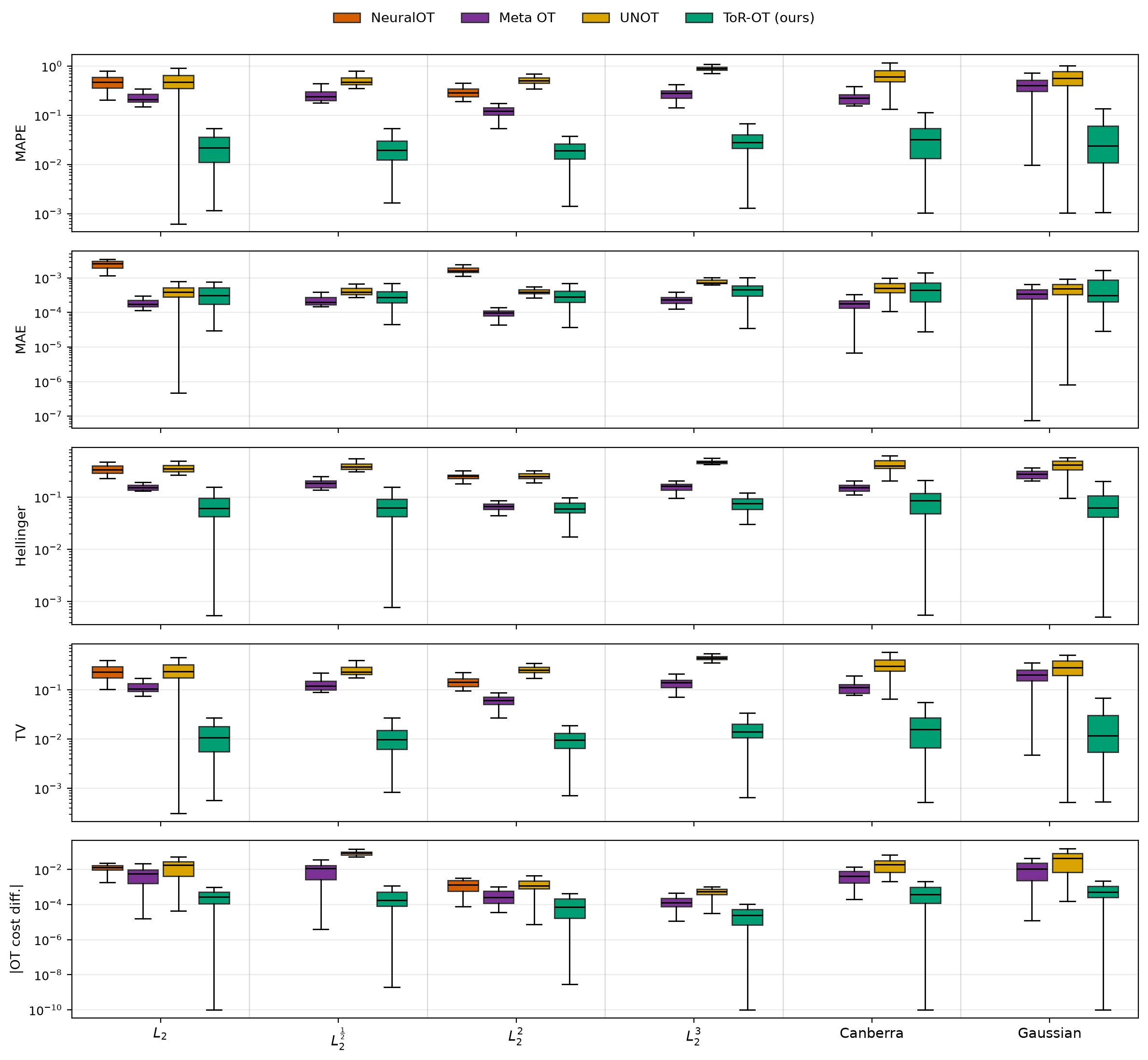}
    \caption{All six costs, 2D OOD set (36 pairs per cost): the five plan metrics, each method against the solution
    of its own problem (box: quartiles; line: median; whiskers: 1.5 IQR; outliers not shown; log scale). NeuralOT:
    $L_2^2$ and $L_2$ only.}
    \label{fig:appF-allcosts-ood}
\end{figure}

\begin{figure}[!ht]
    \centering
    \includegraphics[width=\linewidth]{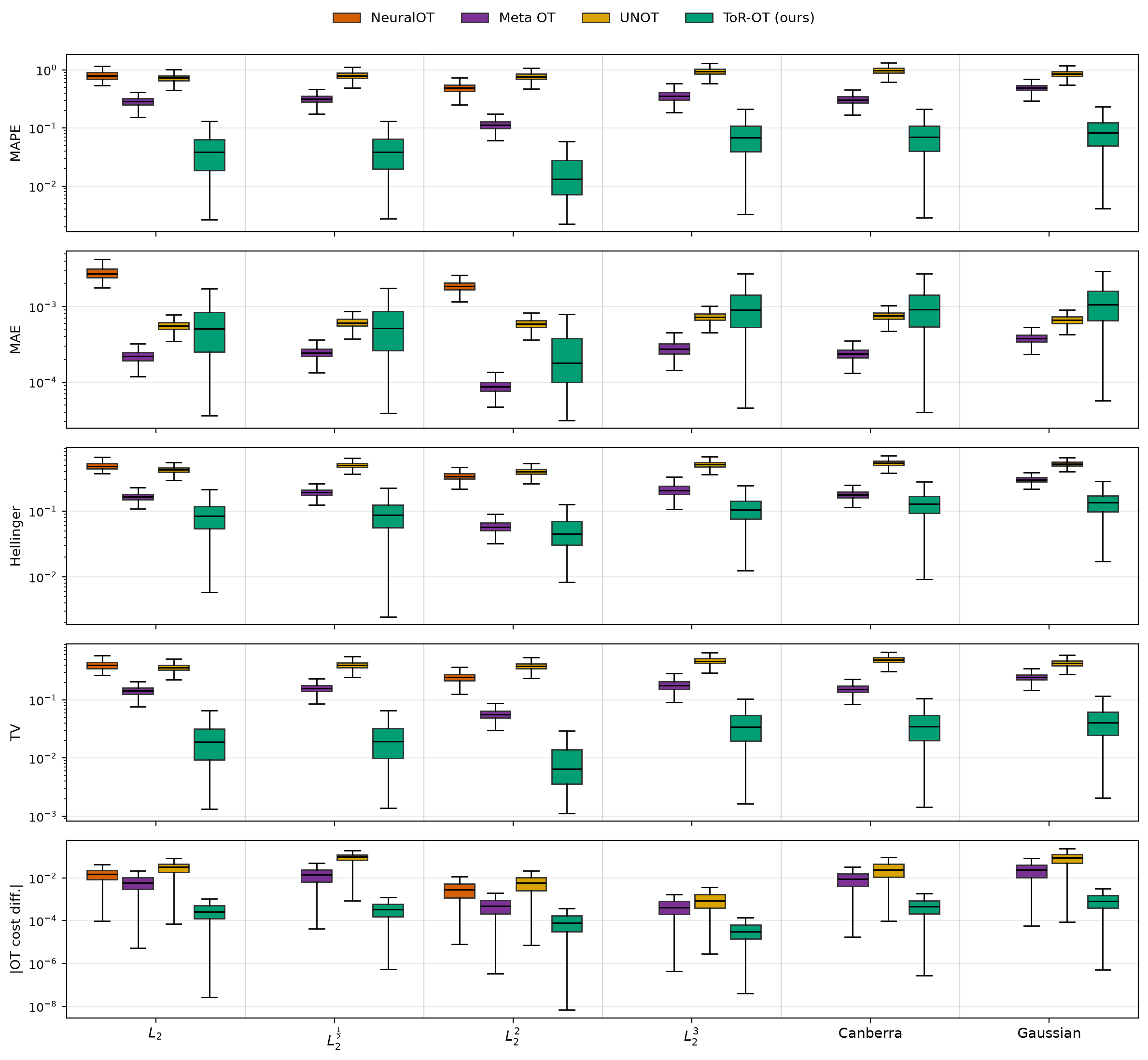}
    \caption{All six costs, 2D test set (1{,}000 pairs per cost; NeuralOT: 1{,}000 pairs for $L_2^2$, 200 for
    $L_2$), as Fig.~\ref{fig:appF-allcosts-ood}.}
    \label{fig:appF-allcosts-test}
\end{figure}

\begin{figure}[ht]
    \centering
    \includegraphics[width=\linewidth]{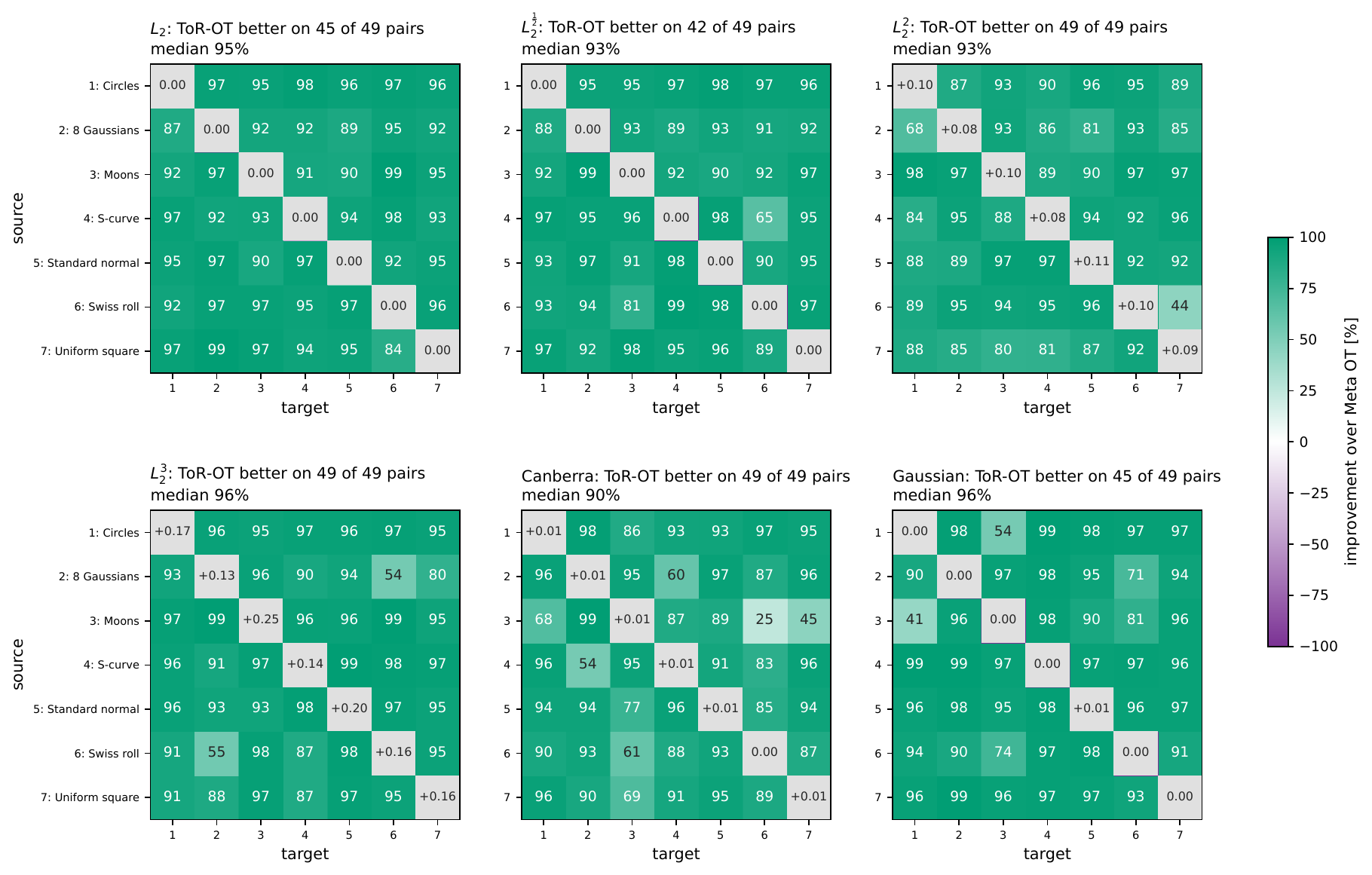}
    \caption{Relative MAPE improvement of ToR-OT over Meta OT, $100\,(\mathrm{MAPE}_{\text{Meta OT}} -
    \mathrm{MAPE}_{\text{ToR-OT}}) / \mathrm{MAPE}_{\text{Meta OT}}$, on each of the 36 2D OOD pairs (row:
    source; column: target; distributions numbered as in Fig.~\ref{fig:appF-distributions}), for each cost; each
    method is scored against the solution of its own problem. Green: ToR-OT better; purple: Meta OT better; the color
    scale is clipped at $\pm 100\%$. Diagonal (grey; source $=$ target): the absolute difference
    $\mathrm{MAPE}_{\text{Meta OT}} - \mathrm{MAPE}_{\text{ToR-OT}}$ instead (positive: ToR-OT better): there
    both errors can be close to 0 (Meta OT's below $10^{-4}$ for some costs), so their ratio is not meaningful. Panel
    titles: the number of pairs on which ToR-OT is better and the median improvement.}
    \label{fig:appF-improvement}
\end{figure}

Figures~\ref{fig:appF-allcosts-ood} and~\ref{fig:appF-allcosts-test} show every method on every cost, one panel per
metric. Fig.~\ref{fig:appF-improvement} shows, per OOD pair, how much ToR-OT improves over Meta OT, the strongest
baseline on every cost: ToR-OT is better on 29 to 35 of the 36 pairs, with median improvements from 81\% to
90\%. Of the 22 pairs (over all six costs) on which Meta OT is better, 18 are on the diagonal (source $=$ target),
where both methods are close to the reference (Meta OT's MAPE is lower by at most 0.011).

\end{document}